\documentclass[pdflatex,sn-mathphys-ay,iicol]{sn-jnl}

\usepackage{graphicx}%
\usepackage{multirow}%
\usepackage{makecell}%
\usepackage{amsmath,amssymb,amsfonts}%
\usepackage{amsthm}%
\usepackage{mathrsfs}%
\usepackage[title]{appendix}%
\usepackage{textcomp}%
\usepackage{manyfoot}%
\usepackage{booktabs}%
\usepackage{algorithm}%
\usepackage{algpseudocode}%
\usepackage{listings}%
\usepackage{xargs}                      
\usepackage[pdftex,dvipsnames]{xcolor}  

\usepackage{url}
\usepackage{etoolbox}
\usepackage{siunitx}
\usepackage{geometry}
\usepackage{esvect}
\usepackage{mathtools}
\usepackage{setspace}
\usepackage{pifont}
\usepackage{tabularx}
\usepackage{bm}
\usepackage{nicematrix}
\usepackage{caption}
\usepackage{subcaption}
\usepackage{threeparttable}
\usepackage{comment}
\usepackage{enumitem}
\usepackage{colortbl}
\usepackage{multirow}
\usepackage{multicol}
\usepackage{adjustbox}

\theoremstyle{thmstyleone}%
\theoremstyle{thmstyletwo}%
\theoremstyle{thmstylethree}%
\newcommand{\cmark}{\ding{51}}%
\newcommand{\xmark}{\ding{55}}%
\newcommand\restr[2]{{
  \left.\kern-\nulldelimiterspace 
  #1 
  \littletaller 
  \right|_{#2} 
  }}
\newcommand{\littletaller}{\mathchoice{\vphantom{\big|}}{}{}{}}
\newcommand{\RN}[1]{%
  \textup{\uppercase\expandafter{\romannumeral#1}}%
}

\begin{document}

\title[Article Title]{Towards Symmetry-sensitive Pose Estimation: A Rotation Representation for Symmetric Object Classes
}

\author*[1,2]{\fnm{Andreas} \sur{Kriegler}}\email{andreas.kriegler@tuwien.ac.at}\author[1]{\fnm{Csaba} \sur{Beleznai}}\email{csaba.beleznai@ait.ac.at}
\author[2]{\fnm{Margrit} \sur{Gelautz}}\email{margrit.gelautz@tuwien.ac.at}

\affil[1]{\orgdiv{Assistive and Autonomous Systems}, \orgname{AIT Austrian Institute of Technology}, \orgaddress{\street{Giefinggasse 4}, \city{Vienna}, \postcode{1210}, \country{Austria}}}
\affil[2]{\orgdiv{Visual Computing and Human-Centered Technology}, \orgname{TU Wien}, \orgaddress{\street{Favoritenstraße 9-11}, \city{Vienna}, \postcode{1040},  \country{Austria}}}

\abstract{Symmetric objects are common in daily life and industry, yet their inherent orientation ambiguities that impede the training of deep learning networks for pose estimation are rarely discussed in the literature. To cope with these ambiguities, existing solutions typically require the design of specific loss functions and network architectures or resort to symmetry-invariant evaluation metrics. In contrast, we focus on the numeric representation of the rotation itself, modifying trigonometric identities with the degrees of symmetry derived from the objects' shapes. We use our representation, SARR, to obtain canonic (symmetry-resolved) poses for the symmetric objects in two popular 6D pose estimation datasets, T-LESS and ITODD, where SARR is unique and continuous w.r.t. the visual appearance. This allows us to use a standard CNN for 3D orientation estimation whose performance is evaluated with the symmetry-sensitive cosine distance AR\textsubscript{C}. Our networks outperform the state of the art using AR\textsubscript{C} and achieve satisfactory performance when using conventional symmetry-invariant measures. Our method does not require any 3D models but only depth, or, as part of an additional experiment, texture-less RGB/grayscale images as input. We also show that networks trained on SARR outperform the same networks trained on rotation matrices, Euler angles, quaternions, standard trigonometrics or the recently popular 6d representation -- even in inference scenarios where no prior knowledge of the objects' symmetry properties is available. Code and a visualization toolkit are available at \url{https://github.com/akriegler/SARR}.}

\maketitle
\noindent\textit{\small This version of the article has been accepted for publication, after peer review but is not the Version of Record and does not reflect post-acceptance improvements or corrections. The Version of Record is available at: \url{http://dx.doi.org/10.1007/s11263-026-02770-x}. Springer Nature SharedIt link is: \url{https://rdcu.be/fb3QU}.}
\section{Introduction}\label{sec_intro}
\begin{figure*}[!t]
    \centering
    \includegraphics[width=\linewidth]{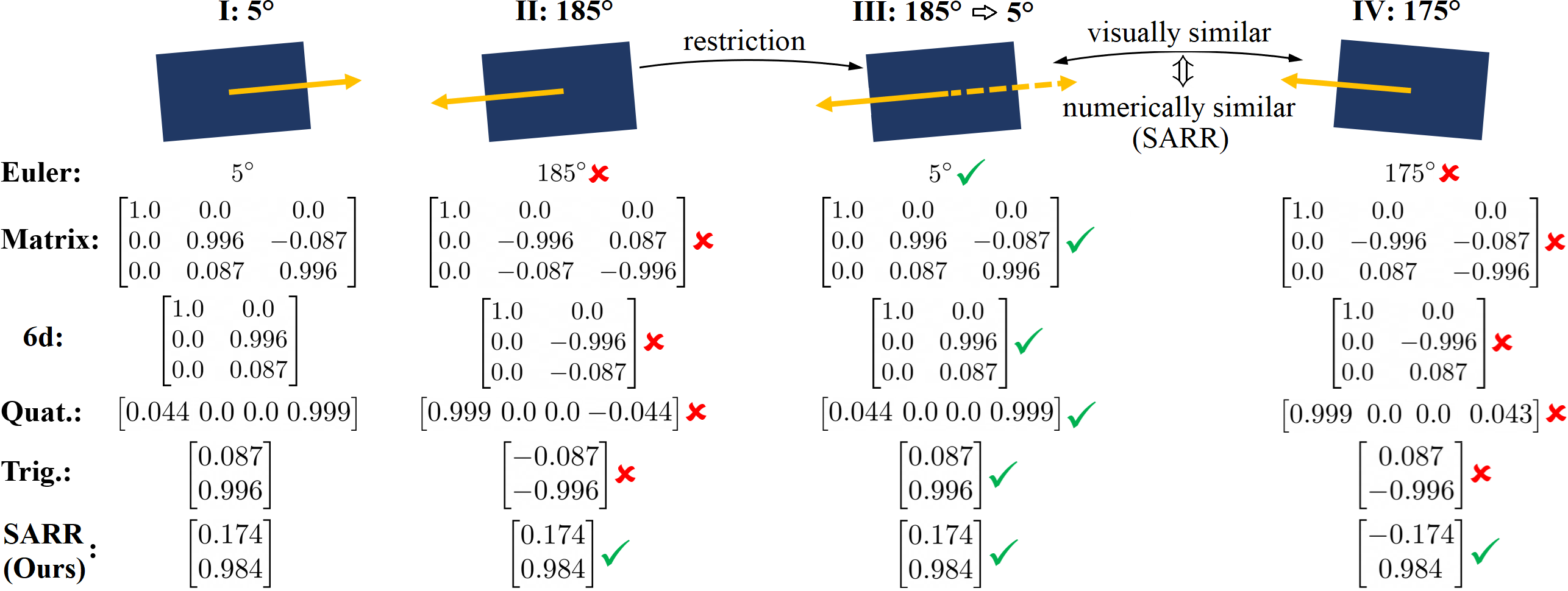}
    \caption{\textbf{Comparing rotation representations on a symmetric shape.} Using a rotation of $5^{\circ}$ as reference (I), rotating the rectangle by another $180^{\circ}$ creates ambiguities due to visual equality but numeric differences (II). A space restriction can resolve this, numerically treating the object as if it were rotated by $5^{\circ}$ (dashed arrow) (III). \textit{Yet this does not ensure continuity for visually similar poses close to the restriction} boundary ($\mathit{175^{\circ} \leftrightarrow 5^{\circ}}$) (IV). Contrarily, our formulation is both unique across the rotation space and continuous across the symmetry boundary transition. \textcolor{green}{\cmark} marks a correct/similar representation, \textcolor{red}{\xmark} is incorrect. ``6d'' denotes the 6d representation proposed by \citet{zhou2019a}.}
    \label{fig_explanation}
\end{figure*}

6D pose estimation is a prerequisite to enable robot grasping and manipulation tasks such as bin picking, as well as virtual and augmented reality applications. With the adoption of deep learning approaches and the usage of large training datasets, significant progress has been made for object pose estimation \citep{chen2023, he2022, irshad2022, periyasamy2022, liu2021, pitteri2021, pitteri2019}, yet the challenges related to rotationally symmetric objects, specifically the ambiguities that arise, are rarely discussed thoroughly. \\
It is sometimes possible to resolve these ambiguities by using symmetry-breaking visual features from the RGB domain such as non-symmetric color patterns like distinctive design elements on manufactured goods. At the same time, an important motivation of our work is that texture-less objects are common in industrial or robotics applications \citep{drost2017, hodan2017, kalra2024}. Furthermore, depth cameras or 3D LiDAR pointclouds are common modalities in robotics \citep{raj2022}. Thus, we focus on the challenging task of object pose estimation where \textit{geometric} symmetries are intrinsic properties due to an object's shape \citep{periyasamy2022, bregier2018} and assume no symmetry-breaking RGB texture to help resolve the ambiguities.\\ 
These ambiguities are prohibitive when optimizing a network to estimate the orientation \citep{hara2017}, since the bijective relation between the \textit{visual} representation $\mathcal{V}(\bm{p})$, e.g. an image of an object in pose $\bm{p}$, and the corresponding \textit{numeric} rotation representation $\mathcal{N}(\bm{p})$, e.g. a rotation matrix, no longer holds: multiple orientations represent the same visual input and the resulting multi-modal distribution cannot be learned unambiguously. To better understand the problems raised by symmetric objects, let us consider the rectangles in columns I and II of \textbf{Fig.}~\ref{fig_explanation}. For other rotation representations -- we consider Euler angles, rotation matrices, the 6d representation \citep{zhou2019a}, quaternions and trigonometric identities -- two numerically different representations apply to visually identical poses $\bm{p}_{\RN{1}}, \bm{p}_{\RN{2}}$:
\begin{equation}\label{eq_1}
\begin{split}
    \mathcal{V}(\bm{p}_{\RN{1}}) = \mathcal{V}(\bm{p}_{\RN{2}}) \text{ but } \mathcal{N}(\bm{p}_{\RN{1}}) \neq \mathcal{N}(\bm{p}_{\RN{2}}).
\end{split}
\end{equation}
While the rectangles are visually indistinguishable, ``physically'' they are oriented differently (marked by the yellow arrow) and thus have different numeric representations. There is therefore no function $\mathcal{F}\text{: }\mathcal{V}(\bm{p}) \mapsto \mathcal{N}(\bm{p})$ for all possible $\bm{p} \in \text{SO(3)}$. As \citet{pitteri2019} state, an attempt to learn such a function with a Machine Learning (ML) algorithm such as a Convolutional Neural Network (CNN) would fail for many loss functions, as the network converges naively to the mean of possible poses (if at all). This is true not only for rotation matrices but also other rotation representations \citep{huynh2009}.

Current pose estimation methods typically attempt to solve this problem in two ways: Firstly, the network architectures and/or loss functions are modified to resolve ambiguities. This was originally proposed by \citet{pitteri2019}, who split the rotation space into multiple bins, train a regression network for each bin as well as a classifier to pick the regressor to invoke. A disadvantage of this discretization approach is that it requires multiple networks and thus extends training times. Secondly, evaluation often relies on metrics that are symmetry-invariant; i.e. metrics that give a full score as long as the network predicts 1 out of the $n$ correct poses. Examples include the VSD, MSSD and MSPD metrics \citep{hodan2020}, which have become standard for pose estimation through the BOP benchmark \citep{hodan2018}. This means there is less incentive for methods to consider the ambiguities that stem from object symmetries  - something we consider problematic, in line with \citet{bregier2017}'s reluctance to rely on ambiguity-invariant error functions.   

A third approach deals with the representation $\mathcal{N}$ itself by restricting the rotation space to only the canonic object poses \citep{rad2017}, see column III of \textbf{Fig.}~\ref{fig_explanation}. The angle is clamped to the canonic space, numerically setting the orientation as marked by the dashed arrow. This restores uniqueness, i.e.:
\begin{equation}\label{eq_2}
\begin{split}
    \mathcal{V}(\bm{p}_{\RN{1}}) = \mathcal{V}(\bm{p}_{\RN{3}}) \Leftrightarrow \mathcal{N}(\bm{p}_{\RN{1}}) = \mathcal{N}(\bm{p}_{\RN{3}}),
\end{split}
\end{equation}\noindent%
however the mathematical properties and consequences were not analysed in detail in that work. Most obviously, any representation built on angles from this canonic space is discontinuous across the boundary of the space, e.g. $180^{\circ}\pm \epsilon \leftrightarrow 0^{\circ}\pm \epsilon$ for the shown rectangle. We take inspiration from their approach as we explore these object symmetries more formally and propose the \textbf{S}ymmetry-\textbf{A}ware-\textbf{R}otation-\textbf{R}epresentation (SARR), which implements a space restriction while remaining continuous across the boundary (column IV in \textbf{Fig.}~ \ref{fig_explanation}). We evaluate SARR using the symmetry classes of the texture-less objects in the T-LESS \citep{hodan2017} and ITODD \citep{drost2017} datasets. These have, as part of the BOP benchmark suite, become highly popular choices for evaluating object pose estimation methods. Our contributions are therefore:
\begin{enumerate}
    \item We formally describe the ambiguities due to object symmetries, targeting the symmetries and rotations of T-LESS and ITODD objects
    \item We propose SARR, a rotation representation that is unique and continuous for those objects 
    \item We present rotation estimation results, comparing to methods from the benchmark and networks trained on common representations
\end{enumerate}
After reviewing related works on symmetry and pose estimation in Section \ref{sec_related}, we provide a formal problem description and introduce the SARR representation in Section \ref{sec_methodology}. We present extensive evaluation results on T-LESS and ITODD in Section \ref{sec_experiments}, before concluding in Section \ref{sec_conclusion}.

\section{Related Work}\label{sec_related}
Pose estimation methods for symmetric objects can be split into two families \citep{periyasamy2022}, a categorization we also follow. Methods from the first category typically estimate only one pose (\ref{subsec_sota_single}), whereas methods from the second category estimate the full distribution of valid poses (\ref{subsec_sota_distribution}). We finish with a brief discussion on the desirable properties of rotation representation for machine learning tasks (\ref{subsec_sota_properties}).

\subsection{Single Valid Pose}\label{subsec_sota_single}
\citet{rad2017}'s work was one of the first to explicitly consider the problem of ambiguities due to symmetries. For their solution, they halve the rotation space of an object with a symmetry of degree 2 and train a classifier to detect if a given object falls outside this restricted space. If so, the image and predicted pose are mirrored. For higher degrees of symmetry, a similar discretization of the viewing sphere was done by \citet{corona2018}. \citet{pitteri2019} provide an analytic solution and implement a mapping algorithm for ambiguous rotations. In their work multiple regressors have to be trained dependent on an object's symmetry. A classifier is additionally required, which gets invoked to choose the correct regressor. Mapping to a canonic pose and introducing empirically set balance parameters was done in \cite{chen2021}, with four such parameters required to control network training.\\
\citet{morrison2020} have recognized the usefulness of the trigonometric representation to learn a grasping angle which is symmetric at $\pm \frac{\pi}{2}$. They restrict the angle to $[-\frac{\pi}{2}, \frac{\pi}{2}]$ and compute the vector $\bm{v} = (\sin2\theta, \cos2\theta)$ for network training. The same approach was taken by \citet{ayoub2023} for grasping tree-logs. Nevertheless, only one axis of symmetry was considered in both of these works that use the trigonometric representation. \\
Another popular idea is to use a specific distance metric \citep{zhao2023, mo2022, shi2021, labbe2020, wang2019}, implemented via a custom loss that fits closely to the evaluation metric. For example, \citet{mo2022} proposed the symmetry-invariant metric A(M)GPD (average (maximum) grouped primitives distance) which we also include in our experiments for evaluation. \citet{shi2021} had to train two separate networks, one for symmetric objects and one for asymmetric objects, each with a proper loss.\\
It is also possible to implicitly learn representations via an Autoencoder, as was done by \citet{sundermeyer2018, sundermeyer2020}. The learned latent representation is (presumably) ambiguity invariant, as visually identical poses will map to the same code. Unfortunately, their method relies strongly on accurate 3D models. \citet{haugaard2023} more recently proposed a way to learn 2D-3D correspondence distributions from color images without prior knowledge about symmetries. The authors admit that these distributions are accurate only ``up to'' symmetry ambiguities and postulate that such global symmetries could be modeled explicitly. In our work we specifically propose such a solution to model the symmetries of T-LESS and ITODD objects with a well-defined rotation representation.

\subsection{Distribution of Pose Hypotheses}\label{subsec_sota_distribution}
Methods of this family estimate not just one pose, but the full set of potential poses according to the object's symmetry. This is commonly done through correspondence matching with a probabilistic approach, where multiple likelihoods for the pose hypotheses are estimated. For example, \citet{zhao2023} learn many-to-many correspondences in two input point clouds, or \citet{li2024} use a render-and-compare approach and soft labels for the classification task. The possible distribution of 3D orientations was learned in the form of a $\text{SO}(3)$ encoder in \citep{cai2022}. As \citet{li2024} mention, the performance of matching-based methods can degrade significantly for imperfect CAD models. As these methods do not resolve to a single canonic pose, they have to resort to reporting results using a symmetry-invariant metric. In contrast, we train our networks without using any 3D models and report results using a symmetry-sensitive metric.

\subsection{Desired Properties of a Rotation Representation}\label{subsec_sota_properties}
While \textit{uniqueness} may be sufficient for analytic settings, for optimization algorithms such as deep neural networks, the target-variables and thus the target function $\mathcal{F}$ should also be \textit{continuous}. This continuity property is desirable, as discontinuities can inhibit learning \citep{hara2017}. \citet{xu2005, xu2004} have shown that functions that are smoother or have stronger continuity properties also have lower approximation errors for a given number of neurons. None of the common representations fulfill this, as they do not consider the symmetries of an object but only its orientation in the form of an abstract, three-dimensional Carthesian coordinate system.\\
To summarize, the works most closest to ours include \citet{pitteri2019}'s mapping to canonic poses, but our proposed SARR requires only a single regressor and preserves continuity, as well as the modified trigonometric representations used by \citet{morrison2020} and \citet{ayoub2023}, but we generalize their concept to multiple axes of symmetry and higher degrees of symmetry.

\section{Methodology}\label{sec_methodology}
We first formally describe the notions of \textit{uniqueness} and \textit{continuity} (\ref{subsec_problem_description}) before reviewing the BOP datasets and selecting T-LESS and ITODD for our analysis (\ref{subsec_bop_datasets}). For the remainder of Section 3 we focus on the symmetric objects in T-LESS (\ref{subsec_tless_symmetries})\footnote{See \textbf{Appendix}~\ref{app_A} for ITODD objects and 3D primitives.}. We then present our rotation representation SARR (\ref{subsec_sarr}) and its inverse mapping (\ref{subsec_sarr_inverse}). The section concludes with a visual validation of our representation (\ref{subsec_sarr_validation}) and a discussion of its limitations (\ref{subsec_sarr_limiations}).

\subsection{Problem Description}\label{subsec_problem_description}
Types of symmetries include translational, reflectional or rotational symmetry in 2D (see \citet{stewart2013}, pg. 3). Yet in 3D, a reflectional symmetry can be understood as rotational symmetry and translational symmetry is less relevant for Computer Vision, since translationally symmetric objects (e.g. a long building with identical parts) are uncommon and CNNs are generally accepted to be translation equivariant \citep{lenc2019}. We therefore limit our discussion to geometric symmetries that arise purely from rotations. We refer to the finite set of symmetric poses of an object as ``discrete'' and use ``continuous'' for the infinite circular symmetries such as those of a cone or sphere. Note that a cylinder actually has both discrete and continuous symmetries, as we define this property per axis. For the subsequent task of pose estimation, since we focus on rotational symmetry, we disregard object translation. That is, we analyse the 3D orientation of objects and not their full 6D pose, i.e. assume $\bm{p} \in \text{SO(3)}$.\\
Then, let $\mathcal{V}(\bm{p})$ be the visual representation of an arbitrary object in pose $\bm{p}$, for example a depth image $\bm{d}\in \mathbb{R}^{w \times h}$ (or an RGB image of a texture-less object). An object is said to be symmetric if there exist one or more rigid motions $\bm{r}$ which, if applied to the object pose, do not change the appearance of the object \citep{pitteri2019}. In this case there exists a non-empty set $\mathcal{R}$:
\begin{equation}\label{eq_3}
    \begin{split}
        \mathcal{R} = &\{\bm{r} \in \text{SO(3)}\setminus I_{3}\text{ s.t. } \\
        &\forall \bm{p} \in \text{SO(3): } \mathcal{V}(\bm{p}) = \mathcal{V}(\bm{r}\cdot\bm{p}).
    \end{split}
\end{equation}
Unlike Pitteri et al. we exclude the identity motion $I_{3}$ as otherwise every object could be considered symmetric. For a rotation representation $\mathcal{N}$ to be considered unique, i.e. to resolve the ambiguities (I and II in \textbf{Fig.}~\ref{fig_explanation}), we require that: 
\begin{equation}\label{eq_4}
    \mathcal{V}(\bm{p}) = \mathcal{V}(\bm{r}\cdot\bm{p}) \Leftrightarrow \mathcal{N}(\bm{p}) = \mathcal{N}(\bm{r}\cdot\bm{p}).
\end{equation}
Operating on the numerical representation level, this can be done by restricting the rotations to the space of only canonic object poses $C \subset \text{SO(3)}$. Such a restriction is both necessary and sufficient to satisfy \textbf{Eq.}~(\ref{eq_3}). However, this effectively disregards all object symmetries and treats objects as non-symmetrical. An obvious drawback is the introduced discontinuity: objects near the symmetry boundary look visually similar (due to their symmetry) yet their numerical representations are drastically different (due to the clamping). More formally, we require that for small rotations $\bm{\epsilon}$ which result in small visual changes, the numeric representations must only differ slightly  - within the canonic space \textit{and across its boundary} - to be considered continuous:
\begin{equation}\label{eq_5}
       \mathcal{V}(\restr{\bm{p}}{C}) \approx \mathcal{V}(\bm{\epsilon} \cdot \bm{r}\cdot\restr{\bm{p}}{C}) \Leftrightarrow \mathcal{N}(\restr{\bm{p}}{C}) \approx \mathcal{N}(\bm{\epsilon} \cdot \bm{r}\cdot\restr{\bm{p}}{C}).
\end{equation}

\citet{pitteri2019} partially solve this problem by training multiple regressors and an additional classifier. Each regressor is continuous within its respective domain $C_{n}$ and invoked via the classifier prediction. They then evaluate their method on T-LESS objects. Our method requires no space discretization and we evaluate not only on the T-LESS dataset.

\subsection{Symmetries in BOP Datasets}\label{subsec_bop_datasets}
\begin{table}[h]
\footnotesize
\setlength{\tabcolsep}{0.2mm} 
\renewcommand{\arraystretch}{1.0} 
\centering
\caption{\textbf{BOP datasets overview.}}
\label{tab_datasets}
\begin{tabular}{|c|cccccc|}
     \hline
     \textbf{Dataset} & \# &  $|\mathcal{S}'_{i}|$ & $|\mathcal{S}_{i, t}|$ & $u$ & GT\textsubscript{V} & GT\textsubscript{T}\\
     \hline
     HANDAL \citep{guo2023} & 40 & 7 & 7  & 2 & \cmark & \xmark\\
     HB \citep{kaskman2019} & 33 & 5 & 3 & 2 & \cmark & \xmark\\
     HOPE \citep{tyree2022} & 28 & 28 & 0 & 0 & \cmark & \xmark\\
     HOT3D \citep{banerjee2025} & 33 & 21 & 10  & 6 & \cmark & \xmark\\
     IC-BIN \citep{doumanoglou2016} & 2 & 1 & 1 & 1 & \xmark & \cmark\\
     IC-MI \citep{tejani2014} & 6 & 3 & 2 & 1 & \xmark & \cmark\\
     IPD \citep{kalra2024} & 10 & 5 & 5 & 5 & \cmark & \cmark\\
     ITODD \citep{drost2017}  & 28 & 18 & 18 & 11 & \cmark & \xmark\\
     ITODD-MV \citep{drost2017} & 28 & 18 & 18 & 11 & \cmark & \xmark\\
     LM \citep{hinterstoisser2013} & 15 & 3 & 3 & 2 & \xmark & \cmark\\
     LM-O \citep{brachmann2014} & 8 & 2 & 2 & 1 & \xmark & \cmark\\
     RU-APC \citep{rennie2016} & 14 & 8 & 1 & 1 & \xmark & \cmark\\
     T-LESS \citep{hodan2017} & 30 & 27 & 27 & 5 & \xmark & \cmark\\
     TUD-L \citep{hodan2018} & 3 & 0 & 0 & 0 & \xmark & \cmark\\
     TYO-L \citep{hodan2018} & 21 & 13 & 8 & 2 & \xmark & \cmark\\
     XYZ-IBD \citep{huang2025} & 15 & 9 & 9 & 5 & \cmark & \cmark\\
     YCB-V \citep{calli2017} & 21 & 14 & 7 & 5 & \xmark & \cmark\\
     \hline
\end{tabular}
\end{table}
\noindent The BOP benchmark features 17 datasets (at the time of writing) with numerous objects -- column \# in \textbf{Table}~\ref{tab_datasets} gives the total number of objects. For every dataset $i$, let $\mathcal{S}_{i}$ denote the set of its symmetric objects, i.e. all objects with non-empty $\mathcal{R}$. $\mathcal{S}_{i}$ itself was never published by the BOP organizers. Organizers did identify $\mathcal{S}_{i}$ -- by using the HALCON software and calculating the Hausdorff distance $h$ between vertices of an object model in the canonical and transformed locations, identifying a symmetry if $h < \text{max}(15mm,0.1d)$, where $d$ is the diameter of the model\footnote{See 7.2 at \url{https://bop.felk.cvut.cz/challenges/bop-challenge-2019/} (Accessed: 2026-01-12)} -- yet $\mathcal{S}_{i}$ is not made public\footnote{See \url{https://github.com/thodan/bop_toolkit/issues/50} (Accessed: 2026-01-12).}. We estimated the sets $\mathcal{S}'_{i}$ instead via visual inspection of the CAD models. Column $|\mathcal{S}'_{i}|$ gives the size of these sets, i.e. the total number of symmetric objects. Different from our approach of considering symmetries defined solely by an object's geometry, the BOP organizers used texture to derive $\mathcal{S}_{i, t} \subseteq \mathcal{S}_{i}$. Specifically, $\mathcal{S}_{i, t}$ consists of objects with ``symmetry transformations that cannot be resolved by the model texture''\footnote{See 7.2 at \url{https://bop.felk.cvut.cz/challenges/bop-challenge-2019/} (Accessed: 2026-01-12).}. For some datasets such as HOPE, which contains objects with highly distinctive textures, this removes the majority of objects from consideration (column $|\mathcal{S}_{i, t}|$). Nevertheless, we consider $\mathcal{S}_{i, t}$ as the starting point for our analysis, as to not deviate from established conventions ($\mathcal{S}_{i, t}$ is provided with every dataset). For every dataset, column $u \coloneqq u(\mathcal{S}_{i, t})$ gives the number of unique symmetry classes amongst all the objects in that dataset. Observing columns  $|\mathcal{S}_{i, t}|$ and $u$, T-LESS, ITODD, ITODD-MV (ITODD with additional multi-view images), HOT3D, XYZ-IBD and TYO-L all feature a large and diverse set of objects and symmetries. We disregard TYO-L, because the leaderboard has been inactive since 2019, HOT3D, because only RGB fisheye image streams are available as a modality (no depth) and XYZ-IBD, because training images were rendered with CAD models whose object origins are defined differently than those from the test set, causing a mismatch. Instead we focus on T-LESS and ITODD, both highly active datasets on the leaderboard. T-LESS has ground-truth 6D pose labels available for its test set (column GT\textsubscript{T}), whereas these labels are only provided for the validation set in ITODD (column GT\textsubscript{V}).

\subsection{Symmetries in T-LESS}\label{subsec_tless_symmetries}
\newcolumntype{C}{>{\centering\arraybackslash}X}%
\begin{table*}[!t]
\caption{\textbf{Symmetry classes of T-LESS objects.} Each of the five symmetry classes has a representation $\text{SARR}_{i}$, vector $\bm{\kappa}_{i}$, a symmetry type, a set of rotations $\mathcal{R}_{i}$ and a rotation space $T_{i}$. T-LESS class IDs of the visualized objects are indicated \textbf{bold} in the bottom row. The default 3D coordinate frame, as shown for class $\RN{2}$, is oriented the same for all objects.}
\centering
\includegraphics[width=1.0\linewidth]{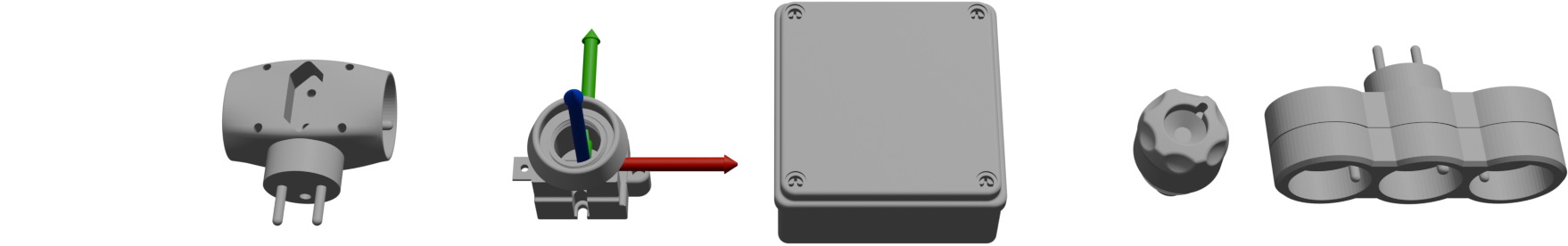}
\label{tab_tless_symmetries}
\setlength{\tabcolsep}{1.7mm} 
\renewcommand{\arraystretch}{1.0} 
\begin{tabular}{lccccc}
\toprule
\textbf{Class} & $\RN{1}$  & $\RN{2}$ & $\RN{3}$ & $\RN{4}$ & $\RN{5}$ \\
$\text{SARR}_{i}$ & $\text{SARR}_{\RN{1}}$ & $\text{SARR}_{\RN{2}}$ & $\text{SARR}_{\RN{3}}$  & $\text{SARR}_{\RN{4}}$ & $\text{SARR}_{\RN{5}}$ \\ 
$\bm{\kappa}_{i}$ & $\bm{\kappa}_{\RN{1}} = [\textcolor{red}{1}, \textcolor{green}{1}, \textcolor{blue}{1}]$ & $\bm{\kappa}_{\RN{2}} = [\textcolor{red}{1}, \textcolor{green}{1}, \textcolor{blue}{2}]$ & $\bm{\kappa}_{\RN{3}} = [\textcolor{red}{1}, \textcolor{green}{1}, \textcolor{blue}{4}]$ & $\bm{\kappa}_{\RN{4}} = [\textcolor{red}{1}, \textcolor{green}{1}, \textcolor{blue}{\infty}]$ & $\bm{\kappa}_{\RN{5}} = [\textcolor{red}{1}, \textcolor{green}{2}, \textcolor{blue}{1}]$ \\
\textbf{Type} & None & Discrete & Discrete & Continuous & Discrete  \\
$\mathcal{R}_{i}$ & $\{\}$ & $\bm{R}_{\textcolor{blue}{z}}^{\gamma} | \gamma \in \{\pi\}$ & $\bm{R}_{\textcolor{blue}{z}}^{\gamma} | \gamma \in \{\frac{\pi}{2}, \pi, \frac{3\pi}{4}\}$ & $\bm{R}_{\textcolor{blue}{z}}^{\gamma} | \gamma \in \{\mathbb{R}\}$  & $\bm{R}_{\textcolor{green}{y}}^{\beta} | \beta \in \{\pi\}$ \\
$T_{i}$ & $T$ & $T$ & $T$ & $T$ & $T_{1}$ \\
\begin{tabular}[c]{@{}c@{}}\textbf{T-LESS}\\ \textbf{IDs} \end{tabular} & \textbf{21}, 22, 18  & \begin{tabular}[c]{@{}c@{}}\textbf{11}, 5, 6, 7, 8, 9, 10,\\ 12, 25, 26, 28, 29 \end{tabular}  & \textbf{27} & \begin{tabular}[c]{@{}c@{}}\textbf{2}, 17, 1, 3, 4, 13,\\ 14, 15, 16, 24, 30 \end{tabular} & \textbf{23}, 19, 20  \\
\bottomrule
\end{tabular}
\end{table*}
\citet{mo2022} categorized the symmetric objects in T-LESS and YCB-V following the Hausdorff distance of model vertices, using a clustering algorithm for simplification while relying on a relaxation-threshold for continuous symmetries. We also classify the symmetries of T-LESS objects (see \textbf{Table}~\ref{tab_tless_symmetries}), but use the original BOP information without any extra parameters to resolve continuous symmetries. All 30 objects are split into one of 5 symmetry classes $i \in \{\RN{1}, \RN{2}, \RN{3}, \RN{4}, \RN{5}\}$ following $u$. Symmetry class $\RN{1}$ includes all non-symmetric objects. Class $\RN{2}$ and $\RN{3}$ are for objects with two-fold and four-fold discrete symmetry about $\textcolor{blue}{z}$, respectively. Class $\RN{4}$ contains all objects with continuous symmetry about $\textcolor{blue}{z}$, while class $\RN{5}$ exhibits a two-fold symmetry about $\textcolor{green}{y}$. This leads to another way of expressing an object's symmetry properties using the symmetry-vectors $\bm{\kappa}_{i} = [\bm{\kappa}_{i, \alpha}, \bm{\kappa}_{i, \beta}, \bm{\kappa}_{i, \gamma}]$, made up of the degrees of symmetry per object axis. For class \RN{2}, $\bm{\kappa}_{\RN{2}} = [\textcolor{red}{1}, \textcolor{green}{1}, \textcolor{blue}{2}]$ can be read as: ``within a full $360^{\circ}$ rotation about $\textcolor{blue}{z}$, there are two poses that are visually identical ($\gamma = 0^{\circ}, \gamma = 180^{\circ}$); only one such pose exists for axes $\textcolor{red}{x}$ and $\textcolor{green}{y}$ (no symmetry)''.
These 30 T-LESS objects appear in a rotation space that is already restricted to partially eliminate duplicate images due to symmetry \citep{hodan2016}. To analyse this, we build up the rotation space $T \subset \text{SO(3)}$ of the T-LESS training set. Image acquisition was divided into two steps, where rotations $\alpha_{1} \in \{5, 15, \dots, 85\}^{\circ}, \beta_{1} = 0^{\circ}, \gamma_{1} \in \{0, 5, \dots, 355\}^{\circ}$ and  $\alpha_{2} \in \{275, 285, \dots, 355\}^{\circ}, \beta_{2} = 0^{\circ}, \gamma_{2} \in \{0, 5, \dots, 355\}^{\circ}$ describe all object poses $\restr{\bm{p}}{T}$: $T_{1} = \{\bm{R}^{\gamma_{1}}_{\textcolor{blue}{z}''}\bm{R}^{\beta_{1}}_{\textcolor{green}{y}'}\bm{R}^{\alpha_{1}}_{\textcolor{red}{x}}\}$ and $T_{2} = \{\bm{R}^{\gamma_{2}}_{\textcolor{blue}{z}''}\bm{R}^{\beta_{2}}_{\textcolor{green}{y}'}\bm{R}^{\alpha_{2}}_{\textcolor{red}{x}}\}$, with $T = T_{1} \cup T_{2}$\footnote{Intrinsic rotations in ``XYZ''-order: $\bm{R}_{\textcolor{red}{x}}^{\alpha}$ then $\bm{R}_{\textcolor{green}{y}'}^{\beta}$ and $\bm{R}_{\textcolor{blue}{z}''}^{\gamma}$.}. $T$ represents a ``full view sphere'' \citep{hodan2016} of these objects with $|T| = 1296$ training images per object class. The jump of $180^{\circ}$  in $\alpha$ was not done for two objects, IDs 19, 20, meaning for those $T = T_1$\footnote{The combined $\alpha$ and $\gamma$ rotations effectively implement a $180^{\circ}$ flip about those object's symmetry axis, axis \textcolor{green}{y}, since objects 19 and 20 belong to symmetry class $\RN{5}$. According to Hodan et al.'s own definition, object 23 also belongs to class $\RN{5}$, yet images of object 23 in $T_{2}$ rotations do exist. We consider this an oversight, as images from $T_{1}$ and $T_{2}$ are visually identical (ignoring texture/lighting changes), and thus disregard these $T_{2}$ images.}. The fact that $\restr{\bm{p}}{T}$ still includes ambiguous poses, and that the default T-LESS annotations do not satisfy \textbf{Eq.}~(\ref{eq_4}) let alone \textbf{Eq.}~(\ref{eq_5}), can be verified by looking at the images in \textbf{Fig.}~\ref{fig_tless_comparison}; it also becomes apparent when considering, for example, that $\gamma \in \{0, 5, \dots, 355\}^{\circ}$ for continuously symmetric objects of class $\RN{4}$. 

\subsection{SARR Representation}\label{subsec_sarr}
To motivate our representation SARR, let us consider object 11 from symmetry class $\RN{2}$ (see \textbf{Table}~\ref{tab_tless_symmetries}) and only a rotation about axis $\textcolor{blue}{z}$. We can then imagine some function $f$ (for example a perceptual hash) that computes the similarity between the visual representations of the object in its default, unrotated pose and a second, rotated pose: $f(\mathcal{V}(\bm{p}), \mathcal{V}(\bm{R}^{\gamma}_{\textcolor{blue}{z}}\cdot\bm{p}))$. It follows that $f$ is periodic, being maximal at symmetrical poses, i.e. where $\gamma \in \{0, \pi, \dots, \frac{2n\pi}{\bm{\kappa}_{\RN{2}, \gamma}}\} \equiv \{n\pi\}, n \in \mathbb{Z}$. Similarly, $f$ is minimal exactly halfway, where the object poses are orthogonal. Considering $f$ instead of only the set $\mathcal{R}_{\RN{2}}$, one can derive that the numeric representation should be designed to show a similar periodic behavior. This is both intuitive for humans to understand and desirable from an optimization standpoint, as it does not introduce any discontinuities since $f$ is smooth. Analogous arguments can be made for all symmetry classes of T-LESS: the degree of symmetry for every axis defines the frequency of $f$ for rotations about that axis. For example, the frequency of $f$ for class $\RN{3}$ and $\textcolor{blue}{z}$ rotations is $\frac{2n\pi}{4}$ since $\bm{\kappa}_{\RN{3}, \gamma}=4$. 

We propose a rotation representation that implements such periodic behaviour by modifying trigonometric identities $\bm{v} = (\sin\theta, \cos\theta)$, where $\bm{v}$ represents a point on the unit circle as seen in \citep{hara2017}. To this end, we formalize and generalize the concept used in \citep{ayoub2023, morrison2020}. In contrast to those works, SARR is formulated generically and can be used to treat different axes of symmetry (\textcolor{green}{y}, \textcolor{blue}{z}) and different degrees of symmetry. For the trigonometric representation $\mathcal{N}_{\text{trig}}$ the composition of trigonometric identities $\bm{v}$ of angles $\alpha, \beta, \gamma$ defines the orientation of an object:
\begin{equation}\label{eq_6}
\begin{split}
    \mathcal{N}_{\text{trig}}(\alpha, \beta, \gamma)
    =
    \begin{bmatrix}
        \sin\alpha & \sin\beta & \sin\gamma \\
        \cos\alpha & \cos\beta & \cos\gamma
    \end{bmatrix}.
    \end{split}
\end{equation}
Combining \textbf{Eq.}~(\ref{eq_6}) with the discussion on $f$, we can define the rotation representation $\text{SARR}_{i}$ for all objects of all five symmetry classes in T-LESS as:
\begin{equation}\label{eq_7}
\begin{split}
    \text{SARR}_{i}(\restr{\alpha}{T_{i}}, \restr{\beta}{T_{i}}, \restr{\gamma}{T_{i}})
    =
    \begin{bmatrix}
        s_{i, \alpha} & s_{i, \beta} & s_{i, \gamma} \\
        c_{i, \alpha} & c_{i, \beta} & c_{i, \gamma} 
    \end{bmatrix}
    \coloneqq
    \\
    \begin{bmatrix}
        \sin(\bm{\kappa}_{i, \alpha}  \restr{\alpha}{T_{i}}) & \sin(\bm{\kappa}_{i, \beta}  \restr{\beta}{T_{i}}) & \sin(\lambda_{i} \restr{\gamma}{T_{i}}) \\
        \cos(\bm{\kappa}_{i, \alpha}  \restr{\alpha}{T_{i}}) & \cos(\bm{\kappa}_{i, \beta}  \restr{\beta}{T_{i}}) & \cos(\lambda_{i}  \restr{\gamma}{T_{i}})
    \end{bmatrix}.
\end{split}
\end{equation}
To resolve the continuous symmetry of class $\RN{4}$, we define
\begin{equation}\label{eq_8}
    \begin{split}
        \lambda_{i} = 
        \begin{cases}
            0 & \text{if } i == \RN{4} \\
            \bm{\kappa}_{i, \gamma} & \text{otherwise.} 
        \end{cases}
    \end{split}
\end{equation}
\noindent
This formulation allows extension to other object symmetries and datasets. Specifically, the representation for symmetry classes of ITODD objects and 3D primitives is presented in \textbf{Appendix}~\ref{app_A}.

\subsection{SARR Inverse Mapping}\label{subsec_sarr_inverse}
While SARR is designed for machine learning tasks, evaluation and visualization requires an inverse mapping to Euler angles $\restr{\alpha_{i}}{C}, \restr{\beta_{i}}{C}$ and $\restr{\gamma_{i}}{C}$, from which other representations such as rotation matrices can be derived:
\begin{equation}\label{eq_9}
    \begin{split}
    \restr{\theta_{i}}{C} = 
        \begin{cases}
            0 \text{ if }  i == \RN{4}, \\
            \frac{1}{\bm{\kappa}_{i, \theta}}(2\pi-\arccos(c_{i, \theta})) & \text{if } s_{i, \theta} < 0, \\
            \frac{1}{\bm{\kappa}_{i, \theta}}\arccos(c_{i, \theta}) & \text{otherwise,}
        \end{cases}
    \end{split}
\end{equation}
for $\theta \in \{\alpha, \beta, \gamma\}$ (angles are in radians). Angles $\restr{\alpha_{i}}{C}, \restr{\beta_{i}}{C}, \restr{\gamma_{i}}{C}$ are now restricted to the subspace $\restr{\cdot}{C}$ and any representations derived from them do not have the continuity property of SARR, but are unique since they define canonic object poses and are thus pertinent for evaluation. Any permutations of these rotations that look visually identical are resolved. \textbf{Algorithm} \ref{alg_sarr} in \textbf{Appendix}~\ref{app_B} shows the full process of forward and inverse mapping.

\subsection{Validation of the Representation}\label{subsec_sarr_validation}
\begin{figure}[t]
\centering
\begin{subfigure}[b]{0.49\linewidth}
\centering
\includegraphics[width=\linewidth]{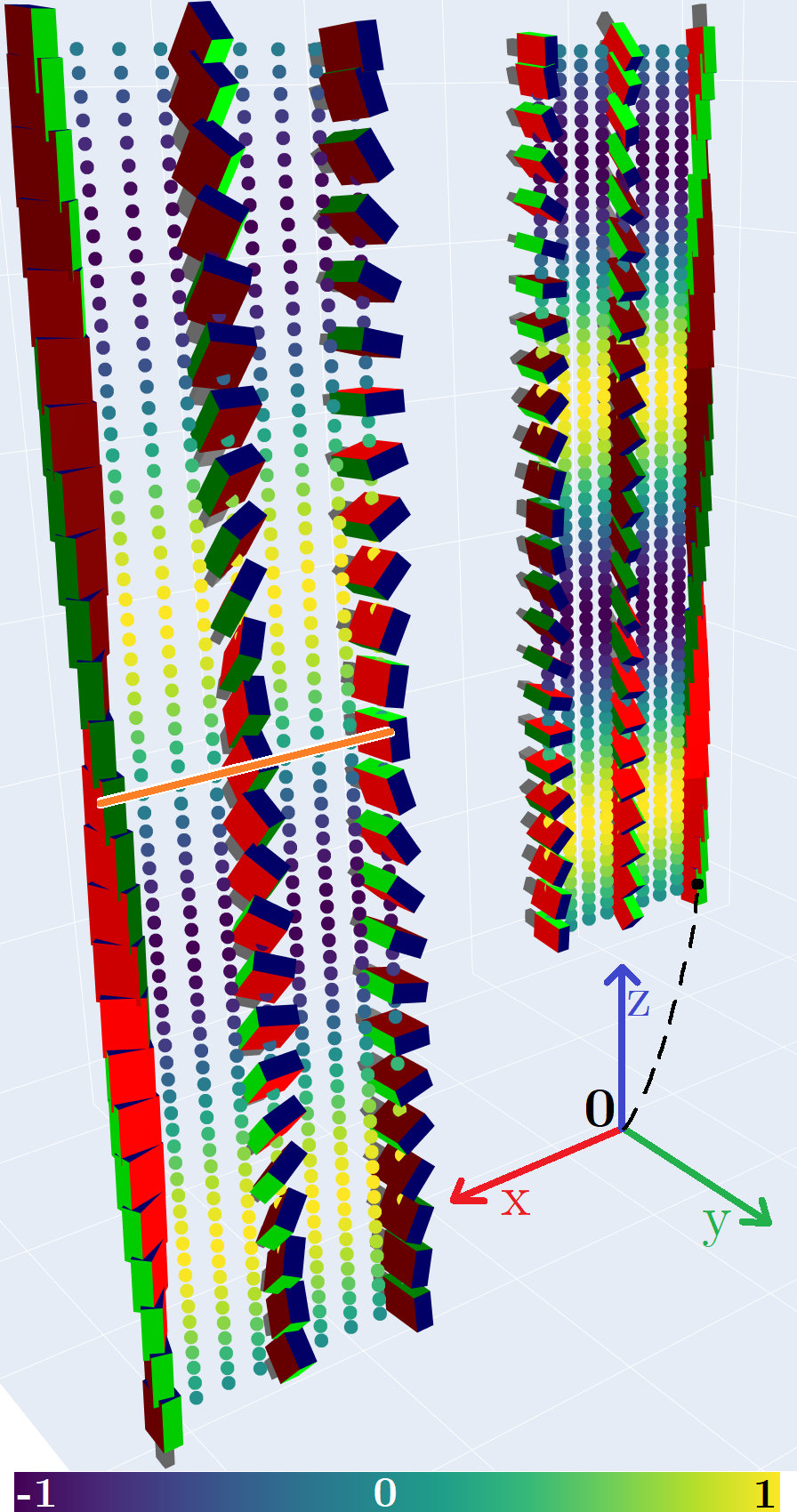}
\caption{Plot for $s_{\RN{2}, \gamma}$}
\label{fig_v1}        
\end{subfigure}
\begin{subfigure}[b]{0.49\linewidth}
\centering
\includegraphics[width=\linewidth]{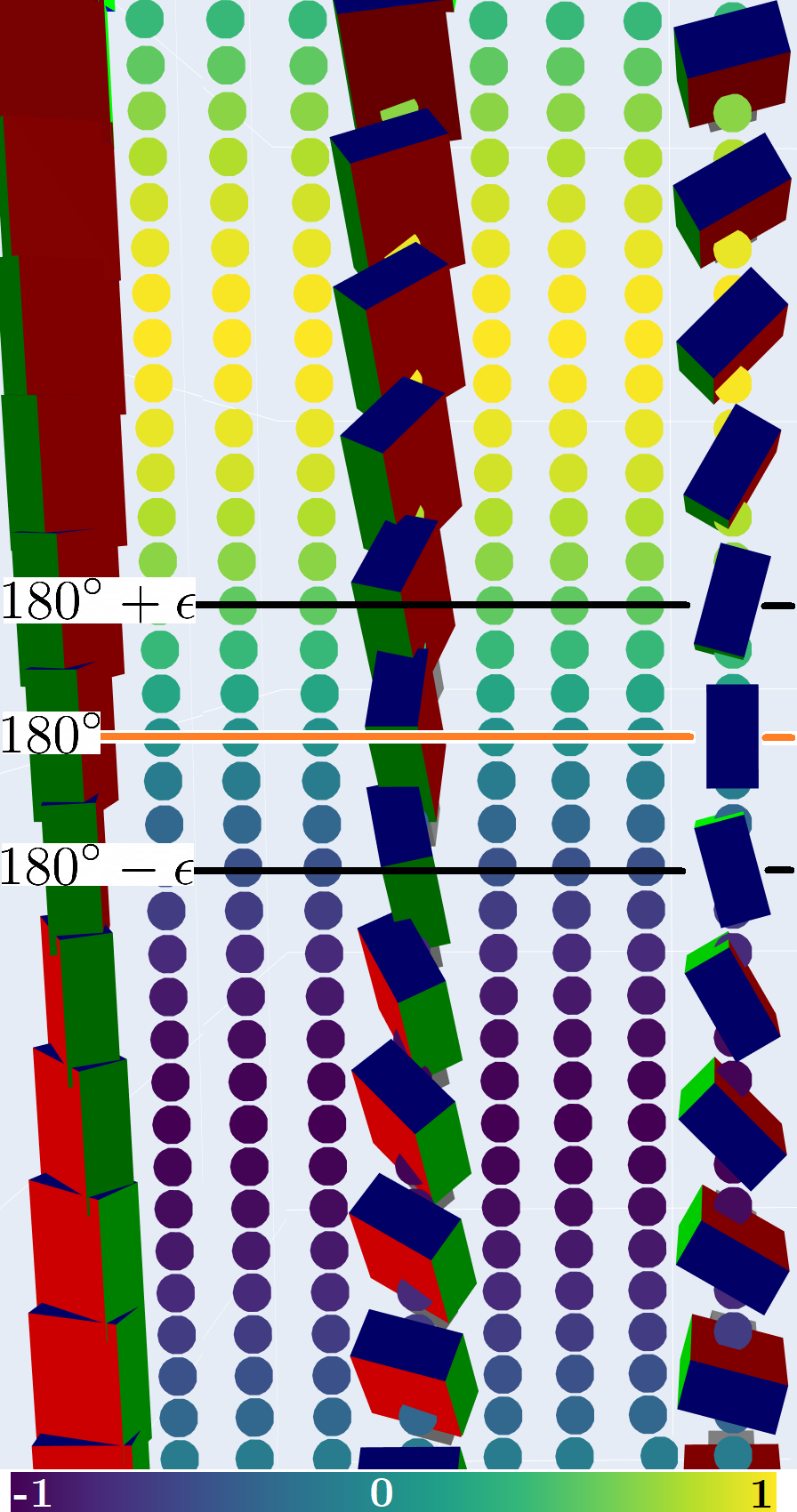}
\caption{$s_{\RN{2}, \gamma}$ near $180^{\circ}$}
\label{fig_v2}        
\end{subfigure}
\caption{\textbf{Visual verification of $\text{SARR}_{\RN{2}}$.} The entire plot for $s_{\RN{2}, \gamma}$ is shown in (a) and zoomed in near the symmetry boundary (orange line) in (b). Color of the grid represents the values of $s_{\RN{2}, \gamma}$.}
\label{fig_validation}
\end{figure}
\textbf{Fig.}~\ref{fig_validation} shows a visual validation for symmetry class $\RN{2}$, symmetric about $\textcolor{blue}{\text{z}}$. A cuboid-like object, with a detail to break the symmetry for axes $\textcolor{red}{\text{x}}$ and $\textcolor{green}{\text{y}}$, is rotated across the space $T$ represented by the grid, with some angular sparsity to avoid cluttering the plot. The colors of this grid follow a heatmap, representing the values of $s_{\RN{2}, \gamma}$. Visually identical objects have dots of equal color (uniqueness), whereas visually similar objects, such as those near the symmetry-boundary at $180^{\circ}$ (horizontal orange line), have similarly colored dots (continuity). This is shown in more detail in the right-hand plot. The visualization tool used to create these plots, which we also make publicly available, can be used to analyse new symmetry classes and extended rotation spaces.

\subsection{Limitations}\label{subsec_sarr_limiations}
Uniqueness and continuity of SARR for these symmetry classes is guaranteed within $T$ but not necessarily for the full space of rotations SO(3). This is because rotations about two non-symmetric axes can result in an object that is visually identical to one rotated about its symmetry axis, for example $\mathcal{V}_{\RN{5}}(180, 0, 180) = \mathcal{V}_{\RN{5}}(0, 180, 0)$ but $\text{SARR}_{\RN{5}}(180, 0, 180) \neq \text{SARR}_{\RN{5}}(0, 0, 0)$. Early results show that an additional space restriction, or mapping to an auxiliary, intermediate representation, can alleviate this problem. Regarding other symmetries, SARR is also valid for objects with multiple axes of symmetry, such as the ITODD objects or 3D primitives, although additional terms in the representation are necessary and inverse mapping can require extra steps. 

\section{Experiments}\label{sec_experiments}
We use the SARR representation to train neural networks to estimate the orientation of objects in images from the T-LESS and ITODD datasets, using depth as the principal input modality. To this end, we first present results of mapping T-LESS rotation annotations $\bm{R}$ to $\restr{\bm{R}}{C}$, as a form of sanity check to verify the correctness of SARR (\ref{subsec_exp_mapping}). We then describe evaluation metrics (\ref{subsec_exp_metrics}) and outline the experimental setup (\ref{subsec_exp_setup}). For the estimation task we compare against both other methods from the leaderboard as well as our network but trained with different rotation representations. We present results on the T-LESS (\ref{subsec_exp_tless}) and ITODD (\ref{subsec_exp_itodd}) datasets, including an ablation study of SARR-networks trained and tested using other input modalities besides depth. The section concludes with a summary and discussion of our results (\ref{subsec_exp_summary}).

\subsection{T-LESS Annotation Mapping}\label{subsec_exp_mapping}
\begin{figure*}[!t]
    \centering
    \begin{subfigure}[b]{0.19\linewidth}
        \centering
        \includegraphics[width=\linewidth]{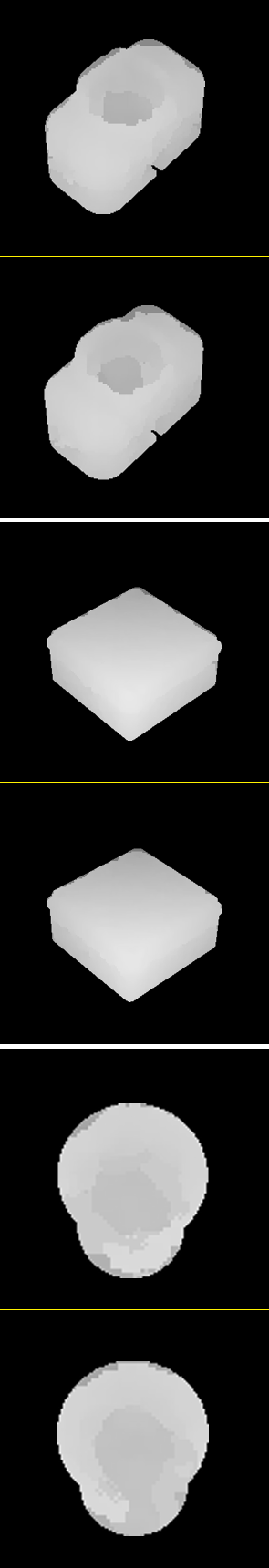}
        \caption{Depth map $\bm{d}$}
        \label{fig_fa}
    \end{subfigure}
    \begin{subfigure}[b]{0.19\linewidth}
        \centering
        \includegraphics[width=\linewidth]{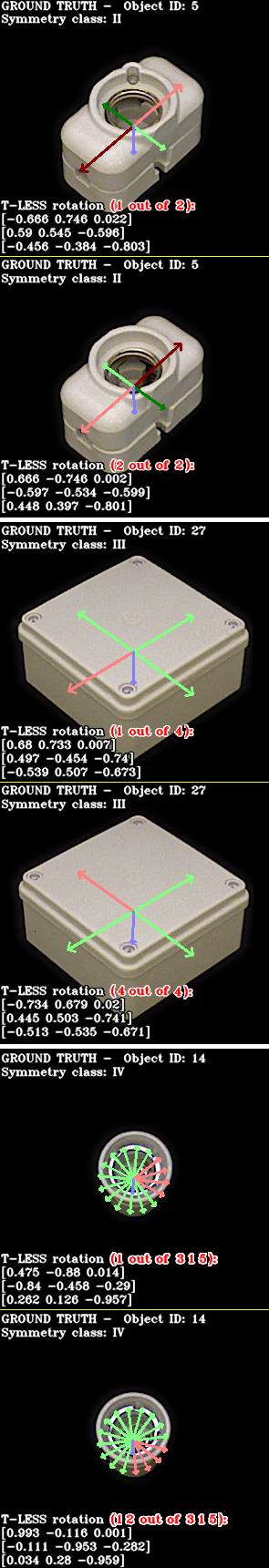}
        \caption{T-LESS GT $\bm{R}$}
        \label{fig_fb}
    \end{subfigure}
    \begin{subfigure}[b]{0.19\linewidth}
        \centering
        \includegraphics[width=\linewidth]{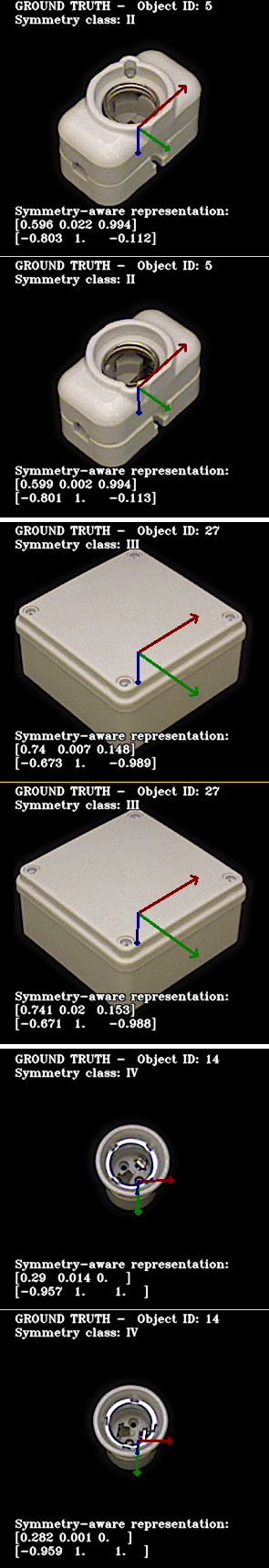}
        \caption{Our GT $\restr{\bm{R}}{C}$}
        \label{fig_fc}
    \end{subfigure}
    \begin{subfigure}[b]{0.19\linewidth}
        \centering
        \includegraphics[width=\linewidth]{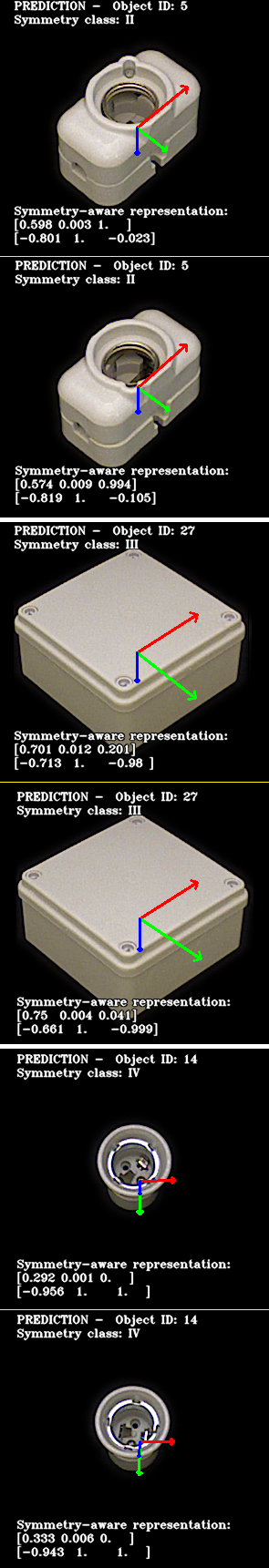}
        \caption{Predictions $\hat{\bm{R}}$}
        \label{fig_fd}
    \end{subfigure}
    \begin{subfigure}[b]{0.19\linewidth}
        \centering
        \includegraphics[width=\linewidth]{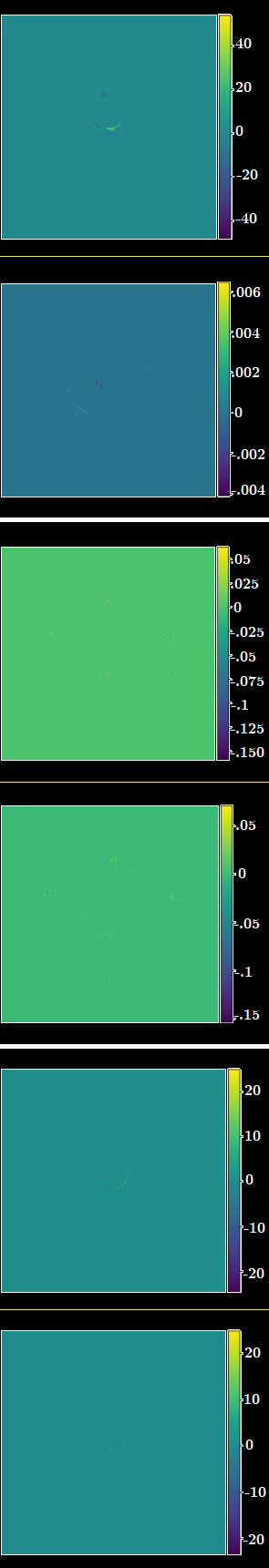}
        \caption{Depth-diff. $\delta$}
        \label{fig_fe}
    \end{subfigure}
    \caption{\textbf{SARR annotation mapping of T-LESS symmetry classes $\RN{2}$, $\RN{3}$ and $\RN{4}$.}}
    \label{fig_tless_comparison}
\end{figure*}

To illustrate the advantage of using our rotation representation over default T-LESS annotations, \textbf{Fig.}~\ref{fig_tless_comparison} presents mapping results for the symmetry classes with ambiguous object poses in $T$: $\RN{2}, \RN{3}, \RN{4}$. For every class there are two rows of object images in poses that are visually (nearly) identical, yet numerically, at least according to the T-LESS annotations, very different. For rows 5 and 6 this discrepancy is especially noticeable: class $\RN{4}$ requires a discretization using T-LESS annotations, per default resulting in 315 distinct yet correct rotation matrices. The columns show (a) the real, pre-processed depth maps $\bm{d}$ we use for training our networks (see \textbf{Appendix}~\ref{app_C1}), (b) the RGB training images used in our ablation study with two of the multiple ambiguous T-LESS rotation matrices $\bm{R}$ overlayed, (c) our unique annotations $\restr{\bm{R}}{C}$, derived from the canonic Euler angles $\restr{\alpha_{i}}{C}, \restr{\beta_{i}}{C}, \restr{\gamma_{i}}{C}$ (d) network predictions $\hat{\bm{R}}$ using our annotations and (e) depth difference $\delta$ as false color maps. Depth difference was computed between rendered depth images\footnote{These depth images were rendered only for the presented calculation.} of objects in poses according to our ground truth $\restr{\bm{R}}{C}$ ($\bm{d}_{Ours}$) vs. T-LESS ground truth $\bm{R}$ ($\bm{d}_{T-LESS}$):
\begin{equation}\label{eq_60}
    \delta = \sum_{O}\sum_{\bm{d} \in \mathcal{D}}\sum_{\text{px} \in \text{mask}} |\bm{d}_{\text{Ours}}  - \bm{d}_{\text{T-LESS}}|,
\end{equation}
for all 30 objects $O$, all depth images $\mathcal{D}$ of those objects and all pixels $\text{px}$ inside the respective visibility masks. We report an average depth discrepancy of $\delta = 0.5\text{mm}$ across the dataset, proving that our process of mapping followed by inverse mapping results in poses that are visually identical, up to deviations due to symmetry imperfections which fall below the threshold used in the HALCON scripts (see Section \ref{subsec_bop_datasets}). Per-object depth difference is highest for object 25, which is a light switch defined to be symmetric, yet the tilted switch-panel noticeably breaks this symmetry in many views. We also report a much higher average difference between real camera depth and rendered T-LESS depth of $\delta_{\text{Cam, T-LESS}} = 7.4\text{mm}$ as well as $\delta_{\text{Cam, Ours}} = 7.8\text{mm}$ for camera depth and our rendered depth due to the camera depth showing various artifacts and corruptions.

\subsection{Evaluation Metrics}\label{subsec_exp_metrics}
For the orientation estimation experiments we calculate five different error functions: $e$, VSD, MSSD, MSPD and A(M)GPD from which we derive three evaluation metrics AR\textsubscript{C}, AR\textsubscript{B} and AR\textsubscript{G}.\\
In line with the BOP benchmark, we report  AR\textsubscript{B} as the average of recalls under VSD, MSSD, and MSPD scores \citep{hodan2020} to evaluate object orientation estimates $\hat{\bm{R}}$. We use the official BOP toolkit\footnote{See \url{https://github.com/thodan/bop_toolkit}, specifically the script \texttt{eval\_bop19\_pose.py} (Accessed: 2026-01-12).} to calculate these results, leaving all parameters unchanged.\\
We also calculate the rotational error $e\in[0, 180]^{\circ}$ derived from the cosine distance \citep{hodan2016}:
\begin{equation}\label{eq_11}
    e(\hat{\bm{R}}, \restr{\bm{R}}{C}) = \arccos(\frac{\text{Tr}(\hat{\bm{R}}\restr{\bm{R}}{C}^{-1})-1}{2})\frac{180}{\pi}.
\end{equation}
In the rare event that a method did not provide a prediction we assign the maximum error of $180^{\circ}$\footnote{When evaluating our networks we encountered two T-LESS depth maps that had NaN values. For other methods listed in \textbf{Table}~\ref{tab_tless_other_results} it is unknown why some estimates are missing.}. We then calculate the average recall AR\textsubscript{C} taking the average of recall scores across the $e$ thresholds $\{2, 5, 10, 15, 25, 40\}^{\circ}$ \citep{kriegler2022, kriegler2023}.\\
We also provide results using the A(M)GPD metric proposed by \citet{mo2022}, which we denote AR\textsubscript{G} for brevity. AR\textsubscript{G} is a pose distance metric that uses the ground-truth and estimated poses, the object class and predefined sets of grouped primitives. These grouped primitives incorporate information on symmetry permutations and were derived from the 3D model by Mo et al.. We reimplemented the calculation of the AR\textsubscript{G} metric using Equations 12 and 13 from \citep{mo2022} as well as their source-code\footnote{Specifically, we created a stand-alone, GPU-independent reimplementation of the \texttt{eval\_metric} function from file \texttt{tless\_gadd\_evaluator.py}: \url{https://github.com/GANWANSHUI/ES6D/blob/master/lib/tless_gadd_evaluator.py}. This reimplementation is available in our repository: \url{https://github.com/akriegler/SARR/blob/main/source/metrics/amgpd.py}.  (Accessed: 2026-01-12).}. We can only report AR\textsubscript{G} results on T-LESS, because the sets of grouped primitives were not defined for the ITODD objects.\\
Metrics AR\textsubscript{B} and AR\textsubscript{G} are both symmetry-invariant by design due to the properties of their underlying error functions, meaning that any one of the $n$ potential poses for the same visual appearance is accepted. In contrast, for AR\textsubscript{C} there is always only one correct orientation: $\restr{\bm{R}}{C}$. In the context of this work, we consider AR\textsubscript{C} to be a stricter metric, punishing methods that are ambiguous in their estimations. AR\textsubscript{B} and AR\textsubscript{G} on the other hand are tolerant towards such ambiguous orientation estimations.\\
Lastly, we report results for two tasks: estimating the orientation for only the most visible instance of every object class in every image (SiSo), as well as for the $n$-most visible instances (ViVo), where $n$ is provided with the annotation and is different for every object and image. When calculating results for the ViVo task, it is necessary to match the sets of pose ground-truths and pose estimates, for all instances of each object class apparent in any image. We treat this as a linear assignment problem, where the sets resemble the two partites, and solve it via minimum weight matching, computing a cost matrix using the error function $e$ from \textbf{Eq.}~(\ref{eq_11}). We do this for metrics AR\textsubscript{C} and AR\textsubscript{G}, while the BOP toolkit handles the matching for AR\textsubscript{B}. We also report the average inference time of the orientation estimation for the ViVo task, where measurement starts right after an input sample $i$ is loaded and ends when the orientation prediction $\hat{\bm{R}}_{i}$ is available (as the final $3\times3$ rotation matrix).

\subsection{Experimental Setup}\label{subsec_exp_setup}
Columns ``Training'' and ``Test'' in \textbf{Table}~\ref{tab_tless_other_results} provide information regarding the training/validation data used for T-LESS, \textbf{Table}~\ref{tab_itodd_other_results} shows the same information for ITODD. Specifically, for training our networks on T-LESS we use the real ``PrimeSense'' depth images as input and no 3D-models, while for our ablation study we train the SARR-networks using RGB images from the same ``PrimeSense'' sensor instead. For training on ITODD we use the synthetic images from a physically-based renderer (PBR) instead: depth images per default and RGB images converted to grayscale for the ablation study  (\textbf{Appendix}~\ref{app_C1} explains the data preprocessing steps).\\
We evaluate on the real test depth/RGB images of T-LESS and the real validation depth/grayscale images of ITODD. The ITODD validation/test sets include no RGB but only grayscale images, and we use the validation set because we require ground truth 6D pose labels to calculate AR\textsubscript{C}, AR\textsubscript{B} and AR\textsubscript{G}, which are not publicly available for the test but only the validation set (see \textbf{Table}~\ref{tab_datasets}).  We use ground-truth translation for all results that we report: the networks we trained ourselves as well as results from other methods. For other methods we downloaded the original result files from the leaderboard and replaced the estimated translation with the ground-truth.\\ 
Considering other works from the benchmark leaderboard, it is common practice to train one network for every single object class, sometimes also across the entire dataset. We extend this approach and train multiple additional networks, one for every symmetry class, and thus identify these different scopes for our experiments:
\begin{description}[labelwidth=1cm, leftmargin=1cm]
    \item[object:] one CNN for each object class, hence 30 for T-LESS and 28 for ITODD,
    \item[symmetry:] one CNN for each symmetry class, hence 5 for T-LESS and 9 for ITODD,
    \item[dataset:] one CNN for T-LESS, one for ITODD, 
    \item[dataset*:] one CNN for T-LESS, one for ITODD -- both with an additional symmetry classification task.
\end{description} 
\textit{Dataset}-networks in the literature are sometimes designed in such a way that they do not require external object class information during inference. But since we need to know the symmetry-class for inverse mapping SARR, which we normally derive from the object-class, we define the additional scope \textit{dataset*}. Here the SARR-\textit{dataset*} networks have to predict the symmetry class of a given object from dataset $i$ from the symmetry classes $u(\mathcal{S}_{i, t})$.\\
We use PyTorch \citep{paszke2019} to train a modified CenterNet \citep{zhou2019} network with HardNet \citep{chao2019} as backbone and optimize using Adam \citep{kingma2017}. We use cosine distance and L1 loss for optimizing the rotation parameters. We do not use the geodesic loss \citep{mahendran2017} for rotation matrices as this would require modifying the network, e.g. by adding another layer \citep{salehi2019}, which would result in an unfair comparison. We use FocalLoss \citep{lin2017} for the \textit{dataset*} networks since we embed the symmetry classification task in heatmaps. For more information regarding implementation and loss functions see \textbf{Appendix}~\ref{app_C2} and \textbf{Appendix}~\ref{app_C3}, respectively.\\

\subsection{T-LESS}\label{subsec_exp_tless}
\begin{table*}[!t]
\resizebox{\linewidth}{!}{%
\begin{threeparttable}[b]
\caption{\textbf{T-LESS orientation estimation.} SARR--Depth-networks outperform all other methods under AR\textsubscript{C}. 1\textsuperscript{st}, 2\textsuperscript{nd} and 3\textsuperscript{rd} best methods are \textbf{bold}, \underline{underlined} and \textit{italic} respectively. Column ``t[s]'' shows the inference time in seconds. D denotes depth, R real and S synthetic images. Methods ModalOcc.rgbd/rgb/depth and ZTE\_PPF are unpublished.}
\label{tab_tless_other_results}
\centering
\setlength{\tabcolsep}{2pt} 
\renewcommand{\arraystretch}{1.0} 
\begin{tabular}{|c|ccccc|ccc|ccc|c|}
\hline
\multirow{2}{*}{\textbf{Method}} & \multirow{2}{*}{\textbf{Training}} & \multirow{2}{*}{\textbf{Test}} & \multirow{2}{*}{\textbf{Scope}} & \multirow{2}{*}{\textbf{Model}} & \multirow{2}{*}{\textbf{Symm.}} & \multicolumn{3}{c|}{\textbf{SiSo}} & \multicolumn{3}{c|}{\textbf{ViVo}} & \multirow{2}{*}{t[s]}\\
 & & & & & & \multicolumn{1}{c}{AR\textsubscript{C}$\uparrow$} & \multicolumn{1}{c}{AR\textsubscript{B}$\uparrow$}  & \multicolumn{1}{c|}{AR\textsubscript{G}$\uparrow$} & \multicolumn{1}{c}{AR\textsubscript{C}}
 & \multicolumn{1}{c}{AR\textsubscript{B}} & \multicolumn{1}{c|}{AR\textsubscript{G}} & \\
\hline
\citet{hodan2015} & templates & RGBD & unknown & CAD & evaluation & 17.1 & 70.1 &  66.2 & 14.8 & 61.0 &  65.4 & 80.1 \\
\citet{sundermeyer2020} & RGB: R,S & RGBD & dataset & CAD & network & 14.0 & 62.3 &  62.4 & 12.7 & 59.7 &  64.7 & .531 \\
\citet{wang2021} & RGBD: R,S & RGBD & object & CAD & loss & 33.5 & \textbf{90.4} &  \textit{90.7} & 30.9  & 91.2 &  \textit{87.3} & 6.63 \\
\citet{su2022} & RGBD: R,S & RGBD & object & reconstr. & data & 38.5 & 85.6 &  \textbf{91.2} & 36.7 & 90.6  &  87.2 & 2.62 \\
ModalOcc.rgbd (2024) & RGBD: S & RGBD & unknown & CAD & unknown & 30.8 & 79.8 &  90.4 & 29.3 & \underline{94.1} &  \underline{87.6} & 7.24 \\
\cite{liu2025} & RGBD: R,S & RGBD & object & reconstr. & loss & 32.0 & \underline{89.4} &  \underline{91.1} & 29.6 & \textbf{95.1} &  86.8 & 2.48 \\
\hline
\cite{cai2022} & RGB: S & RGB & dataset & \ding{55} & network & 30.2 & 75.1 &  85.9 & 27.4 & 87.1 & 86.0 & 50.8\\
\citet{castro2023} & RGB: S  & RGB & dataset & CAD & loss & 24.8 & 73.9 & 83.7 & 23.2 & 80.5 & 80.9 & .059 \\
ModalOcc.rgb (2024) & RGBD: S  & RGB & unknown & CAD & unknown & 32.6 & 69.1 & 90.0 & 30.3 & 93.1 & \textbf{88.3} & 7.75 \\
\cite{liu2025} & RGB: R,S  & RGB & object & reconstr. & loss & 33.7 & \textit{89.1} & 90.6 & 30.2 & \textit{93.4} & 85.6 & .214 \\
\hline
\citet{drost2010} & templates & D & unknown & reconstr. & repres. & 17.9 & 70.3 &  64.2 & 15.5 & 60.9 &  63.8 & 9.20 \\
\citet{vidal2018} & templates & D & unknown & reconstr. & repres. & 22.6 & 76.0 &  74.0 & 19.5 & 68.9 &  71.5 & 7.06 \\
ZTE\_PPF (2022) & templates & D & unknown & reconstr. & unknown & 21.9 & 79.3 &  73.8 & 19.3 & 75.1 &  74.5 & .846 \\
ModalOcc.depth (2024) & RGBD: S & D & unknown & CAD & unknown & 27.3 & 81.8 &  83.4 & 23.6 & 80.6 &  82.0 & 4.72 \\
\hline
\multirow{4}{*}{\parbox{2.3cm}{\textbf{SARR}--Depth}} & D: R & D & \textit{object} & \ding{55} & repres. & 45.0 & 48.0 &  68.4 & 41.9 & 58.1 &  67.0 & .077 \\
 & D: R & D & \textit{symmetry} & \ding{55} & repres. & \underline{47.5} & 56.3 &  70.0 & \underline{44.0} & 60.1 &  69.4 & .078  \\
 & D: R & D & \textit{dataset} & \ding{55} & repres. & \textbf{48.0} & 50.3 &  69.8 & \textbf{44.8} & 60.0 &  69.4 & .074 \\
& D: R & D & \textit{dataset*} & \ding{55} & repres. & 46.7 & 54.9 &  68.8 & 42.9 & 59.0 &  66.7 & .077 \\
\hline
\multirow{4}{*}{\parbox{2.1cm}{\textbf{SARR}--RGB}} & RGB: R & RGB & \textit{object} & \ding{55} & repres. & 32.6 & 46.8 & 58.2 & 30.8 & 45.8 & 58.8 & .077 \\
 & RGB: R & RGB & \textit{symmetry} & \ding{55} & repres. & 29.1 & 42.4 & 49.9 & 28.9 & 42.1 & 53.3 & .077 \\
 & RGB: R & RGB & \textit{dataset} & \ding{55} & repres. & \textit{47.3} & 49.0 & 69.3 & \textit{43.6} & 58.7 & 69.1 & .078 \\
 & RGB: R & RGB & \textit{dataset*} & \ding{55} & repres. & 45.8 & 52.1 & 69.1 & 41.1 & 58.7 & 67.4 & .080 \\
\hline
\end{tabular}
\end{threeparttable}%
}
\end{table*}
\begin{table*}[!t]
\centering
\resizebox{\linewidth}{!}{%
\setlength{\tabcolsep}{2pt} 
\begin{threeparttable}[b]
\caption{\textbf{T-LESS representation comparison.} SARR is best suited for pose estimation of symmetric objects, outperforming the other representations across the different scopes, tasks and evaluation metrics.}
\label{tab_tless_rep_results}
\renewcommand{\arraystretch}{1.0} 
\begin{tabular}{|c|ccc|ccc|ccc|}
\hline
\multirow{2}{*}{\textbf{Representation}} & \multicolumn{3}{c|}{\textbf{SiSo} - AR\textsubscript{C}} & \multicolumn{3}{c|}{\textbf{SiSo} - AR\textsubscript{B}} & \multicolumn{3}{c|}{\textbf{SiSo} - AR\textsubscript{G}} \\
& \textit{object} & \textit{symmetry} & \multicolumn{1}{c|}{\textit{dataset(*)}} & \textit{object} & \textit{symmetry} & \multicolumn{1}{c|}{\textit{dataset(*)}} &\textit{object} & \textit{symmetry} & \multicolumn{1}{c|}{\textit{dataset(*)}} \\\hline
Euler & 16.1 & 8.50 & 7.80 & 39.5 & 39.3 & 44.1 & 54.1 & 47.1 & 49.7\\
Rotation-Matrix & 6.80 & 4.40 & 3.30 & 14.7 & 28.5 & 27.6 & 46.1 & 46.8 & 41.7 \\
6d & 5.70 & 7.40 & 1.10 & 16.9 & 27.8 &  19.6 & 35.1 & 38.1 & 34.5 \\
Quaternion & 7.80 & 9.10 & 1.40 & 41.9 & 51.0 & 21.0 & 52.3 & 52.6 & 30.6 \\
Trigonometric & 9.90 & 6.90 & 10.5 & 42.4 & 43.2 & 43.1 & 56.7 & 49.0 & 58.1 \\
\hline
$\restr{\text{Euler}}{C}$ & 32.2 & 30.3 & 22.5 & 39.9 & 48.3 & 46.9 & 56.1 & 53.9 & 53.5 \\
$\restr{\text{Rotation-Matrix}}{C}$ & \textit{38.8} & \textit{41.3} & 24.2 & 42.5 & \textit{54.0} & 46.1 & \underline{64.1} & \textit{63.7} & 51.3 \\
$\restr{\text{6d}}{C}$ & 34.8 & 40.7 & \textit{41.5} & 41.6 & 48.0 & \textit{51.1} & 58.4 & 63.3 & \textit{65.0} \\
$\restr{\text{Quaternion}}{C}$ & 33.7 & 38.0 & 28.5 & \underline{47.4} & 53.3 & \textbf{57.6} & 60.2 & 61.6 & 56.8 \\
$\restr{\text{Trigonometric}}{C}$ & \underline{40.0} & \underline{43.7} & 41.1 & \textit{46.7} & \underline{54.5} & 49.4 & \textit{63.9} & \underline{66.3} & 64.1 \\
\hline
\textbf{SARR}--Depth  & \textbf{45.0} & \textbf{47.5} & \textbf{48.0}(\underline{46.7}) &  \textbf{48.0} & \textbf{56.3} & 50.3(\underline{54.9})  & \textbf{68.4} & \textbf{70.0} & \textbf{69.8}(\underline{68.8}) \\
\hline
\hline
\multirow{2}{*}{\textbf{Representation}} & \multicolumn{3}{c|}{\textbf{ViVo} - AR\textsubscript{C}} & \multicolumn{3}{c|}{\textbf{ViVo} - AR\textsubscript{B}}  & \multicolumn{3}{c|}{\textbf{ViVo} - AR\textsubscript{G}} \\
& \textit{object} & \textit{symmetry} & \multicolumn{1}{c|}{\textit{dataset(*)}} & \textit{object} & \textit{symmetry} & \multicolumn{1}{c|}{\textit{dataset(*)}} &\textit{object} & \textit{symmetry} & \multicolumn{1}{c|}{\textit{dataset(*)}} \\\hline
Euler & 15.3 & 8.00 & 7.70 & 44.4 & 37.7 & 41.3 & 54.9 & 49.1 & 52.0 \\
Rotation-Matrix & 6.00 & 3.60 & 3.30 & 29.1 & 24.2 & 24.5 & 44.9 & 48.8 & 43.7 \\
6d & 5.30 & 7.10 & 1.10 & 26.1 & 27.5 & 23.0 & 38.3 & 40.3 & 37.6 \\
Quaternion & 6.90 & 8.30 & 1.40 & 38.9 & 41.1 & 23.2 & 52.3 & 53.2 & 35.3 \\
Trigonometric & 8.50 & 6.10 & 9.10 & 44.6 & 39.5 & 47.6 & 56.5 & 49.8 &57.4 \\
\hline
$\restr{\text{Euler}}{C}$ & 30.5 & 29.7 & 20.9 & 46.2 & 45.3 & 44.0 & 56.9 & 56.4 & 54.5 \\
$\restr{\text{Rotation-Matrix}}{C}$ & \textit{36.3} & \textit{38.7} & 22.2 & \textit{52.1} & \textit{54.0} & 40.6 & \underline{63.7} & \textit{64.1} & 52.9 \\
$\restr{\text{6d}}{C}$ & 32.3 & 38.3 & \textit{38.6} & 47.4 & 53.7 & \textit{55.5} & 56.7 & 63.9 & \textit{64.2}\\
$\restr{\text{Quaternion}}{C}$ & 31.1 & 36.2 & 25.9 & 47.6 & 51.5 & 46.1 & 59.6 & 62.6 & 57.2 \\
$\restr{\text{Trigonometric}}{C}$ & \underline{37.2} & \underline{41.0} & 38.3 & \underline{52.9} & \underline{56.4} & 54.9 & \textit{62.5} & \underline{66.6} & 63.4 \\
\hline
\textbf{SARR}--Depth & \textbf{41.9} & \textbf{44.0} & \textbf{44.8}(\underline{42.9}) & \textbf{58.1} & \textbf{60.1} & \textbf{60.0}(\underline{59.0})  & \textbf{67.0} & \textbf{69.4} & \textbf{69.4}(\underline{66.7}) \\
\hline
\end{tabular}
\end{threeparttable}%
}
\end{table*}
 We compare to some of the best performing methods from the BOP T-LESS leaderboard in \textbf{Table}~\ref{tab_tless_other_results}. This includes not only RGB-D based methods, which have historically performed the best, but also RGB-only methods, which have become increasingly better in recent times, as well as depth-only methods. Under metric AR\textsubscript{B}, SARR--Depth networks achieve satisfactory results, especially considering that we use only a small number of real depth (D: R) images but no 3D models (see columns ``Training'' and ``Model'') and we did not place significant emphasis on using the latest network architecture. Our networks already outperform some existing methods under AR\textsubscript{G} and finally outperform all other methods under the symmetry-sensitive metric AR\textsubscript{C}. The SARR--Depth-\textit{dataset*} network, trained with the additional task of symmetry classification, performed only marginally worse in terms of orientation estimation than the SARR--Depth-\textit{dataset} network, while achieving a mean classification accuracy of 78.6\% for the SiSo task, and 77.8\% for ViVo (see \textbf{Appendix}~\ref{app_D}, specifically \textbf{Fig.}~\ref{fig_cfm_a} to \textbf{Fig.}~\ref{fig_cfm_d}). As can be seen in column `Symm.', besides \citet{drost2010} and \citet{vidal2018}, our method is the only one to augment the numeric rotation representation itself to deal with symmetry ambiguities\footnote{Vidal et al. actually used the representation proposed by Drost et al. to handle the symmetries.}. Other methods rely on the evaluation metrics, network/loss modifications or additional data. The last four rows show the 3D object orientation estimation results obtained using RGB images instead. Performance decreases notably compared to SARR--Depth for the \textit{object} and \textit{symmetry} scope networks, but for the \textit{dataset(*)} networks, trained on a much larger single training set, performance stays comparable. See \textbf{Fig.}~\ref{fig_cfm_rgb_a} through \textbf{Fig.}~\ref{fig_cfm_rgb_d} for symmetry classification confusion matrices of the SARR--RGB-\textit{dataset*} network.\\
To show the merit of using SARR for rotation estimation of symmetric objects in comparison to other representations, we trained 30 additional networks using depth as input modality: One for each of the five other rotation representations (Euler angles, rotation matrices, the 6d representation from \citep{zhou2019a}, quaternions\footnote{For quaternions we map the redundant double cover to the ``canonic'' single cover, see \url{https://docs.scipy.org/doc/scipy/reference/generated/scipy.spatial.transform.Rotation.as_quat.html} (Accessed: 2026-01-12).} and trigonometrics) and for each of the three scopes, with annotations derived from either the ambiguous T-LESS rotation matrices or our canonic Euler angles $\restr{\cdot}{C}$. \textbf{Table}~\ref{tab_tless_rep_results} shows these results for both tasks, SiSo on the top and ViVo on the bottom. With the exception of the AR\textsubscript{B} score of $\restr{\text{Quaternion}}{C}$-\textit{dataset} on SiSo, networks trained on the SARR representation outperform all other representations across the different scopes, evaluation metrics and tasks. The table highlights the merit of paying respect to object symmetries, as all networks trained on unrestricted representations (upper halves in the SiSO and ViVO blocks) performed poorly, especially under the metric AR\textsubscript{C} but also using the symmetry-invariant metrics AR\textsubscript{B} and AR\textsubscript{G}. 

\subsection{ITODD}\label{subsec_exp_itodd}
\begin{table*}[ht]
\centering
\resizebox{\linewidth}{!}{%
\begin{threeparttable}[b]
\caption{\textbf{ITODD orientation estimation.} SARR--Depth-networks outperform SC6D under the strict AR\textsubscript{C} metric, the \textit{dataset*}-network gives the best results. Changing the input to grayscale substantially decreases performance.}
\label{tab_itodd_other_results}
\begin{tabular}{|c|ccccc|cc|cc|c|}
\hline
\multirow{2}{*}{\textbf{Method}} & \multirow{2}{*}{\textbf{Training}} & \multirow{2}{*}{\textbf{Test}}  &\multirow{2}{*}{\textbf{Scope}} & \multirow{2}{*}{\textbf{Model}} & \multirow{2}{*}{\textbf{Symmetry}}  & \multicolumn{2}{c|}{\textbf{SiSo}} & \multicolumn{2}{c|}{\textbf{ViVo}} & \multirow{2}{*}{t[s]}\\
 & & &  & & & \multicolumn{1}{c}{AR\textsubscript{C}} & \multicolumn{1}{c|}{AR\textsubscript{B}} & \multicolumn{1}{c}{AR\textsubscript{C}} & \multicolumn{1}{c|}{AR\textsubscript{B}} & \\
\hline
\cite{cai2022} & RGB: S & RGB &dataset & \ding{55} & network & 36.7 & \textbf{74.0} & 30.6 & \textbf{82.0} & .053 \\
\hline
\multirow{4}{*}{\parbox{2.3cm}{\textbf{SARR}--Depth}} & D: S & D & \textit{object} & \ding{55} & representation & 38.3 & 62.1 & 42.1 & \textit{67.7} & .071 \\
& D: S & D & \textit{symmetry} & \ding{55} & representation & \underline{41.0} & \underline{68.7} & \underline{43.0} & 63.7 & .079 \\
& D: S & D & \textit{dataset} & \ding{55} & representation & \underline{41.0} & \textit{63.7} & \textit{42.1} & 65.1 & .066 \\
& D: S & D & \textit{dataset*} & \ding{55} & representation & \textbf{44.8} & 56.3 & \textbf{48.1} & \underline{71.1} & .078 \\
\hline
\multirow{4}{*}{\parbox{2.1cm}{\textbf{SARR}--Gray}} & RGB2Gray: S & Gray & \textit{object} & \ding{55} & representation & 10.8 & 23.6 & 14.2 & 33.7 & .080 \\
& RGB2Gray: S & Gray & \textit{symmetry} & \ding{55} & representation & 7.1 & 24.0 & 8.1 & 28.7 & .088 \\
& RGB2Gray: S & Gray & \textit{dataset} & \ding{55} & representation & 2.8 & 6.9 & 4.6 & 15.4 & .073 \\
& RGB2Gray: S & Gray & \textit{dataset*} & \ding{55} & representation & 6.5 & 22.5 & 6.6 & 24.2 & .082 \\
\hline
\end{tabular}
\end{threeparttable}
}
\end{table*}
\begin{table*}[ht]
\centering
\resizebox{\linewidth}{!}{%
\begin{threeparttable}[b]
\caption{\textbf{ITODD representation comparison.} The SARR representation outperforms most other representations.}
\label{tab_itodd_rep_results}
\renewcommand{\arraystretch}{1.0} 
\setlength{\tabcolsep}{0.8mm} 
\begin{tabular}{|c|ccc|ccc|ccc|ccc|}
\hline
\multirow{2}{*}{\textbf{Representation}} & \multicolumn{3}{c|}{\textbf{SiSo} - AR\textsubscript{C}} & \multicolumn{3}{c|}{\textbf{SiSo} - AR\textsubscript{B}} & \multicolumn{3}{c|}{\textbf{ViVo} - AR\textsubscript{C}} & \multicolumn{3}{c|}{\textbf{ViVo} - AR\textsubscript{B}}\\
& \textit{obj.} & \textit{symm.} & \multicolumn{1}{c|}{\textit{dataset(*)}} & \textit{obj.} & \textit{symm.} & \multicolumn{1}{c|}{\textit{dataset(*)}} & \textit{obj.} & \textit{symm.} & \multicolumn{1}{c|}{\textit{dataset(*)}} & \textit{obj.} & \textit{symm.} & \multicolumn{1}{c|}{\textit{dataset(*)}} \\
\hline
Euler & 17.9 & 8.6 & 9.6 & 51.0 & 40.2 & 32.5 & 20.7 & 10.4 & 10.8 & 52.4 & 37.9 & 32.8 \\ 
Rot.-Matrix & 11.7 & 4.3 & 0.0 & 30.2 & 15.5 & 15.6 & 12.1 & 4.6 & 0.1 & 35.4 & 22.6 & 12.5 \\  
6d & 8.6 & 0.3 & 1.2 & 24.4 & 14.8 & 14.9 & 9.6 & 1.1 & 2.2 & 38.4 & 15.3 & 18.9 \\ 
Quaternion & 13.3 & 1.2 & 0.9 & 27.0 & 29.6 & 17.0 & 10.6 & 1.9 & 0.9 & 45.1 & 34.7 & 23.9 \\ 
Trigonometric & 19.4 & 20.5 & 6.8 & 49.5 & 51.7 & 38.3 & 19.2 & 8.8 & 5.7 & \textit{60.0} & 53.1 & 42.7 \\ 
\hline
$\restr{\text{Euler}}{C}$ & 27.8 & 21.6 & 9.0 & \textbf{68.5} & 51.0 & 39.2 & 29.8 & 22.9 & 10.0 & 58.3 & 43.8 & 34.6 \\ 
$\restr{\text{Rot.-Matrix}}{C}$ & 35.5 & \textit{31.2} & \textit{34.6} & 58.8 & \textit{52.2} & \textit{59.6} & \textit{34.4} & \textit{34.6} & \textit{35.5} & 55.4 & \textit{52.9} & \textit{61.3} \\ 
$\restr{\text{6d}}{C}$ & \underline{38.9} & 23.1 & 18.5 & 56.0 & 42.6 & 46.9 & 33.9 & 22.8 & 19.2 & 54.5 & 41.0 & 41.3 \\ 
$\restr{\text{Quaternion}}{C}$ & 28.7 & 24.7 & 11.7 & 44.4 & 47.1 & 32.9 & 26.8 & 26.4 & 11.8 & 48.9 & 44.5 & 31.1 \\ 
$\restr{\text{Trigonometric}}{C}$ & \textbf{40.1} & \underline{39.2} & 34.0 & \textit{61.2} & \underline{60.8} & \underline{61.4} & \underline{41.9} & \underline{39.3} & 34.6  & \underline{66.0} & \underline{62.7} & 60.7 \\ 
\hline
\textbf{SARR}--Depth & \textit{38.3} & \textbf{41.0} & \underline{41.0}(\textbf{44.8}) & \underline{62.1} & \textbf{68.7} & \textbf{63.7}(56.3) & \textbf{42.1} & \textbf{43.0} & \underline{42.1}(\textbf{48.1}) & \textbf{67.7} & \textbf{63.7} & \underline{65.1}(\textbf{71.1})\\
\hline
\end{tabular}
\end{threeparttable}%
}
\end{table*}
\begin{figure*}[!ht]
    \centering
    \begin{subfigure}[b]{1.0\linewidth}
        \centering
        \includegraphics[width=\linewidth]{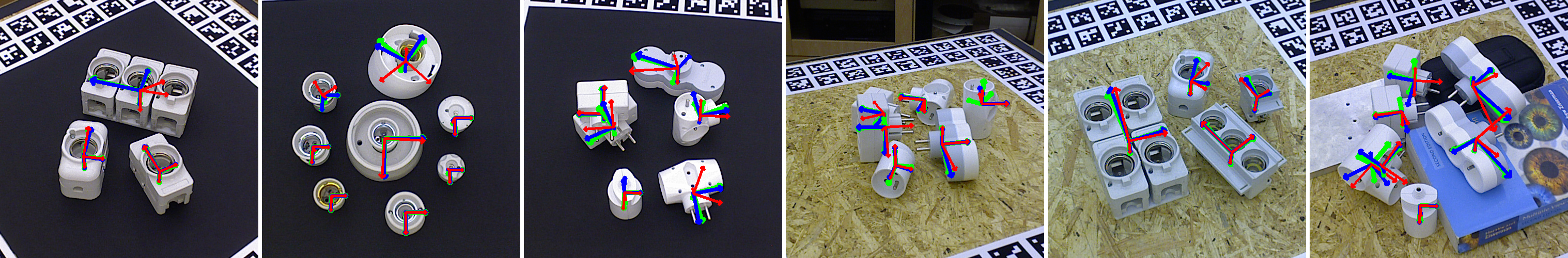}
        \caption{\textbf{T-LESS visualizations.} \textcolor{blue}{SARR--Depth}-\textit{dataset} predictions in blue, \textcolor{red}{$\restr{\text{Trigonometric}}{C}$-\textit{dataset}} in red.}
        \label{fig_vis_tless}
    \end{subfigure}
    \begin{subfigure}[b]{1.0\linewidth}
        \centering
        \includegraphics[width=\linewidth]{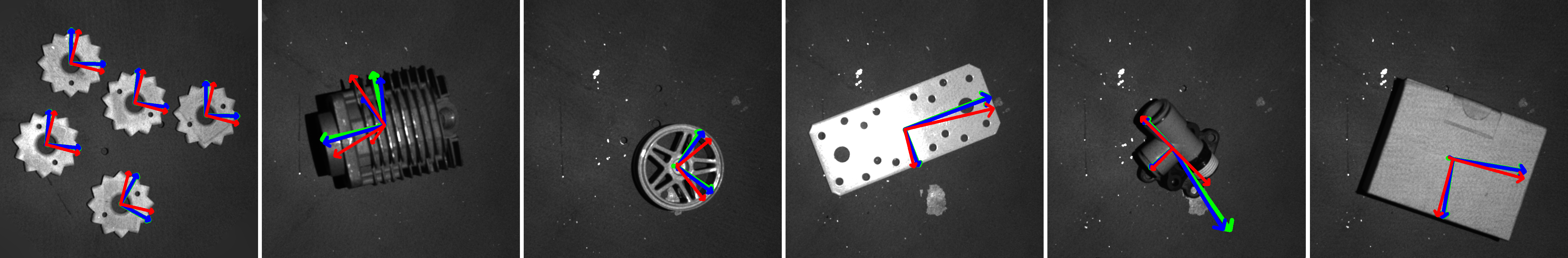}
        \caption{\textbf{ITODD visualizations.} \textcolor{blue}{SARR--Depth}-\textit{symmetry} predictions in blue, \textcolor{red}{$\restr{\text{Trigonometric}}{C}$-\textit{object}} in red.}
        \label{fig_vis_itodd}
    \end{subfigure}
    \caption{\textbf{T-LESS and ITODD pose predictions.} \textcolor{green}{Ground-truths} are given in green.}
    \label{fig_pred_results}
\end{figure*}
While the ITODD BOP leaderboard is well populated with different methods, public results are only available for the test set, not the validation set. This means that we have to obtain pose estimations, and calculate results, for other methods ourselves, which in turn requires a maintained, publicly available framework. We additionally had to exclude all methods from consideration that used the real validation images for fine-tuning (which is a common practice), since the validation set constitutes our test set. With this in mind, we were able to set-up and deploy the SC6D method by \citet{cai2022} for comparison. Their method also puts emphasis on object symmetries and does not use the 3D CAD models, like ours. \textbf{Table}~\ref{tab_itodd_other_results} displays results for this comparison. While SC6D outperforms our networks under AR\textsubscript{B}, using our SARR representation and depth images leads to better results following the stricter AR\textsubscript{C} metric. Interestingly, the SARR--Depth-\textit{dataset*} network, which achieved symmetry classification accuracies of 91.0\% for SiSo and 94.2\% for ViVo (see \textbf{Appendix}~\ref{app_D}, specifically \textbf{Fig.}~\ref{fig_cfm_e} to \textbf{Fig.}~\ref{fig_cfm_h}), also performed the best in terms of orientation estimation. It seems that the additional task of learning to classify the symmetry positively impacted the rotation estimation optimization. We hypothesize that the symmetry class labels may act as guidance for the optimizer to more easily find and distinguish the respective areas of the loss surface, since these areas are topologically quite different from one another when using the SARR representation. Additional experiments with the SARR--Gray networks yielded poor results, most likely because the synthetic grayscale training images converted from RGB exhibit substantial difference in appearance compared to the real grayscale evaluation images.\\
We compare against other rotation representations in \textbf{Table}~\ref{tab_itodd_rep_results}. Using our representation again yields networks that provide the best results in almost all comparisons. When training the $\restr{\text{Quaternion}}{C}$-\textit{dataset} network, we observed a NaN error that occurred during backpropagation in epoch 15 out of 40. This effectively halted the optimization at that point and might be an indicator of poor stability of the Quaternion representation for this learning task. 

\subsection{Results Summary}\label{subsec_exp_summary}
\begin{table}[ht]
\centering
\caption{\textbf{Summary of results.}}
\label{tab_summary_results}
\renewcommand{\arraystretch}{1.0} 
\setlength{\tabcolsep}{1.2mm} 
\begin{tabular}{|cc|ccc|cc|}
\hline
\multicolumn{2}{|c|}{\multirow{2}{*}{\textbf{Representation}}} & \multicolumn{3}{c|}{\textbf{T-LESS}} & \multicolumn{2}{c|}{\textbf{ITODD}} \\
& & AR\textsubscript{C} & AR\textsubscript{B} & AR\textsubscript{G} & AR\textsubscript{C} & AR\textsubscript{B} \\
\hline
\multicolumn{2}{|c|}{Euler} & 10.6 & 41.1 & 51.2 & 13.0 & 41.1 \\
\multicolumn{2}{|c|}{Rotation-Matrix} & 4.57 & 24.8 & 45.3 & 5.47 & 22.0 \\
\multicolumn{2}{|c|}{6d} & 4.62 & 23.5 & 37.3 & 3.83 & 21.1\\
\multicolumn{2}{|c|}{Quaternion} & 5.82 & 36.2 & 46.1 & 4.80 & 29.6 \\
\multicolumn{2}{|c|}{Trigonometric} & 8.50 & 43.4 & 54.6 & 13.4 & 49.2 \\
\hline
\multicolumn{2}{|c|}{$\restr{\text{Euler}}{C}$} & 27.7 & 45.1 & 55.2 & 20.2 & 49.2 \\
\multicolumn{2}{|c|}{$\restr{\text{Rotation-Matrix}}{C}$} & 33.6 & 48.2 & 60.0 & 34.3 & 56.7 \\
\multicolumn{2}{|c|}{$\restr{\text{6d}}{C}$} & 37.7 & 49.6 & 61.9 & 26.1 & 47.1 \\
\multicolumn{2}{|c|}{$\restr{\text{Quaternion}}{C}$} & 32.2 & 50.6 & 59.7 & 21.7 & 41.5 \\
\multicolumn{2}{|c|}{$\restr{t\text{Trigonometric}}{C}$} & \textit{40.2} & \textit{52.5} & \textit{64.5} & \textit{38.2} & \textit{62.1} \\
\hline
\multirow{2}{*}{\textbf{SARR}} & /w \textit{dataset} & \textbf{45.2} & \underline{55.5} & \textbf{69.0} & \underline{41.3} & \textbf{65.2} \\ 
& /w \textit{dataset*} & \underline{44.7} & \textbf{56.1} & \underline{68.4} & \textbf{42.9} & \underline{64.9}\\
\hline
\end{tabular}
\end{table}
By averaging the representation comparison results from \textbf{Table}~\ref{tab_tless_rep_results} and \textbf{Table}~\ref{tab_itodd_rep_results} across the three scopes and two tasks, we summarize our results in \textbf{Table}~\ref{tab_summary_results}, which leads to four key observations. Firstly, using our proposed SARR representation leads to an absolute performance increase over the second best representation $\restr{\text{Trigonometric}}{C}$ of about $3-5\%$, or a relative increase of up to $11\%$. Secondly, the representation generalizes well across two diverse datasets, with T-LESS and ITODD sharing only a couple of symmetry classes. Thirdly, comparing the last two rows shows that the additional task of classifying the object symmetry (\textit{dataset*}) is barely detrimental and sometimes even conducive to orientation estimation, which is relevant for scenarios where the symmetry class is not known a priori. Finally, relying solely on the symmetry-agnostic nature of an evaluation metric and using standard rotation representations leads to poor performance due to the ambiguities during optimization (upper half).\\
Lastly, \textbf{Fig.}~\ref{fig_pred_results} shows visualizations of predicted orientations from networks trained on depth images and with different rotation representations, featuring various objects from both T-LESS and ITODD. While in some cases the other representations yield clearly wrong estimates, the advantages of the SARR networks are well visible, especially in the difficult T-LESS images (textured background in \textbf{Fig.}~\ref{fig_vis_tless}) and the first ITODD image (\textbf{Fig.}~\ref{fig_vis_itodd}).\\

\section{Conclusion}\label{sec_conclusion}
In this work, we focused on the challenges that arise when training a ML method to estimate the 3D orientation of symmetric objects. We focused on the symmetry classes in the popular T-LESS and ITODD datasets and proposed a representation which builds on trigonometric identities that resolves symmetry ambiguities to a canonic pose, while preserving continuity across symmetry boundaries. Our method operates on the annotations directly, allowing any pose estimation network to take into account the symmetries of objects, solving the problem of symmetries more naturally than existing works. The proposed representation scheme is generic and could be extended to additional symmetry classes beyond the ones analysed in this paper. Therefore, future work includes the analysis of other object symmetries in different datasets, further extending the rotation space, and incorporating translation prediction to allow full 6D pose estimation. Disentangling the symmetry classification and pose regression tasks could also prove beneficial. Fusing depth and RGB information while remaining texture-agnostic in our symmetry definitions could open new avenues as well. Finally, we expect a performance increase from more powerful network architectures such as Transformers.

\backmatter
\section*{Statements and Declarations}
\bmhead{Supplementary information} The online version of this paper contains supplementary material available at (link).
\bmhead{Author Contributions} All authors contributed to the conception of the paper, the formulation of the method, and the interpretation of the results. The concept of the rotation representation, the implementation and the experiments were developed by Andreas Kriegler. The first draft of the manuscript was written by Andreas Kriegler, and all authors continuously revised the manuscript. All authors read and approved the final manuscript.
\bmhead{Funding} This work was funded by the Lighthouse Project AI-Enabled Sustainable Automation and Robotics of the Austrian Institute of Technology (AIT).
\bmhead{Data Availability} All experiments make use of the T-LESS dataset (version 2), publicly available at \url{https://huggingface.co/datasets/bop-benchmark/tless} (Accessed: 2026-01-12). Predictions used to calculate the results from Tables 3, 4, 5 and 6 (other methods and ours) are available at \url{https://github.com/akriegler/SARR/tree/main/results} (Accessed: 2026-01-12).
\bmhead{Competing Interests} The authors have no interests to declare that had an influence on the content of this article. 
\bmhead{Code Availability} Code has been released on GitHub alongside this article and is available at \url{https://github.com/akriegler/SARR} (Accessed: 2026-01-12).
\bmhead{Materials Availability} Not applicable.
\bmhead{Ethics Approval} Not applicable.
\bmhead{Consent for Publication} Not applicable.
\bmhead{Consent for Participation} Not applicable.
\bigskip
\bibliography{literature}

@inproceedings{ayoub2023,
  title = {Grasp {{Planning}} with {{CNN}} for {{Log-loading Forestry Machine}}},
  booktitle = {2023 {{IEEE International Conference}} on {{Robotics}} and {{Automation}} ({{ICRA}})},
  author = {Ayoub, Elie and Levesque, Patrick and Sharf, Inna},
  year = 2023,
  month = may,
  pages = {11802--11808},
  publisher = {IEEE},
  address = {London, United Kingdom},
  doi = {10.1109/ICRA48891.2023.10161562},
  urldate = {2024-08-22},
  copyright = {https://doi.org/10.15223/policy-029},
  isbn = {979-8-3503-2365-8}
}

@inproceedings{banerjee2025,
  title = {{{HOT3D}}: {{Hand}} and {{Object Tracking}} in {{3D}} from {{Egocentric Multi-View Videos}}},
  booktitle = {2025 {{IEEE}}/{{CVF Conference}} on {{Computer Vision}} and {{Pattern Recognition}} ({{CVPR}})},
  author = {Banerjee, Prithviraj and Shkodrani, Sindi and Moulon, Pierre and Hampali, Shreyas and Han, Shangchen and Zhang, Fan and Zhang, Linguang and Fountain, Jade and Miller, Edward and Basol, Selen and Newcombe, Richard and Wang, Robert and Engel, Jakob Julian and Hodan, Tomas},
  year = 2025,
  pages = {7061--7071},
  publisher = {IEEE},
  address = {Nashville, TN, USA},
  doi = {10.1109/CVPR52734.2025.00662},
  langid = {english}
}

@inproceedings{brachmann2014,
  title = {Learning {{6D Object Pose Estimation Using 3D Object Coordinates}}},
  booktitle = {European {{Conference}} on {{Computer Vision}} ({{ECCV}})},
  author = {Brachmann, Eric and Krull, Alexander and Michel, Frank and Gumhold, Stefan and Shotton, Jamie and Rother, Carsten},
  year = 2014,
  volume = {8690},
  pages = {536--551},
  publisher = {Springer International Publishing},
  address = {Zurich, Switzerland},
  doi = {10.1007/978-3-319-10605-2_35},
  urldate = {2025-07-01},
  copyright = {http://www.springer.com/tdm},
  isbn = {978-3-319-10604-5 978-3-319-10605-2},
  langid = {english}
}

@inproceedings{bregier2017,
  title = {Symmetry {{Aware Evaluation}} of {{3D Object Detection}} and {{Pose Estimation}} in {{Scenes}} of {{Many Parts}} in {{Bulk}}},
  booktitle = {2017 {{IEEE International Conference}} on {{Computer Vision Workshops}} ({{ICCVW}})},
  author = {Bregier, Romain and Devernay, Frederic and Leyrit, Laetitia and Crowley, James L.},
  year = 2017,
  month = oct,
  pages = {2209--2218},
  publisher = {IEEE},
  address = {Venice, Italy},
  doi = {10.1109/ICCVW.2017.258},
  urldate = {2024-08-22},
  isbn = {978-1-5386-1034-3}
}

@article{bregier2018,
  title = {Defining the {{Pose}} of {{Any 3D Rigid Object}} and an {{Associated Distance}}},
  author = {Br{\'e}gier, Romain and Devernay, Fr{\'e}d{\'e}ric and Leyrit, Laetitia and Crowley, James L.},
  year = 2018,
  month = jun,
  journal = {International Journal of Computer Vision (IJCV)},
  volume = {126},
  number = {6},
  pages = {571--596},
  issn = {0920-5691, 1573-1405},
  doi = {10.1007/s11263-017-1052-4},
  urldate = {2024-08-22},
  langid = {english}
}

@inproceedings{cai2022,
  title = {{{SC6D}}: {{Symmetry-agnostic}} and {{Correspondence-free 6D Object Pose Estimation}}},
  shorttitle = {{{SC6D}}},
  booktitle = {2022 {{International Conference}} on {{3D Vision}} ({{3DV}})},
  author = {Cai, Dingding and Heikkil{\"a}, Janne and Rahtu, Esa},
  year = 2022,
  month = sep,
  pages = {536--546},
  publisher = {IEEE},
  address = {Prague, Czech Republic},
  doi = {10.1109/3DV57658.2022.00065},
  urldate = {2024-08-22},
  copyright = {https://doi.org/10.15223/policy-029},
  isbn = {978-1-6654-5670-8}
}

@article{calli2017,
  title = {Yale-{{CMU-Berkeley}} Dataset for Robotic Manipulation Research},
  author = {Calli, Berk and Singh, Arjun and Bruce, James and Walsman, Aaron and Konolige, Kurt and Srinivasa, Siddhartha and Abbeel, Pieter and Dollar, Aaron M.},
  year = 2017,
  journal = {International Journal of Robotics Research},
  volume = {36},
  number = {3},
  pages = {261--268},
  issn = {17413176},
  doi = {10.1177/0278364917700714}
}

@inproceedings{castro2023,
  title = {{{CRT-6D}}: {{Fast 6D Object Pose Estimation}} with {{Cascaded Refinement Transformers}}},
  shorttitle = {{{CRT-6D}}},
  booktitle = {2023 {{IEEE}}/{{CVF Winter Conference}} on {{Applications}} of {{Computer Vision}} ({{WACV}})},
  author = {Castro, Pedro and Kim, Tae-Kyun},
  year = 2023,
  month = jan,
  pages = {5735--5744},
  publisher = {IEEE},
  address = {Waikoloa, HI, USA},
  doi = {10.1109/WACV56688.2023.00570},
  urldate = {2025-10-16},
  copyright = {https://doi.org/10.15223/policy-029},
  isbn = {978-1-6654-9346-8}
}

@inproceedings{chao2019,
  title = {{{HarDNet}}: {{A Low Memory Traffic Network}}},
  shorttitle = {{{HarDNet}}},
  booktitle = {2019 {{IEEE}}/{{CVF International Conference}} on {{Computer Vision}} ({{ICCV}})},
  author = {Chao, Ping and Kao, Chao-Yang and Ruan, Yushan and Huang, Chien-Hsiang and Lin, Youn-Long},
  year = 2019,
  month = oct,
  pages = {3551--3560},
  publisher = {IEEE},
  address = {Seoul, Korea (South)},
  issn = {2380-7504},
  doi = {10.1109/ICCV.2019.00365},
  urldate = {2024-08-23}
}

@inproceedings{chen2021,
  title = {{{FS-Net}}: {{Fast Shape-based Network}} for {{Category-Level 6D Object Pose Estimation}} with {{Decoupled Rotation Mechanism}}},
  booktitle = {2021 {{IEEE}}/{{CVF Conference}} on {{Computer Vision}} and {{Pattern Recognition}} ({{CVPR}})},
  author = {Chen, Wei and Jia, Xi and Chang, Hyung Jin and Duan, Jinming and Shen, Linlin and Leonardis, Ale{\v s}},
  year = 2021,
  eprint = {2103.07054v1},
  pages = {1581--1590},
  publisher = {IEEE},
  address = {Nasville, TN, USA},
  doi = {10.1109/CVPR46437.2021.00163},
  archiveprefix = {arXiv}
}

@inproceedings{chen2023,
  title = {{{StereoPose}}: {{Category-Level 6D Transparent Object Pose Estimation}} from {{Stereo Images}} via {{Back-View NOCS}}},
  shorttitle = {{{StereoPose}}},
  booktitle = {2023 {{IEEE International Conference}} on {{Robotics}} and {{Automation}} ({{ICRA}})},
  author = {Chen, Kai and James, Stephen and Sui, Congying and Liu, Yun-Hui and Abbeel, Pieter and Dou, Qi},
  year = 2023,
  month = may,
  pages = {2855--2861},
  publisher = {IEEE},
  address = {London, United Kingdom},
  doi = {10.1109/ICRA48891.2023.10160780},
  urldate = {2024-08-22},
  copyright = {https://doi.org/10.15223/policy-029},
  isbn = {979-8-3503-2365-8}
}

@inproceedings{corona2018,
  title = {Pose {{Estimation}} for {{Objects}} with {{Rotational Symmetry}}},
  booktitle = {2018 {{IEEE}}/{{RSJ International Conference}} on {{Intelligent Robots}} and {{Systems}} ({{IROS}})},
  author = {Corona, Enric and Kundu, Kaustav and Fidler, Sanja},
  year = 2018,
  month = oct,
  pages = {7215--7222},
  publisher = {IEEE},
  address = {Madrid, Spain},
  doi = {10.1109/IROS.2018.8594282},
  urldate = {2024-08-22},
  isbn = {978-1-5386-8094-0}
}

@inproceedings{doumanoglou2016,
  title = {Recovering {{6D Object Pose}} and {{Predicting Next-Best-View}} in the {{Crowd}}},
  booktitle = {2016 {{IEEE Conference}} on {{Computer Vision}} and {{Pattern Recognition}} ({{CVPR}})},
  author = {Doumanoglou, Andreas and Kouskouridas, Rigas and Malassiotis, Sotiris and Kim, Tae-Kyun},
  year = 2016,
  month = jun,
  pages = {3583--3592},
  publisher = {IEEE},
  address = {Las Vegas, NV, USA},
  doi = {10.1109/CVPR.2016.390},
  urldate = {2025-07-01},
  isbn = {978-1-4673-8851-1},
  langid = {english}
}

@inproceedings{drost2010,
  title = {Model Globally, Match Locally: {{Efficient}} and Robust {{3D}} Object Recognition},
  shorttitle = {Model Globally, Match Locally},
  booktitle = {2010 {{IEEE Computer Society Conference}} on {{Computer Vision}} and {{Pattern Recognition}} ({{CVPR}})},
  author = {Drost, Bertram and Ulrich, Markus and Navab, Nassir and Ilic, Slobodan},
  year = 2010,
  month = jun,
  pages = {998--1005},
  publisher = {IEEE},
  address = {San Francisco, CA, USA},
  doi = {10.1109/CVPR.2010.5540108},
  urldate = {2024-08-22},
  isbn = {978-1-4244-6984-0}
}

@inproceedings{drost2017,
  title = {Introducing {{MVTec ITODD}} --- {{A Dataset}} for {{3D Object Recognition}} in {{Industry}}},
  booktitle = {2017 {{IEEE International Conference}} on {{Computer Vision Workshops}} ({{ICCVW}})},
  author = {Drost, Bertram and Ulrich, Markus and Bergmann, Paul and Hartinger, Philipp and Steger, Carsten},
  year = 2017,
  pages = {2200--2208},
  publisher = {IEEE},
  address = {Venice, Italy},
  doi = {10.1109/ICCVW.2017.257},
  urldate = {2025-01-28},
  isbn = {978-1-5386-1034-3}
}

@inproceedings{guo2023,
  title = {{{HANDAL}}: {{A Dataset}} of {{Real-World Manipulable Object Categories}} with {{Pose Annotations}}, {{Affordances}}, and {{Reconstructions}}},
  shorttitle = {{{HANDAL}}},
  booktitle = {2023 {{IEEE}}/{{RSJ International Conference}} on {{Intelligent Robots}} and {{Systems}} ({{IROS}})},
  author = {Guo, Andrew and Wen, Bowen and Yuan, Jianhe and Tremblay, Jonathan and Tyree, Stephen and Smith, Jeffrey and Birchfield, Stan},
  year = 2023,
  month = oct,
  pages = {11428--11435},
  publisher = {IEEE},
  address = {Detroit, MI, USA},
  doi = {10.1109/IROS55552.2023.10341672},
  urldate = {2025-07-01},
  copyright = {https://doi.org/10.15223/policy-029},
  isbn = {978-1-6654-9190-7}
}

@misc{hara2017,
  title = {Designing {{Deep Convolutional Neural Networks}} for {{Continuous Object Orientation Estimation}}},
  author = {Hara, Kota and Vemulapalli, Raviteja and Chellappa, Rama},
  year = 2017,
  month = feb,
  number = {arXiv:1702.01499},
  eprint = {1702.01499},
  primaryclass = {cs},
  publisher = {arXiv},
  doi = {10.48550/arXiv.1702.01499},
  urldate = {2024-08-22},
  archiveprefix = {arXiv}
}

@inproceedings{haugaard2023,
  title = {{{SpyroPose}}: {{SE}}(3) {{Pyramids}} for {{Object Pose Distribution Estimation}}},
  shorttitle = {{{SpyroPose}}},
  booktitle = {2023 {{IEEE}}/{{CVF International Conference}} on {{Computer Vision Workshops}} ({{ICCVW}})},
  author = {Haugaard, Rasmus Laurvig and Hagelskj{\ae}r, Frederik and Iversen, Thorbj{\o}rn Mosekj{\ae}r},
  year = 2023,
  month = oct,
  pages = {2074--2083},
  publisher = {IEEE},
  address = {Paris, France},
  doi = {10.1109/ICCVW60793.2023.00222},
  urldate = {2024-08-22},
  copyright = {https://doi.org/10.15223/policy-029},
  isbn = {979-8-3503-0744-3}
}

@inproceedings{he2022,
  title = {{{FS6D}}: {{Few-Shot 6D Pose Estimation}} of {{Novel Objects}}},
  shorttitle = {{{FS6D}}},
  booktitle = {2022 {{IEEE}}/{{CVF Conference}} on {{Computer Vision}} and {{Pattern Recognition}} ({{CVPR}})},
  author = {He, Yisheng and Wang, Yao and Fan, Haoqiang and Sun, Jian and Chen, Qifeng},
  year = 2022,
  month = jun,
  pages = {6804--6814},
  address = {New Orleans, LA, USA},
  issn = {2575-7075},
  doi = {10.1109/CVPR52688.2022.00669},
  urldate = {2024-08-23}
}

@inproceedings{hinterstoisser2013,
  title = {Model {{Based Training}}, {{Detection}} and {{Pose Estimation}} of {{Texture-Less 3D Objects}} in {{Heavily Cluttered Scenes}}},
  booktitle = {Computer {{Vision}} -- {{ACCV}} 2012},
  author = {Hinterstoisser, Stefan and Lepetit, Vincent and Ilic, Slobodan and Holzer, Stefan and Bradski, Gary and Konolige, Kurt and Navab, Nassir},
  editor = {Hutchison, David and Kanade, Takeo and Kittler, Josef and Kleinberg, Jon M. and Mattern, Friedemann and Mitchell, John C. and Naor, Moni and Nierstrasz, Oscar and Pandu Rangan, C. and Steffen, Bernhard and Sudan, Madhu and Terzopoulos, Demetri and Tygar, Doug and Vardi, Moshe Y. and Weikum, Gerhard and Lee, Kyoung Mu and Matsushita, Yasuyuki and Rehg, James M. and Hu, Zhanyi},
  year = 2013,
  volume = {7724},
  pages = {548--562},
  publisher = {Springer Berlin Heidelberg},
  address = {Berlin, Heidelberg},
  doi = {10.1007/978-3-642-37331-2_42},
  urldate = {2025-07-01},
  isbn = {978-3-642-37330-5 978-3-642-37331-2}
}

@inproceedings{hodan2015,
  title = {Detection and {{Fine 3D Pose Estimation}} of {{Texture-less Objects}} in {{RGB-D Images}}},
  booktitle = {2015 {{IEEE}}/{{RSJ International Conference}} on {{Intelligent Robots}} and {{Systems}} ({{IROS}})},
  author = {Hoda{\v n}, Tom{\'a}{\v s} and Zabulis, Xenophon and Lourakis, Manolis and Obdrzalek, Stepan and Matas, Jiri},
  year = 2015,
  month = sep,
  pages = {4421--4428},
  publisher = {IEEE},
  address = {Hamburg, Germany},
  doi = {10.1109/IROS.2015.7354005},
  urldate = {2024-10-02},
  isbn = {978-1-4799-9994-1},
  langid = {english}
}

@inproceedings{hodan2016,
  title = {On {{Evaluation}} of {{6D Object Pose Estimation}}},
  booktitle = {European {{Conference}} on {{Computer Vision Workshops}} ({{ECCVW}})},
  author = {Hoda{\v n}, Tom{\'a}{\v s} and Matas, Ji{\v r}{\'i} and Obdr{\v z}{\'a}lek, {\v S}t{\v e}p{\'a}n},
  year = 2016,
  pages = {606--619},
  publisher = {Springer International Publishing},
  address = {Amsterdam, Netherlands},
  doi = {10.1007/978-3-319-49409-8_52},
  urldate = {2024-08-22},
  isbn = {978-3-319-49408-1 978-3-319-49409-8},
  langid = {english}
}

@inproceedings{hodan2017,
  title = {T-{{LESS}}: {{An RGB-D Dataset}} for {{6D Pose Estimation}} of {{Texture-Less Objects}}},
  shorttitle = {T-{{LESS}}},
  booktitle = {2017 {{IEEE Winter Conference}} on {{Applications}} of {{Computer Vision}} ({{WACV}})},
  author = {Hoda{\v n}, Tom{\'a}{\v s} and Haluza, Pavel and Obdrzalek, Stepan and Matas, Jiri and Lourakis, Manolis and Zabulis, Xenophon},
  year = 2017,
  month = mar,
  pages = {880--888},
  publisher = {IEEE},
  address = {Santa Rosa, CA, USA},
  doi = {10.1109/WACV.2017.103},
  urldate = {2024-08-22},
  isbn = {978-1-5090-4822-9}
}

@inproceedings{hodan2018,
  title = {{{BOP}}: {{Benchmark}} for {{6D Object Pose Estimation}}},
  shorttitle = {{{BOP}}},
  booktitle = {European {{Conference}} on {{Computer Vision}} ({{ECCV}})},
  author = {Hoda{\v n}, Tom{\'a}{\v s} and Michel, Frank and Brachmann, Eric and Kehl, Wadim and Buch, Anders Glent and Kraft, Dirk and Drost, Bertram and Vidal, Joel and Ihrke, Stephan and Zabulis, Xenophon and Sahin, Caner and Manhardt, Fabian and Tombari, Federico and Kim, Tae-Kyun and Matas, Ji{\v r}{\'i} and Rother, Carsten},
  year = 2018,
  pages = {19--35},
  publisher = {Springer International Publishing},
  address = {Munich, Germany},
  doi = {10.1007/978-3-030-01249-6_2},
  urldate = {2024-08-22},
  langid = {english}
}

@inproceedings{hodan2020,
  title = {{{BOP Challenge}} 2020 on {{6D Object Localization}}},
  booktitle = {European {{Conference}} on {{Computer Vision Workshops}} ({{ECCVW}})},
  author = {Hoda{\v n}, Tom{\'a}{\v s} and Sundermeyer, Martin and Drost, Bertram and Labb{\'e}, Yann and Brachmann, Eric and Michel, Frank and Rother, Carsten and Matas, Ji{\v r}{\'i}},
  year = 2020,
  pages = {577--594},
  publisher = {Springer International Publishing},
  address = {Online},
  doi = {10.1007/978-3-030-66096-3_39},
  urldate = {2024-08-22},
  isbn = {978-3-030-66095-6 978-3-030-66096-3},
  langid = {english}
}

@misc{huang2025,
  title = {{{XYZ-IBD}}: {{A High-precision Bin-picking Dataset}} for {{Object 6D Pose Estimation Capturing Real-world Industrial Complexity}}},
  shorttitle = {{{XYZ-IBD}}},
  author = {Huang, Junwen and Liang, Jizhong and Hu, Jiaqi and Sundermeyer, Martin and Yu, Peter KT and Navab, Nassir and Busam, Benjamin},
  year = 2025,
  publisher = {arXiv},
  doi = {10.48550/arXiv.2506.00599},
  urldate = {2026-01-12},
  copyright = {Creative Commons Attribution Non Commercial Share Alike 4.0 International}
}

@article{huynh2009,
  title = {Metrics for {{3D Rotations}}: {{Comparison}} and {{Analysis}}},
  shorttitle = {Metrics for {{3D Rotations}}},
  author = {Huynh, Du Q.},
  year = 2009,
  month = oct,
  journal = {Journal of Mathematical Imaging and Vision},
  volume = {35},
  number = {2},
  pages = {155--164},
  issn = {0924-9907, 1573-7683},
  doi = {10.1007/s10851-009-0161-2},
  urldate = {2024-08-22},
  copyright = {http://www.springer.com/tdm},
  langid = {english}
}

@inproceedings{irshad2022,
  title = {{{CenterSnap}}: {{Single-Shot Multi-Object 3D Shape Reconstruction}} and {{Categorical 6D Pose}} and {{Size Estimation}}},
  shorttitle = {{{CenterSnap}}},
  booktitle = {2022 {{International Conference}} on {{Robotics}} and {{Automation}} ({{ICRA}})},
  author = {Irshad, Muhammad Zubair and Kollar, Thomas and Laskey, Michael and Stone, Kevin and Kira, Zsolt},
  year = 2022,
  month = may,
  pages = {10632--10640},
  publisher = {IEEE},
  address = {Philadelphia, PA, USA},
  doi = {10.1109/ICRA46639.2022.9811799},
  urldate = {2024-08-23}
}

@inproceedings{kalra2024,
  title = {Towards {{Co-Evaluation}} of {{Cameras}}, {{HDR}}, and {{Algorithms}} for {{Industrial-Grade 6DoF Pose Estimation}}},
  booktitle = {2024 {{IEEE}}/{{CVF Conference}} on {{Computer Vision}} and {{Pattern Recognition}} ({{CVPR}})},
  author = {Kalra, Agastya and Stoppi, Guy and Marin, Dmitrii and Taamazyan, Vage and Shandilya, Aarrushi and Agarwal, Rishav and Boykov, Anton and Chong, Tze Hao and Stark, Michael},
  year = 2024,
  month = jun,
  pages = {22691--22701},
  publisher = {IEEE},
  address = {Seattle, WA, USA},
  doi = {10.1109/CVPR52733.2024.02141},
  urldate = {2025-07-01},
  copyright = {https://doi.org/10.15223/policy-029},
  isbn = {979-8-3503-5300-6},
  langid = {english}
}

@inproceedings{kaskman2019,
  title = {{{HomebrewedDB}}: {{RGB-D Dataset}} for {{6D Pose Estimation}} of {{3D Objects}}},
  shorttitle = {{{HomebrewedDB}}},
  booktitle = {2019 {{IEEE}}/{{CVF International Conference}} on {{Computer Vision Workshop}} ({{ICCVW}})},
  author = {Kaskman, Roman and Zakharov, Sergey and Shugurov, Ivan and Ilic, Slobodan},
  year = 2019,
  month = oct,
  pages = {2767--2776},
  publisher = {IEEE},
  address = {Seoul, Korea (South)},
  doi = {10.1109/ICCVW.2019.00338},
  urldate = {2025-07-01},
  copyright = {https://ieeexplore.ieee.org/Xplorehelp/downloads/license-information/IEEE.html},
  isbn = {978-1-7281-5023-9},
  langid = {english}
}

@inproceedings{kingma2017,
  title = {Adam: {{A Method}} for {{Stochastic Optimization}}},
  shorttitle = {Adam},
  booktitle = {International {{Conference}} on {{Learning Representations}} ({{ICLR}})},
  author = {Kingma, Diederik P. and Ba, Jimmy},
  year = 2017,
  month = jan,
  eprint = {11245/1.505367},
  eprinttype = {hdl},
  publisher = {Ithaca, NY: ArXiv},
  address = {San Diego, CA, US},
  doi = {https://hdl.handle.net/11245/1.505367},
  urldate = {2024-08-30},
  arxiv = {1412.6980 [cs]},
  langid = {english}
}

@inproceedings{kriegler2022,
  title = {{{PrimitivePose}}: {{3D Bounding Box Prediction}} of {{Unseen Objects}} via {{Synthetic Geometric Primitives}}},
  shorttitle = {{{PrimitivePose}}},
  booktitle = {2022 {{Sixth IEEE International Conference}} on {{Robotic Computing}} ({{IRC}})},
  author = {Kriegler, Andreas and Beleznai, Csaba and Murschitz, Markus and G{\"o}bel, Kai and Gelautz, Margrit},
  year = 2022,
  month = dec,
  pages = {190--197},
  publisher = {IEEE},
  address = {Naples, Italy},
  doi = {10.1109/IRC55401.2022.00040},
  urldate = {2024-08-22},
  copyright = {https://doi.org/10.15223/policy-029},
  isbn = {978-1-6654-7260-9}
}

@article{kriegler2023,
  title = {{{PrimitivePose}}: {{Generic Model}} and {{Representation}} for {{3D Bounding Box Prediction}} of {{Unseen Objects}}},
  shorttitle = {{{PrimitivePose}}},
  author = {Kriegler, Andreas and Beleznai, Csaba and Gelautz, Margrit and Murschitz, Markus and G{\"o}bel, Kai},
  year = 2023,
  month = sep,
  journal = {International Journal of Semantic Computing},
  volume = {17},
  number = {3},
  pages = {387--410},
  issn = {1793-351X, 1793-7108},
  doi = {10.1142/S1793351X23620027},
  urldate = {2024-08-23},
  langid = {english}
}

@inproceedings{labbe2020,
  title = {{{CosyPose}}: {{Consistent Multi-view Multi-object 6D Pose Estimation}}},
  shorttitle = {{{CosyPose}}},
  booktitle = {European {{Conference}} on {{Computer Vision}} ({{ECCV}})},
  author = {Labb{\'e}, Yann and Carpentier, Justin and Aubry, Mathieu and Sivic, Josef},
  year = 2020,
  pages = {574--591},
  publisher = {Springer International Publishing},
  address = {Online},
  doi = {10.1007/978-3-030-58520-4_34},
  urldate = {2024-08-22},
  isbn = {978-3-030-58519-8 978-3-030-58520-4},
  langid = {english}
}

@article{lenc2019,
  title = {Understanding {{Image Representations}} by {{Measuring Their Equivariance}} and {{Equivalence}}},
  author = {Lenc, Karel and Vedaldi, Andrea},
  year = 2019,
  month = may,
  journal = {International Journal of Computer Vision},
  volume = {127},
  number = {5},
  pages = {456--476},
  issn = {0920-5691, 1573-1405},
  doi = {10.1007/s11263-018-1098-y},
  urldate = {2024-08-22},
  langid = {english}
}

@inproceedings{li2024,
  title = {{{MRC-Net}}: 6-{{DoF Pose Estimation}} with {{MultiScale Residual Correlation}}},
  shorttitle = {{{MRC-Net}}},
  booktitle = {2024 {{IEEE}}/{{CVF Conference}} on {{Computer Vision}} and {{Pattern Recognition}} ({{CVPR}})},
  author = {Li, Yuelong and Mao, Yafei and Bala, Raja and Hadap, Sunil},
  year = 2024,
  pages = {10476--10486},
  publisher = {IEEE},
  address = {Seattle, WA, USA},
  doi = {10.1109/CVPR52733.2024.00997},
  urldate = {2024-08-22}
}

@inproceedings{lin2017,
  title = {Focal {{Loss}} for {{Dense Object Detection}}},
  booktitle = {2017 {{IEEE International Conference}} on {{Computer Vision}} ({{ICCV}})},
  author = {Lin, Tsung-Yi and Goyal, Priya and Girshick, Ross and He, Kaiming and Dollar, Piotr},
  year = 2017,
  month = oct,
  pages = {2999--3007},
  publisher = {IEEE},
  address = {Venice, Italy},
  doi = {10.1109/ICCV.2017.324},
  urldate = {2024-08-22},
  isbn = {978-1-5386-1032-9}
}

@inproceedings{liu2021,
  title = {{{StereOBJ-1M}}: {{Large-scale Stereo Image Dataset}} for {{6D Object Pose Estimation}}},
  shorttitle = {{{StereOBJ-1M}}},
  booktitle = {2021 {{IEEE}}/{{CVF International Conference}} on {{Computer Vision}} ({{ICCV}})},
  author = {Liu, Xingyu and Iwase, Shun and Kitani, Kris M.},
  year = 2021,
  month = oct,
  pages = {10850--10859},
  publisher = {IEEE},
  address = {Montreal, QC, Canada},
  doi = {10.1109/ICCV48922.2021.01069},
  urldate = {2024-08-22},
  copyright = {https://doi.org/10.15223/policy-029},
  isbn = {978-1-6654-2812-5}
}

@article{liu2025,
  title = {{{GDRNPP}}: {{A Geometry-guided}} and {{Fully Learning-based Object Pose Estimator}}},
  shorttitle = {{{GDRNPP}}},
  author = {Liu, Xingyu and Zhang, Ruida and Zhang, Chenyangguang and Wang, Gu and Tang, Jiwen and Li, Zhigang and Ji, Xiangyang},
  year = 2025,
  month = mar,
  journal = {IEEE Transactions on Pattern Analysis and Machine Intelligence (TPAMI)},
  volume = {47},
  number = {7},
  pages = {5742--5759},
  issn = {0162-8828, 2160-9292, 1939-3539},
  doi = {10.1109/TPAMI.2025.3553485},
  urldate = {2025-05-12},
  copyright = {https://ieeexplore.ieee.org/Xplorehelp/downloads/license-information/IEEE.html}
}

@inproceedings{mahendran2017,
  title = {{{3D Pose Regression Using Convolutional Neural Networks}}},
  booktitle = {2017 {{IEEE Conference}} on {{Computer Vision}} and {{Pattern Recognition Workshops}} ({{CVPRW}})},
  author = {Mahendran, Siddharth and Ali, Haider and Vidal, Ren{\'e}},
  year = 2017,
  month = jul,
  pages = {494--495},
  publisher = {IEEE},
  address = {Honolulu, HI, USA},
  issn = {2160-7516},
  doi = {10.1109/CVPRW.2017.73},
  urldate = {2024-11-14}
}

@inproceedings{mo2022,
  title = {{{ES6D}}: {{A Computation Efficient}} and {{Symmetry-Aware 6D Pose Regression Framework}}},
  shorttitle = {{{ES6D}}},
  booktitle = {2022 {{IEEE}}/{{CVF Conference}} on {{Computer Vision}} and {{Pattern Recognition}} ({{CVPR}})},
  author = {Mo, Ningkai and Gan, Wanshui and Yokoya, Naoto and Chen, Shifeng},
  year = 2022,
  pages = {6708--6717},
  publisher = {IEEE},
  address = {New Orleans, LA, USA},
  doi = {10.1109/CVPR52688.2022.00660},
  urldate = {2024-08-22}
}

@article{morrison2020,
  title = {Learning Robust, Real-Time, Reactive Robotic Grasping},
  author = {Morrison, Douglas and Corke, Peter and Leitner, J{\"u}rgen},
  year = 2020,
  month = mar,
  journal = {The International Journal of Robotics Research},
  volume = {39},
  number = {2-3},
  pages = {183--201},
  issn = {0278-3649, 1741-3176},
  doi = {10.1177/0278364919859066},
  urldate = {2024-08-22},
  langid = {english}
}

@inproceedings{paszke2019,
  title = {{{PyTorch}}: {{An Imperative Style}}, {{High-Performance Deep Learning Library}}},
  booktitle = {Proceedings of the 33rd {{International Conference}} on {{Neural Information Processing Systems}} ({{NIPS}})},
  author = {Paszke, Adam and Gross, Sam and Massa, Francisco and Lerer, Adam and Bradbury, James and Chanan, Gregory and Killeen, Trevor and Lin, Zeming and Gimelshein, Natalia and Antiga, Luca and Desmaison, Alban and Kopf, Andreas and Yang, Edward and DeVito, Zachary and Raison, Martin and Tejani, Alykhan and Chilamkurthy, Sasank and Steiner, Benoit and Fang, Lu and Bai, Junjie and Chintala, Soumith},
  year = 2019,
  pages = {8026--8037},
  publisher = {ACM},
  address = {Vancouver, BC, Canada},
  doi = {10.5555/3454287.3455008},
  langid = {english}
}

@inproceedings{periyasamy2022,
  title = {Learning {{Implicit Probability Distribution Functions}} for {{Symmetric Orientation Estimation}} from {{RGB Images Without Pose Labels}}},
  booktitle = {2022 {{Sixth IEEE International Conference}} on {{Robotic Computing}} ({{IRC}})},
  author = {Periyasamy, Arul Selvam and Denninger, Luis and Behnke, Sven},
  year = 2022,
  month = dec,
  pages = {221--228},
  publisher = {IEEE},
  address = {Naples, Italy},
  doi = {10.1109/IRC55401.2022.00044},
  urldate = {2024-08-22},
  copyright = {https://doi.org/10.15223/policy-029},
  isbn = {978-1-6654-7260-9}
}

@inproceedings{pitteri2019,
  title = {On {{Object Symmetries}} and {{6D Pose Estimation}} from {{Images}}},
  booktitle = {2019 {{International Conference}} on {{3D Vision}} ({{3DV}})},
  author = {Pitteri, Giorgia and Ramamonjisoa, Michael and Ilic, Slobodan and Lepetit, Vincent},
  year = 2019,
  month = sep,
  pages = {614--622},
  publisher = {IEEE},
  address = {Qu\'ebec City, QC, Canada},
  doi = {10.1109/3DV.2019.00073},
  urldate = {2024-08-22},
  copyright = {https://ieeexplore.ieee.org/Xplorehelp/downloads/license-information/IEEE.html},
  isbn = {978-1-7281-3131-3}
}

@inproceedings{pitteri2021,
  title = {{{3D Object Detection}} and {{Pose Estimation}} of {{Unseen Objects}} in {{Color Images}} with {{Local Surface Embeddings}}},
  booktitle = {Asian {{Conference}} on {{Computer Vision}} ({{ACCV}})},
  author = {Pitteri, Giorgia and Bugeau, Aur{\'e}lie and Ilic, Slobodan and Lepetit, Vincent},
  year = 2021,
  pages = {38--54},
  publisher = {Springer International Publishing},
  address = {Cham},
  doi = {10.1007/978-3-030-69525-5_3},
  urldate = {2024-08-23},
  isbn = {978-3-030-69524-8 978-3-030-69525-5},
  langid = {english}
}

@inproceedings{rad2017,
  title = {{{BB8}}: {{A Scalable}}, {{Accurate}}, {{Robust}} to {{Partial Occlusion Method}} for {{Predicting}} the {{3D Poses}} of {{Challenging Objects}} without {{Using Depth}}},
  shorttitle = {{{BB8}}},
  booktitle = {2017 {{IEEE International Conference}} on {{Computer Vision}} ({{ICCV}})},
  author = {Rad, Mahdi and Lepetit, Vincent},
  year = 2017,
  month = oct,
  pages = {3848--3856},
  publisher = {IEEE},
  address = {Venice, Italy},
  doi = {10.1109/ICCV.2017.413},
  urldate = {2024-08-22},
  isbn = {978-1-5386-1032-9}
}

@inproceedings{raj2022,
  title = {Towards {{Object Agnostic}} and {{Robust}} 4-{{DoF Table-Top Grasping}}},
  booktitle = {2022 {{IEEE}} 18th {{International Conference}} on {{Automation Science}} and {{Engineering}} ({{CASE}})},
  author = {Raj, Prem and Kumar, Ashish and Sanap, Vipul and Sandhan, Tushar and Behera, Laxmidhar},
  year = 2022,
  month = aug,
  pages = {963--970},
  address = {Mexico City, Mexico},
  issn = {2161-8089},
  doi = {10.1109/CASE49997.2022.9926708},
  urldate = {2025-01-28}
}

@article{rennie2016,
  title = {A {{Dataset}} for {{Improved RGBD-Based Object Detection}} and {{Pose Estimation}} for {{Warehouse Pick-and-Place}}},
  author = {Rennie, Colin and Shome, Rahul and Bekris, Kostas E. and De Souza, Alberto F.},
  year = 2016,
  month = jul,
  journal = {IEEE Robotics and Automation Letters},
  volume = {1},
  number = {2},
  pages = {1179--1185},
  issn = {2377-3766, 2377-3774},
  doi = {10.1109/LRA.2016.2532924},
  urldate = {2025-07-01},
  copyright = {https://ieeexplore.ieee.org/Xplorehelp/downloads/license-information/IEEE.html}
}

@article{salehi2019,
  title = {Real-{{Time Deep Pose Estimation With Geodesic Loss}} for {{Image-to-Template Rigid Registration}}},
  author = {Salehi, Seyed Sadegh Mohseni and Khan, Shadab and Erdogmus, Deniz and Gholipour, Ali},
  year = 2019,
  month = feb,
  journal = {IEEE Transactions on Medical Imaging},
  volume = {38},
  number = {2},
  pages = {470--481},
  issn = {0278-0062, 1558-254X},
  doi = {10.1109/TMI.2018.2866442},
  urldate = {2024-11-14},
  copyright = {https://ieeexplore.ieee.org/Xplorehelp/downloads/license-information/IEEE.html}
}

@inproceedings{shi2021,
  title = {{{StablePose}}: {{Learning 6D Object Poses}} from {{Geometrically Stable Patches}}},
  shorttitle = {{{StablePose}}},
  booktitle = {2021 {{IEEE}}/{{CVF Conference}} on {{Computer Vision}} and {{Pattern Recognition}} ({{CVPR}})},
  author = {Shi, Yifei and Huang, Junwen and Xu, Xin and Zhang, Yifan and Xu, Kai},
  year = 2021,
  month = jun,
  pages = {15217--15226},
  publisher = {IEEE},
  address = {Nashville, TN, USA},
  issn = {2575-7075},
  doi = {10.1109/CVPR46437.2021.01497},
  urldate = {2024-08-23}
}

@book{stewart2013,
  title = {Symmetry: {{A Very Short Introduction}}},
  shorttitle = {Symmetry},
  author = {Stewart, Ian},
  year = 2013,
  series = {Very Short Introductions},
  edition = {1st},
  number = {353},
  publisher = {Oxford University Press},
  address = {Oxford},
  isbn = {978-0-19-965198-6},
  lccn = {QA174.7.S96 S74 2013}
}

@inproceedings{su2022,
  title = {{{ZebraPose}}: {{Coarse}} to {{Fine Surface Encoding}} for {{6DoF Object Pose Estimation}}},
  shorttitle = {{{ZebraPose}}},
  booktitle = {2022 {{IEEE}}/{{CVF Conference}} on {{Computer Vision}} and {{Pattern Recognition}} ({{CVPR}})},
  author = {Su, Yongzhi and Saleh, Mahdi and Fetzer, Torben and Rambach, Jason and Navab, Nassir and Busam, Benjamin and Stricker, Didier and Tombari, Federico},
  year = 2022,
  month = jun,
  pages = {6728--6738},
  publisher = {IEEE},
  address = {New Orleans, LA, USA},
  doi = {10.1109/CVPR52688.2022.00662},
  urldate = {2024-10-01},
  copyright = {https://doi.org/10.15223/policy-029},
  isbn = {978-1-6654-6946-3},
  langid = {english}
}

@inproceedings{sundermeyer2018,
  title = {Implicit {{3D Orientation Learning}} for {{6D Object Detection}} from {{RGB Images}}},
  booktitle = {European {{Conference}} on {{Computer Vision}} ({{ECCV}})},
  author = {Sundermeyer, Martin and Marton, Zoltan-Csaba and Durner, Maximilian and Brucker, Manuel and Triebel, Rudolph},
  year = 2018,
  pages = {712--729},
  publisher = {Springer International Publishing},
  address = {Munich, Germany},
  doi = {10.1007/978-3-030-01231-1_43},
  urldate = {2024-08-22},
  isbn = {978-3-030-01230-4 978-3-030-01231-1},
  langid = {english}
}

@article{sundermeyer2020,
  title = {Augmented {{Autoencoders}}: {{Implicit 3D Orientation Learning}} for {{6D Object Detection}}},
  shorttitle = {Augmented {{Autoencoders}}},
  author = {Sundermeyer, Martin and Marton, Zoltan-Csaba and Durner, Maximilian and Triebel, Rudolph},
  year = 2020,
  month = mar,
  journal = {International Journal of Computer Vision},
  volume = {128},
  number = {3},
  pages = {714--729},
  issn = {0920-5691, 1573-1405},
  doi = {10.1007/s11263-019-01243-8},
  urldate = {2024-08-22},
  langid = {english}
}

@inproceedings{tejani2014,
  title = {Latent-{{Class Hough Forests}} for {{3D Object Detection}} and {{Pose Estimation}}},
  booktitle = {European {{Conference}} on {{Computer Vision}} ({{ECCV}})},
  author = {Tejani, Alykhan and Tang, Danhang and Kouskouridas, Rigas and Kim, Tae-Kyun},
  year = 2014,
  pages = {462--477},
  publisher = {Springer International Publishing},
  address = {Zurich, Switzerland},
  doi = {10.1007/978-3-319-10599-4_30},
  urldate = {2025-07-01},
  copyright = {http://www.springer.com/tdm},
  isbn = {978-3-319-10598-7 978-3-319-10599-4},
  langid = {english}
}

@inproceedings{tyree2022,
  title = {6-{{DoF Pose Estimation}} of {{Household Objects}} for {{Robotic Manipulation}}: {{An Accessible Dataset}} and {{Benchmark}}},
  shorttitle = {6-{{DoF Pose Estimation}} of {{Household Objects}} for {{Robotic Manipulation}}},
  booktitle = {2022 {{IEEE}}/{{RSJ International Conference}} on {{Intelligent Robots}} and {{Systems}} ({{IROS}})},
  author = {Tyree, Stephen and Tremblay, Jonathan and To, Thang and Cheng, Jia and Mosier, Terry and Smith, Jeffrey and Birchfield, Stan},
  year = 2022,
  month = oct,
  pages = {13081--13088},
  publisher = {IEEE},
  address = {Kyoto, Japan},
  doi = {10.1109/IROS47612.2022.9981838},
  urldate = {2025-07-01},
  copyright = {https://doi.org/10.15223/policy-029},
  isbn = {978-1-6654-7927-1}
}

@article{vidal2018,
  title = {A {{Method}} for {{6D Pose Estimation}} of {{Free-Form Rigid Objects Using Point Pair Features}} on {{Range Data}}},
  author = {Vidal, Joel and Lin, Chyi-Yeu and Llad{\'o}, Xavier and Mart{\'i}, Robert},
  year = 2018,
  month = aug,
  journal = {Sensors},
  volume = {18},
  number = {8},
  pages = {2678},
  issn = {1424-8220},
  doi = {10.3390/s18082678},
  urldate = {2024-08-22},
  copyright = {https://creativecommons.org/licenses/by/4.0/},
  langid = {english}
}

@inproceedings{wang2019,
  title = {{{DenseFusion}}: {{6D Object Pose Estimation}} by {{Iterative Dense Fusion}}},
  shorttitle = {{{DenseFusion}}},
  booktitle = {2019 {{IEEE}}/{{CVF Conference}} on {{Computer Vision}} and {{Pattern Recognition}} ({{CVPR}})},
  author = {Wang, Chen and Xu, Danfei and Zhu, Yuke and {Martin-Martin}, Roberto and Lu, Cewu and {Fei-Fei}, Li and Savarese, Silvio},
  year = 2019,
  month = jun,
  pages = {3338--3347},
  publisher = {IEEE},
  address = {Long Beach, CA, USA},
  doi = {10.1109/CVPR.2019.00346},
  urldate = {2024-08-22},
  copyright = {https://ieeexplore.ieee.org/Xplorehelp/downloads/license-information/IEEE.html},
  isbn = {978-1-7281-3293-8}
}

@inproceedings{wang2021,
  title = {{{GDR-Net}}: {{Geometry-Guided Direct Regression Network}} for {{Monocular 6D Object Pose Estimation}}},
  shorttitle = {{{GDR-Net}}},
  booktitle = {2021 {{IEEE}}/{{CVF Conference}} on {{Computer Vision}} and {{Pattern Recognition}} ({{CVPR}})},
  author = {Wang, Gu and Manhardt, Fabian and Tombari, Federico and Ji, Xiangyang},
  year = 2021,
  month = jun,
  pages = {16606--16616},
  publisher = {IEEE},
  address = {Nashville, TN, USA},
  doi = {10.1109/CVPR46437.2021.01634},
  urldate = {2025-02-17},
  copyright = {https://ieeexplore.ieee.org/Xplorehelp/downloads/license-information/IEEE.html},
  isbn = {978-1-6654-4509-2},
  langid = {english}
}

@article{xu2004,
  title = {The Essential Order of Approximation for Neural Networks},
  author = {Xu, Zongben and Cao, Feilong},
  year = 2004,
  journal = {Science in China Series F: Information Sciences},
  volume = {47},
  number = {1},
  pages = {97--112},
  issn = {1009-2757},
  doi = {10.1360/02yf0221},
  urldate = {2024-08-22},
  langid = {english}
}

@article{xu2005,
  title = {Simultaneous {{Lp-approximation}} Order for Neural Networks},
  author = {Xu, Zong-Ben and Cao, Fei-Long},
  year = 2005,
  month = sep,
  journal = {Neural Networks},
  volume = {18},
  number = {7},
  pages = {914--923},
  issn = {08936080},
  doi = {10.1016/j.neunet.2005.03.013},
  urldate = {2024-08-22},
  copyright = {https://www.elsevier.com/tdm/userlicense/1.0/},
  langid = {english}
}

@inproceedings{zhao2023,
  title = {Learning {{Symmetry-Aware Geometry Correspondences}} for {{6D Object Pose Estimation}}},
  booktitle = {2023 {{IEEE}}/{{CVF International Conference}} on {{Computer Vision}} ({{ICCV}})},
  author = {Zhao, Heng and Wei, Shenxing and Shi, Dahu and Tan, Wenming and Li, Zheyang and Ren, Ye and Wei, Xing and Yang, Yi and Pu, Shiliang},
  year = 2023,
  month = oct,
  pages = {13999--14008},
  publisher = {IEEE},
  address = {Paris, France},
  doi = {10.1109/ICCV51070.2023.01291},
  urldate = {2024-08-22},
  copyright = {https://doi.org/10.15223/policy-029},
  isbn = {979-8-3503-0718-4}
}

@inproceedings{zhou2019a,
  title = {On the {{Continuity}} of {{Rotation Representations}} in {{Neural Networks}}},
  booktitle = {2019 {{IEEE}}/{{CVF Conference}} on {{Computer Vision}} and {{Pattern Recognition}} ({{CVPR}})},
  author = {Zhou, Yi and Barnes, Connelly and Lu, Jingwan and Yang, Jimei and Li, Hao},
  year = 2019,
  month = jun,
  pages = {5738--5746},
  publisher = {IEEE},
  address = {Long Beach, CA, USA},
  doi = {10.1109/CVPR.2019.00589},
  urldate = {2024-08-22},
  copyright = {https://ieeexplore.ieee.org/Xplorehelp/downloads/license-information/IEEE.html},
  isbn = {978-1-7281-3293-8}
}

@misc{zhou2019,
  title = {Objects as {{Points}}},
  author = {Zhou, Xingyi and Wang, Dequan and Kr{\"a}henb{\"u}hl, Philipp},
  year = 2019,
  month = apr,
  number = {arXiv:1904.07850},
  eprint = {1904.07850},
  primaryclass = {cs},
  publisher = {arXiv},
  doi = {10.48550/arXiv.1904.07850},
  urldate = {2024-08-26},
  archiveprefix = {arXiv}
}

\begin{appendices}
\section{SARR Generalization}\label{app_A}%
This section generalizes our proposed symmetry-aware rotation representation SARR. The following definitions were verified for the symmetry classes of ITODD and the chosen 3D geometric primitives (taking into account the limitations discussed in Section \ref{subsec_sarr_limiations}). For any such 3D objects with symmetry class $i$ and symmetry vector $\bm{\kappa}_{i}$, we restrict the Euler angles to the respective canonic space $C_{i}$:
\begin{equation}\label{eq_12}
\begin{split}
    \restr{\alpha}{C_{i}} = \alpha \in \bigg[0, \dots, \frac{2\pi}{\bm{\kappa}_{i, \alpha}}\bigg[, \\
    \restr{\beta}{C_{i}} = \beta \in \bigg[0, \dots, \frac{2\pi}{\bm{\kappa}_{i, \beta}}\bigg[, \\
    \restr{\gamma}{C_{i}} = \gamma \in \bigg[0, \dots, \frac{2\pi}{\bm{\kappa}_{i, \gamma}}\bigg[.
\end{split}
\end{equation}
Assuming intrinsic rotations in `XYZ'-order, the representation SARR is defined as:
\begin{equation}\label{eq_13}
\begin{split}
    \text{SARR}_{i}(\restr{\alpha}{C_{i}}, \restr{\beta}{C_{i}}, \restr{\gamma}{C_{i}}) =
    \begin{bmatrix}
        s_{i, \alpha} & s_{i, \beta} & s_{i, \gamma} \\
        c_{i, \alpha} & c_{i, \beta} & c_{i, \gamma} \\
    \end{bmatrix}
    \coloneqq
    \\
    \begin{bmatrix}
        \sin(\bm{\lambda}_{\alpha}  \restr{\alpha}{C_{i}}) & \sin(\bm{\lambda}_{\beta}  \restr{\beta}{C_{i}})\nu_{\alpha} & 
        \sin(\bm{\lambda}_{\gamma} \restr{\gamma}{C_{i}})\nu_{\alpha}\nu_{\beta} \\
        \cos(\bm{\lambda}_{\alpha}  \restr{\alpha}{C_{i}}) & \cos(\bm{\lambda}_{\beta}  \restr{\beta}{C_{i}}) & \cos(\bm{\lambda}_{\gamma}  \restr{\gamma}{C_{i}})
    \end{bmatrix}.
\end{split}
\end{equation}
The terms
\begin{equation}\label{eq_14}
    \begin{split}
        \bm{\lambda}_{\alpha} = 
        \begin{cases}
            0 & \text{if } \bm{\kappa}_{i, \alpha} \text{ is } \infty, \\
            \bm{\kappa}_{i, \alpha} & \text{otherwise,} 
        \end{cases} \\
                \bm{\lambda}_{\beta} = 
        \begin{cases}
            0 & \text{if } \bm{\kappa}_{i, \beta} \text{ is } \infty, \\
            \bm{\kappa}_{i, \beta} & \text{otherwise,} 
        \end{cases} \\
                \bm{\lambda}_{\gamma} = 
        \begin{cases}
            0 & \text{if } \bm{\kappa}_{i, \gamma} \text{ is } \infty, \\
            \bm{\kappa}_{i, \gamma} & \text{otherwise,} 
        \end{cases}
    \end{split}
\end{equation}
resolve continuous symmetries as shown in \textbf{Eq.}~(\ref{eq_8}). The difference to the representation for T-LESS are the additional terms 
\begin{equation}\label{eq_15}
    \begin{split}
        \nu_{\alpha} = 
        \begin{cases}
            \cos(\restr{\alpha}{C_{i}}) & \text{if } 1 < \bm{\kappa}_{i, \alpha} < \infty, \\
            1.0 & \text{otherwise,} 
        \end{cases}\\
        \nu_{\beta} = 
        \begin{cases}
            \cos(\restr{\beta}{C_{i}}) & \text{if } 1 < \bm{\kappa}_{i, \beta} < \infty, \\
            1.0 & \text{otherwise.} 
        \end{cases}
    \end{split}
\end{equation}
These additional terms are necessary to ensure that visually distinct poses have different numeric representations. For example, one can imagine the rectangular cuboid, $\bm{\kappa}_{\text{CUBOID}} = [\textcolor{red}{2}, \textcolor{green}{2}, \textcolor{blue}{2}]$, with distinct images $\mathcal{V}_{\text{CUBOID}}(0, 0, 45) \neq \mathcal{V}_{\text{CUBOID}}(0, 180, 45)$, yet without $\nu_{\alpha}, \nu_{\beta}$, the  representation would be identical.
\begin{table*}[!t]
\centering
\caption{\textbf{Symmetry classes of ITODD objects.} Each of the symmetry classes $i$ has a representation $\text{SARR}_{i}$, vector $\bm{\kappa}_{i}$, a symmetry type and a set of rotations $\mathcal{R}_{i}$. T-LESS class IDs of the visualized objects are indicated \textbf{bold} in the bottom rows. The default 3D coordinate frame, as shown for class $\RN{2}$, is oriented the same for all objects.}
\includegraphics[width=1.0\linewidth]{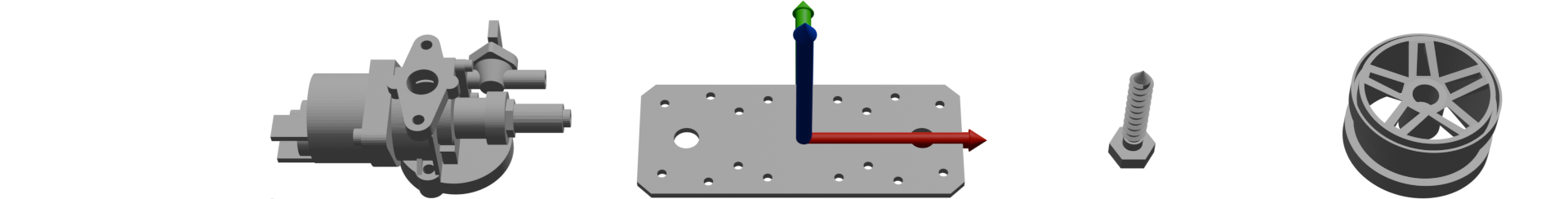}
\label{tab_itodd_symmetries}
\setlength{\tabcolsep}{3.5mm} 
\renewcommand{\arraystretch}{1.0} 
\begin{tabular}{ccccc}
\toprule
\textbf{Class} & $\RN{1}$ & $\RN{2}$ & $\RN{3}$ & $\RN{4}$ \\
\begin{tabular}[c]{@{}c@{}}\textbf{SARR\textsubscript{i}}\\ (ITODD)\end{tabular} & $\text{SARR}_{\RN{1}}$ & $\text{SARR}_{\RN{2}}$ & $\text{SARR}_{\RN{3}}$  & $\text{SARR}_{\RN{4}}$ \\ 
$\bm{\kappa}_{i}$ & $\bm{\kappa}_{\RN{1}} = [\textcolor{red}{1}, \textcolor{green}{1}, \textcolor{blue}{1}]$ & $\bm{\kappa}_{\RN{2}} = [\textcolor{red}{2}, \textcolor{green}{2}, \textcolor{blue}{2}]$ & $\bm{\kappa}_{\RN{3}} = [\textcolor{red}{1}, \textcolor{green}{1}, \textcolor{blue}{\infty}]$ & $\bm{\kappa}_{\RN{4}} = [\textcolor{red}{1}, \textcolor{green}{1}, \textcolor{blue}{5}]$ \\
\textbf{Type} & None & Discrete & Continuous & Discrete  \\
$\mathcal{R}_{i}$ & $\{\}$ & $ \bm{R}_{\textcolor{red}{x}, \textcolor{green}{y}, \textcolor{blue}{z}}^{\alpha, \beta, \gamma} | \alpha,\beta,\gamma \in \{\pi\}$ & $\bm{R}_{\textcolor{blue}{z}}^{\gamma} | \gamma \in \{\mathbb{R}\}$ & \begin{tabular}[c]{@{}c@{}}$\bm{R}_{\textcolor{blue}{z}}^{\gamma} | \gamma \in \{\frac{2n\pi}{5}\},$\\ $n \in \{1,\dots,4\}$ \end{tabular}\\
\begin{tabular}[c]{@{}c@{}}\textbf{ITODD}\\ \textbf{IDs} \end{tabular} & \begin{tabular}[c]{@{}c@{}}\textbf{20}, 1, 2, 4, 5, 6, 10,\\ 13, 15, 16, 21, 22, 26 \end{tabular} & \begin{tabular}[c]{@{}c@{}}\textbf{11}, 3, 19 \end{tabular}  & \textbf{23}, 7, 24, 27 & \textbf{8} \\
\bottomrule
\end{tabular}
\includegraphics[width=1.0\linewidth]{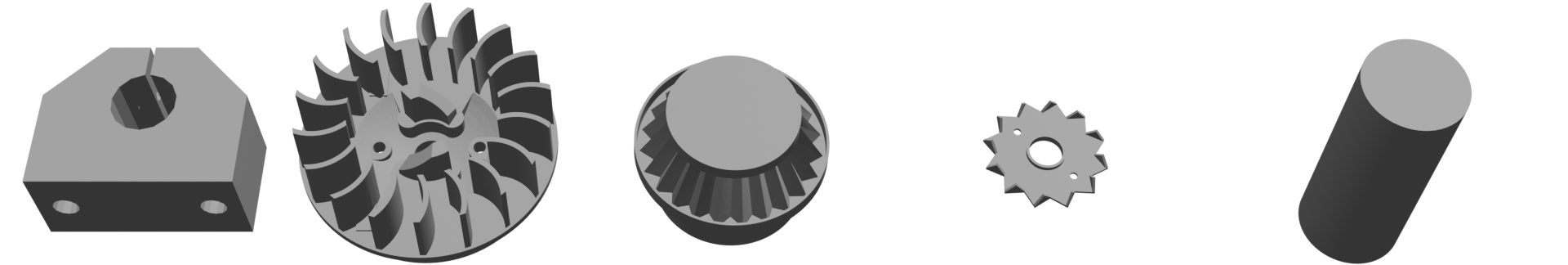}
\setlength{\tabcolsep}{4mm} 
\renewcommand{\arraystretch}{1.0} 
\begin{tabular}{ccccc}
$\RN{5}$ & $\RN{6}$ & $\RN{7}$ & $\RN{8}$ & $\RN{9}$\\
$\text{SARR}_{\RN{5}}$ & $\text{SARR}_{\RN{6}}$ & $\text{SARR}_{\RN{7}}$  & $\text{SARR}_{\RN{8}}$  & $\text{SARR}_{\RN{9}}$ \\ 
$\bm{\kappa}_{\RN{5}} = [\textcolor{red}{1}, \textcolor{green}{2}, \textcolor{blue}{1}]$ & $\bm{\kappa}_{\RN{6}} = [\textcolor{red}{1}, \textcolor{green}{1}, \textcolor{blue}{18}]$ & $\bm{\kappa}_{\RN{7}} = [\textcolor{red}{1}, \textcolor{green}{1}, \textcolor{blue}{23}]$ & $\bm{\kappa}_{\RN{8}} = [\textcolor{red}{1}, \textcolor{green}{1}, \textcolor{blue}{12}]$ & $\bm{\kappa}_{\RN{9}} = [\textcolor{red}{2}, \textcolor{green}{2}, \textcolor{blue}\infty{}]$\\
 Discrete & Discrete & Discrete & Discrete & \begin{tabular}[c]{@{}c@{}}Discrete \&\\Continuous \end{tabular}\\
$\bm{R}_{\textcolor{green}{y}}^{\beta} | \beta \in \{\pi\}$  &  \begin{tabular}[c]{@{}c@{}}$\bm{R}_{\textcolor{blue}{z}}^{\gamma} | \gamma \in \{\frac{2n\pi}{18}\}$, \\$n \in \{1,\dots,17\}$ \end{tabular} & \begin{tabular}[c]{@{}c@{}}$\bm{R}_{\textcolor{blue}{z}}^{\gamma} | \gamma \in \{\frac{2n\pi}{23}\}$, \\$n \in \{1,\dots,22\}$ \end{tabular} & \begin{tabular}[c]{@{}c@{}}$\bm{R}_{\textcolor{blue}{z}}^{\gamma} | \gamma \in \{\frac{2n\pi}{12}\}$, \\$n \in \{1,\dots,11\}$ \end{tabular} & \begin{tabular}[c]{@{}c@{}}$\bm{R}_{\textcolor{red}{x}, \textcolor{green}{y}, \textcolor{blue}{z}}^{\alpha, \beta, \gamma} | \alpha,\beta \in \{\pi\}$, \\$\gamma \in \{\mathbb{R}\}$ \end{tabular} \\
\textbf{9}, 18 & \textbf{14}  & \textbf{17} & \textbf{25} & \textbf{12}, 28  \\
\bottomrule
\end{tabular}
\end{table*}\\
Inverse mapping the SARR representation is done sequentially, starting with the rotation first in chain, $\alpha$. The $\nu$ terms are computed as in \textbf{Eq.}~(\ref{eq_15}), but done so now on the fly, using the calculated angle from the previous step:
\begin{equation}\label{eq_16}
    \begin{split}
        \restr{\alpha}{C_{i}} = 
        \begin{cases}
            0 & \text{if } \bm{\kappa}_{i, \alpha} \text{ is } \infty, \\
            \frac{1}{\bm{\kappa}_{i, \alpha}}(2\pi-\arccos(c_{i, \alpha})) & \text{if } s_{i, \alpha} < 0, \\
            \frac{1}{\bm{\kappa}_{i, \alpha}}\arccos(c_{i, \alpha}) & \text{otherwise,}
        \end{cases}
        \\
        \restr{\beta}{C_{i}} = 
        \begin{cases}
            0 & \text{if } \bm{\kappa}_{i, \beta} \text{ is } \infty, \\
            \frac{1}{\bm{\kappa}_{i, \beta}}(2\pi-\arccos(c_{i, \beta})) & \text{if } \frac{s_{i, \beta}}{\nu_{\alpha}} < 0, \\
            \frac{1}{\bm{\kappa}_{i, \beta}}\arccos(c_{i, \beta}) & \text{otherwise,}
        \end{cases}
        \\
        \restr{\gamma}{C_{i}} = 
        \begin{cases}
            0 & \text{if } \bm{\kappa}_{i, \gamma} \text{ is } \infty, \\
            \frac{1}{\bm{\kappa}_{i, \gamma}}(2\pi-\frac{\arccos(c_{i, \gamma})}{\bm{\kappa}_{i, \gamma}}\color{black}) & \text{if } \frac{s_{i, \gamma}}{\nu_{\alpha}\nu_{\beta}} < 0, \\
            \frac{1}{\bm{\kappa}_{i, \gamma}}\arccos(c_{i, \gamma}) & \text{otherwise.}
        \end{cases}
    \end{split}
\end{equation}
As can be seen from equations \textbf{Eq.}~(\ref{eq_13}) and \textbf{Eq.}~(\ref{eq_16}), the $\nu$ terms introduce a singularity at $90^{\circ}$ for symmetry classes of type $\bm{\kappa} = [\{2, 3\}, \{2, 3\}, n], n \in \mathbb{Z}$. Fortunately, this is mostly a theoretic problem, because it is exceedingly rare that the symmetry class $\RN{2}$ objects from ITODD, or any object in the real world for that matter, have a rotation of \textit{exactly} $90^{\circ}$ -- our networks trained on those objects have converged successfully. For symmetry class $\RN{5}$ of ITODD, and when considering the full SO(3) rotation space, inverse mapping is done not via \textbf{Eq.}~(\ref{eq_16}) but instead via:
\begin{equation}\label{eq_17}
    \begin{split}
        \restr{\alpha}{C_{i}} = 
        \begin{cases}
            2\pi-\arccos(c_{i, \alpha}) & \text{if } s_{i, \alpha} < 0, \\
            \arccos(c_{i, \alpha}) & \text{otherwise,}
        \end{cases}
        \\
        \restr{\beta}{C_{i}} = 
        \begin{cases}
            \frac{2\pi}{\bm{\kappa}_{i, \beta}} - \frac{\arccos(c_{i, \beta})}{\bm{\kappa}_{i, \beta}} & \text{if } s_{i, \beta} < 0, \\
            \frac{1}{\bm{\kappa}_{i, \beta}}\arccos(c_{i, \beta}) & \text{otherwise,}
        \end{cases}
        \\
        \restr{\gamma'}{C_{i}} = 
        \begin{cases}
            2\pi-\arccos(c_{i, \gamma}) & \text{if } s_{i, \gamma} < 0, \\
            \arccos(c_{i, \gamma}) & \text{otherwise,}
        \end{cases}
        \\
         \restr{\gamma}{C_{i}} = 
        \begin{cases}
            -\restr{\gamma'}{C_{i}} & \text{if } s_{i, \beta} < 0, \\
             \restr{\gamma'}{C_{i}} & \text{otherwise.}
        \end{cases}
    \end{split}
\end{equation}
An overview of all the ITODD symmetry classes is shown in \textbf{Table}~\ref{tab_itodd_symmetries}. We provide an alternative definition for the symmetry classes of some ITODD objects. Specifically, we define object 23, a screw, to be continuously symmetric, i.e. to belong to class $\RN{3}$. We also consider objects 2, 4 and 5 non-symmetric. In each case, their object centroid in the 3D model was defined at a location s.t. no single rotation $\bm{r}$ about any axis could result in a visually identical pose. For object 2, because the centroid is shifted far off the symmetry-axis thus requiring a rototranslation, and for objects 4 and 5 because the transformation requires chaining multiple elementary rotations. We use this alternative symmetry classification for training all our networks and evaluating under the AR\textsubscript{C} metric, yet we use the original definition from BOP when computing AR\textsubscript{B} scores.\\
Regarding 3D geometric primitives, the above definition also works for many common primitives. The ones we consider are cuboids, cylinders, torii and spheres. Cuboids themselves can be split into five different symmetry classes:
\begin{description}[labelwidth=1cm, leftmargin=1cm]
    \item[CUBOID:] rectangular prism, every face is a non-square rectangle,
    \item[CUB\_XY:] the two faces parallel to the XY plane are squares, the other four are non-square rectangles,
    \item[CUB\_XZ:] the two faces parallel to the XZ plane are squares, the other four are non-square rectangles,
    \item[CUB\_YZ:] the two faces parallel to the YZ plane are squares, the other four are non-square rectangles,
    \item[CUBE:] a cube, all faces are squares.
\end{description}
Assuming upright default pose (see the default coordinate frame in \textbf{Table}~\ref{tab_itodd_symmetries}), the symmetry vectors for these primitives are: $\bm{\kappa}_{\text{CUBOID}} = [\textcolor{red}{2}, \textcolor{green}{2}, \textcolor{blue}{2}]$, $\bm{\kappa}_{\text{CUB\_XY}} = [\textcolor{red}{2}, \textcolor{green}{2}, \textcolor{blue}{4}]$, $ \bm{\kappa}_{\text{CUB\_XZ}} = [\textcolor{red}{2}, \textcolor{green}{4}, \textcolor{blue}{2}]$, $\bm{\kappa}_{\text{CUB\_YZ}} = [\textcolor{red}{4}, \textcolor{green}{2}, \textcolor{blue}{2}]$, $\bm{\kappa}_{\text{CUBE}} = [\textcolor{red}{4}, \textcolor{green}{4}, \textcolor{blue}{4}]$, $\bm{\kappa}_{\text{CYLINDER}} \equiv \bm{\kappa}_{\text{TORUS}} = [\textcolor{red}{2}, \textcolor{green}{2}, \textcolor{blue}{\infty}]$, $\bm{\kappa}_{\text{SPHERE}} = [\textcolor{red}{\infty}, \textcolor{green}{\infty}, \textcolor{blue}{\infty}]$.\\
\section{SARR Algorithm}\label{app_B}%
\textbf{Algorithm} \ref{alg_sarr} shows the entire process of mapping a rotation representation, here a $3\times3$ rotation matrix, to our proposed SARR representation, followed by inverse mapping back to a rotation matrix $\restr{\mathbf{R}}{C}$. While the SARR representation, which is used in the algorithm as an intermediate result, fulfills the uniqueness and continuity properties, $\restr{\mathbf{R}}{C}$ on the other hand is also unique for the given symmetry classes w.r.t. the visual appearance, but not continuous.  This generalized algorithm is valid for the symmetry classes and rotation spaces as shown in \textbf{Table}~\ref{tab_tless_symmetries} and \textbf{Table}~\ref{tab_itodd_symmetries} with the exception of class V (both datasets). Mapping for symmetry class V, due to its specific property of having one axis of symmetry which is not the last in the defined rotation order, requires some modifications which are shown at the bottom of the algorithm.
\begin{algorithm*}
  \caption{Mapping a standard rot.-mat. $\mathbf{R}$ to a canonic rot.-mat. $\restr{\mathbf{R}}{C}$ via the SARR representation.}\label{alg_sarr}
  \begin{algorithmic}[1]
    \Function{CanonicViaSARR}{$\mathbf{R}$, sym\_cls}\Comment{\begin{footnotesize}$\mathbf{R} \in \text{SO(3)}, \text{sym\_cls} \in \{\RN{1}, \dots, \RN{9}\}$\end{footnotesize}}
        \State $\bm{\kappa} \equiv [\bm{\kappa}_{\alpha}, \bm{\kappa}_{\beta}, \bm{\kappa}_{\gamma}] \coloneqq \bm{\kappa}_{\text{sym\_cls}} $ \Comment{\begin{footnotesize}$\bm{\kappa}_{\alpha}, \bm{\kappa}_{\beta}, \bm{\kappa}_{\gamma} \in \mathbb{Z}^{+}$\end{footnotesize}}
        \State $\text{SARR} \gets \text{\textsc{Forward}}(\mathbf{R}, \bm{\kappa})$  \Comment{\begin{footnotesize}$\text{SARR} \in \mathbb{R}^{2\times3}$, SARR\textsubscript{i, j} $\in [-1, 1]$\end{footnotesize}}
        \State $\restr{\mathbf{R}}{C} \gets \text{\textsc{Inverse}}(\text{SARR}, \bm{\kappa})$ \Comment{\begin{footnotesize}$\restr{\mathbf{R}}{C} \in C \text{ with } C \subset$ \text{SO(3) if any}$(\bm{\kappa}_{i}) > 1$\text{, else} $C \equiv $ SO(3)\end{footnotesize}}
    \EndFunction
    \Statex
    \Function{Forward}{$\mathbf{R}, \bm{\kappa}$}
        \State $\alpha, \beta, \gamma \gets \text{R\_to\_Euler}(\mathbf{R}\text{, `XYZ', `radians'})$ \Comment{\begin{footnotesize}Convert $\mathbf{R}$ to Euler angles, in intrinsic `XYZ' order\end{footnotesize}}
        \State $\restr{\alpha}{C} \gets (\alpha \bmod \frac{2\pi}{\bm{\kappa}_{\alpha}}) \frac{\bm{\kappa}_{\alpha} \bmod 10^{3}}{\bm{\kappa}_{\alpha}}$ \Comment{\begin{footnotesize}Clamp the three angles to unique subspaces\end{footnotesize}}
        \State $\restr{\beta}{C} \gets (\beta \bmod \frac{2\pi}{\bm{\kappa}_{\beta}}) \frac{\bm{\kappa}_{\beta} \bmod 10^{3}}{\bm{\kappa}_{\beta}}$\Comment{\begin{footnotesize}$\bm{\kappa}_{i}=10^{3}$ handles continuous symmetry\end{footnotesize}}
        \State $\restr{\gamma}{C} \gets (\gamma \bmod \frac{2\pi}{\bm{\kappa}_{\gamma}}) \frac{\bm{\kappa}_{\gamma} \bmod 10^{3}}{\bm{\kappa}_{\gamma}}$ \Comment{\begin{footnotesize}A different clamping is required for class $\RN{5}$, see lines 29-33\end{footnotesize}}
        \State $\nu_{\alpha} \gets \cos(\restr{\alpha}{C}) \text{ if } 2 <= \bm{\kappa}_{\alpha} < 10^{3} \text{ else } 1$\Comment{\begin{footnotesize}Handles objects with multiple symmetry axes\end{footnotesize}}  
        \State $\nu_{\beta} \gets \cos(\restr{\beta}{C}) \text{ if } 2 <= \bm{\kappa}_{\beta} < 10^{3} \text{ else } 1$
        \State $s_{\alpha} \gets \sin(\bm{\kappa}_{\alpha}\restr{\alpha}{C})$\Comment{\begin{footnotesize}Build multi-valued trigonometrics\end{footnotesize}}
        \State $c_{\alpha} \gets \cos(\bm{\kappa}_{\alpha}\restr{\alpha}{C})$\Comment{\begin{footnotesize}Six parameters make up the SARR representation\end{footnotesize}}
        \State $s_{\beta} \gets \sin(\bm{\kappa}_{\beta}\restr{\beta}{C})\nu_{\alpha}$ 
        \State $c_{\beta} \gets \cos(\bm{\kappa}_{\beta}\restr{\beta}{C})$
        \State $s_{\gamma} \gets \sin(\bm{\kappa}_{\gamma}\restr{\gamma}{C})\nu_{\alpha}\nu_{\beta}$
        \State $c_{\gamma} \gets \cos(\bm{\kappa}_{\gamma}\restr{\gamma}{C})$
        \State \textbf{return} SARR $\gets$ concat\_transpose($s_{\alpha}$, $c_{\alpha}$, $s_{\beta}$, $c_{\beta}$, $s_{\gamma}$, $c_{\gamma}$)\Comment{\begin{footnotesize}Get SARR in desired shape, i.e. $2\times3$\end{footnotesize}}
    \EndFunction
    \Statex
    \Function{Inverse}{SARR, $\bm{\kappa}$}\Comment{\begin{footnotesize}Reconstruct the angles\end{footnotesize}}
        \State $\restr{\alpha}{C} \gets 0$ if $\bm{\kappa}_{\alpha} = 10^{3}$ else $\frac{2\pi-\arccos(c_{\alpha})}{\bm{\kappa}_{\alpha}}$ if $s_{\alpha}<0$ else $\frac{\arccos(c_{\alpha})}{\bm{\kappa}_{\alpha}}$
        \State $\nu_{\alpha} \gets \cos(\restr{\alpha}{C}) \text{ if } 2 <= \bm{\kappa}_{\alpha} < 10^{3} \text{ else } 1$\Comment{\begin{footnotesize}via unit-circle endpoints\end{footnotesize}}
        \State $\restr{\beta}{C} \gets 0$ if $\bm{\kappa}_{\beta} = 10^{3}$ else $\frac{2\pi-\arccos(c_{\beta})}{\bm{\kappa}_{\beta}}$ if $\frac{s_{\alpha}}{\nu_{\alpha}}<0$ else$\frac{\arccos(c_{\alpha})}{\bm{\kappa}_{\alpha}}$
        \State $\nu_{\beta} \gets \cos(\restr{\beta}{C}) \text{ if } 2 <= \bm{\kappa}_{\beta} < 10^{3} \text{ else } 1$\Comment{\begin{footnotesize}$\nu_{i}$ are computed on-the-fly\end{footnotesize}}
        \State $\restr{\gamma}{C} \gets 0 $ if $\bm{\kappa}_{\gamma} = 10^{3}$ else $\frac{1}{\bm{\kappa}_{\gamma}}(2\pi-\frac{\arccos(c_{\gamma})}{\bm{\kappa}_{\gamma}})$ if $\frac{s_{\alpha}}{\nu_{\alpha}\nu_{\beta}}<0$ else $\frac{\arccos(c_{\gamma})}{\bm{\kappa}_{\gamma}}$
        \State \textbf{return} $\restr{\mathbf{R}}{C} \gets$ Euler\_to\_R$(\restr{\alpha}{C}, \restr{\beta}{C}, \restr{\gamma}{C}$, `XYZ') \Comment{\begin{footnotesize}Converts Euler angles to a rot.-mat.\end{footnotesize}}
    \EndFunction
    \Statex
    \If{$\alpha\bmod2\pi > \pi$}\Comment{\begin{footnotesize}Replaces lines 8-10 for symmetry class V\end{footnotesize}}
        \State $\restr{\alpha}{C} \gets (\alpha - \pi) \bmod 2\pi$
        \State $\restr{\beta}{C} \gets -\beta$
        \State $\restr{\gamma}{C} \gets (\pi - \gamma) \bmod 2\pi$
    \EndIf
    \Statex
    \State $\restr{\alpha}{C} \gets2\pi-\arccos(c_{\alpha})$ if $s_{\alpha}<0$ else $\arccos(c_{\alpha})$\Comment{\begin{footnotesize}Replaces lines 22-26 for symmetry class V\end{footnotesize}}
    \State $\restr{\beta}{C} \gets \frac{2\pi-\arccos(c_{\beta})}{\bm{\kappa}_{\beta}}$ if $s_{\beta}<0$ else $\frac{\arccos(c_{\beta})}{\bm{\kappa}_{\beta}}$
    \State $\restr{\gamma}{C} \gets (-1$ if $s_{\beta} < 0)(2\pi-\arccos(c_{\gamma})$ if $s_{\gamma}<0$ else $\arccos(c_{\gamma}))$
    \end{algorithmic}
\end{algorithm*}
\color{black}
\section{Experimental Setup}\label{app_C}%
This section provides details regarding data pre-processing, implementation details of our networks and a discussion regarding loss functions.
\subsection{Image Preprocessing}\label{app_C1}%
\textbf{T-LESS:} We take a centered crop and set the pixels not part of the ground-truth object mask to 0. Masking allows the network to focus on the object by removing spurious background signals, which is especially relevant for the test images, as there is significant object clutter in the test scenes. For depth images, we calculate the mean pixel value using the visibility mask (a different mask, which only includes actually visible pixels), and use this mean to fill pixels that fall outside the defined minimum-maximum depth range: $d_{min}=530, d_{max}=929$ millimeters \mbox{\citep{hodan2016}}. We then cut out the object using the object mask, resize and pad the patch to $384\times384$ while preserving the aspect ratio (patches of size $384\times384$ are expected by CenterNet). We then clamp the depth map to $[d_{min}, d_{max}]$, since resizing can result in depth values outside this range, before inversing the depth, i.e. further objects now appear darker. We use these depth and RGB patches to calculate the mean and standard deviation for data standardization, for every object class, symmetry class and across the entire dataset, in line with the different scopes used during network training.\\
For test images we perform the same augmentations, but before masking we also divide every depth map $\bm{d}_{c, i}$ by $\bar{t}_{c, z}$, where $\bm{d}_{c, i}$ is the depth map of instance $i$ of object class $c$, and $\bar{t}_{c, z}$ is the average translation in \textcolor{blue}{z}-direction, or ``center-depth'', of all training instances of object class $c$. This effectively removes the impact of translations on the depth distribution, resulting in depth maps with variance that stems only from the shape and orientation of the objects, which in turn aligns training and test samples more closely.\\
\textbf{ITODD:} We use the 50k synthetic PBR training images, from which we select only the samples with a visibility of at least $60\%$ and require that the center-depth, given via $t_{z}$, satisfies $695mm < t_{z} < 761mm$. This depth range was derived from the statistics of the validation set. We then crop out and mask the object instance. We resize-pad to $384\times384$ while preserving the aspect ratio, clamp the depth map to $[650, 770]$ millimeters, scaled to $[0, 1]$. For the ablation study we convert the RGB patches to grayscale using \texttt{cv2.cvtColor(img, cv2.COLOR\_RGB2GRAY)}. We do not perform any standardization or depth map scaling via the average center-depth like we did for T-LESS.\\
The real depth images from the validation set used for the evaluation feature many invalid depth pixels due to reflections on the surfaces of the metallic objects. We therefore fill the depth patch after cropping and masking in a reverse-watershed like manner, using simple dilation, guided by the object mask. Depth and grayscale patches are then resize-padded, depth samples are additionally scaled like the training patches.

\subsection{Network Implementation}\label{app_C2}%
We use PyTorch \citep{paszke2019} to train a modified CenterNet \citep{zhou2019} network, using HardNet \citep{chao2019} as the backbone. We train all networks for $40$ epochs, optimizing with Adam \citep{kingma2017}. We use PyTorch's \texttt{ReduceLROnPlateau}\footnote{\url{https://docs.pytorch.org/docs/stable/generated/torch.optim.lr_scheduler.ReduceLROnPlateau.html} (Accessed: 2026-01-12)}. function as a learning rate scheduler. We use \texttt{`sum'} for loss reduction (taking the sum of the losses of all samples in the batch for backpropagation) and a batch size of $4$ per GPU. Loss terms for the three elementary rotations are weighted equally with $1$. We use the heatmap head to learn the symmetry class when training the SARR-\textit{dataset*} networks, using FocalLoss \citep{lin2017}. Number of total epochs, initial learning rate, learning rate decay parameters and batch size were initially optimized by observing the loss-plots of networks trained using SARR on T-LESS. \textbf{Table}~\ref{tab_hyper} shows all relevant hyperparameters of our experiments.
\begin{table}[!ht]
    \centering
    \caption{\textbf{Network hyperparameters.}}
    \label{tab_hyper}
    \setlength{\tabcolsep}{1.0pt} 
    \renewcommand{\arraystretch}{1.0} 
    \begin{tabular}{|c|cc|}
        \hline
        \textbf{Parameter} & \textbf{T-LESS} & \textbf{ITODD} \\
        \hline
        Backbone & HardNet & HardNet \\
        Epochs & 40 & 40 \\
        Batch-size & 4 & 12 \\
        Input-size & $384\times384$ & $384\times384$ \\
        Optimizer & Adam & Adam \\
        Rotation-loss & Cosine/L1 & Cosine/L1 \\
        Symm.-Class-loss & FocalLoss & FocalLoss \\
        Loss-reduction & \texttt{`sum'} & \texttt{`sum'} \\
        Loss-weights & 1.0 for all & 1.0 for all \\
        Learning rate & $6*10^{-4}$ & $6*10^{-4}$ \\
        LR-scheduler & \texttt{ReduceLROnPlateau} & \texttt{ReduceLROnPlateau} \\
        LR-factor & 0.5 & 0.5 \\
        LR-patience & 2 & 2 \\
        LR-threshold & 0.15 & 0.1 \\
        LR-cooldown & 1 & 1 \\
        GPUs & 1 $\times$ RTX 3090 & 3 $\times$ RTX 3090 \\
        \hline
    \end{tabular}
\end{table}
\subsection{Loss Functions}\label{app_C3}%
To ensure a fair comparison between the different rotation representations we want to train all networks as similar as possible, yet each representation has slightly different mathematical properties (dimensionality, range). A common ground was found using the cosine distance, interpreted as the cosine similarity loss function. For ground truths and predictions, the quaternion ($n=4$), trigonometric ($n=6$), rotation matrix ($n=9$), 6d ($n=6$) and SARR ($n=6$) representations are flattened to a $1\times n$ vector and the difference between true and predicted vectors is computed as the cosine distance. Since the cosine similarity loss does not enforce magnitude but only directionality between the two vectors, resulting predictions are normalized in a post-processing step for visualization and evaluation. This normalization happens column-wise for the trigonometric, rotation matrix and SARR representations, where each column is necessarily unit-length. Versors (quaternions that represent rotations) are also unit-length and thus the entire quaternion can be normalized. When mapping the 6d representation to a rotation matrix, normalization happens by definition. Euler angles do not allow normalization  - we therefore use standard L1 loss for Euler angles. Regarding the geodesic loss \citep{mahendran2017} that is often used for rotation matrices: the trace-operations involved in the calculation of the loss during training require matrices to already have unit-length columns. This means the output of the network has to be constrained in some way, for example by adding another layer \citep{salehi2019}. The comparison would then no longer be with networks of identical architecture, resulting in an unfair comparison.

\section{Symmetry Classification}\label{app_D}%
\begin{figure*}[p]
    \begin{multicols}{2}
    \centering
    \begin{subfigure}{0.65\linewidth}
    \centering
    \includegraphics[width=1.0\linewidth]{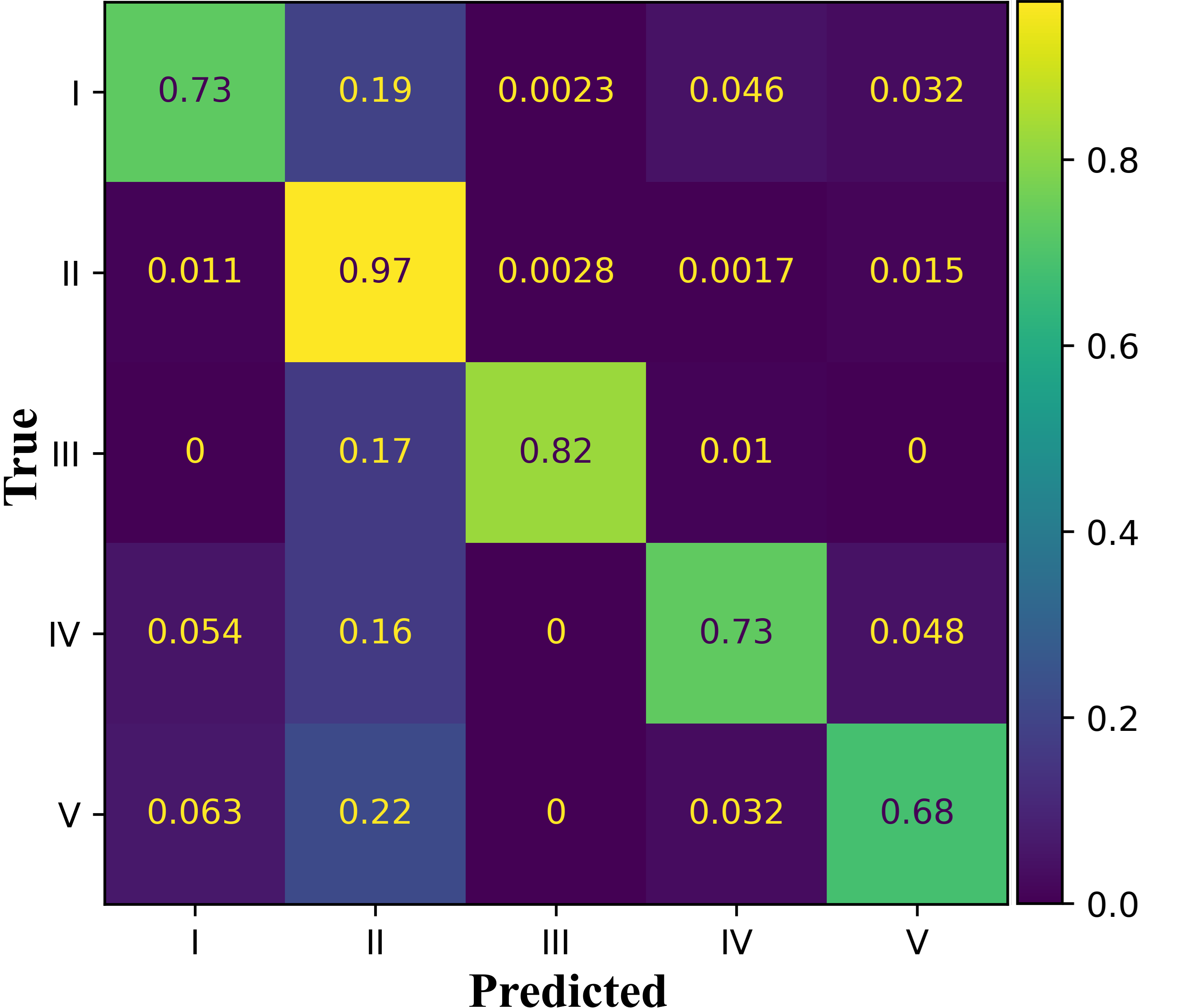}
    \caption{Normalized CFM for SiSo task}
    \label{fig_cfm_a}
    \end{subfigure}
    \hfill
    \begin{subfigure}{0.65\linewidth}
    \centering
    \includegraphics[width=1.0\linewidth]{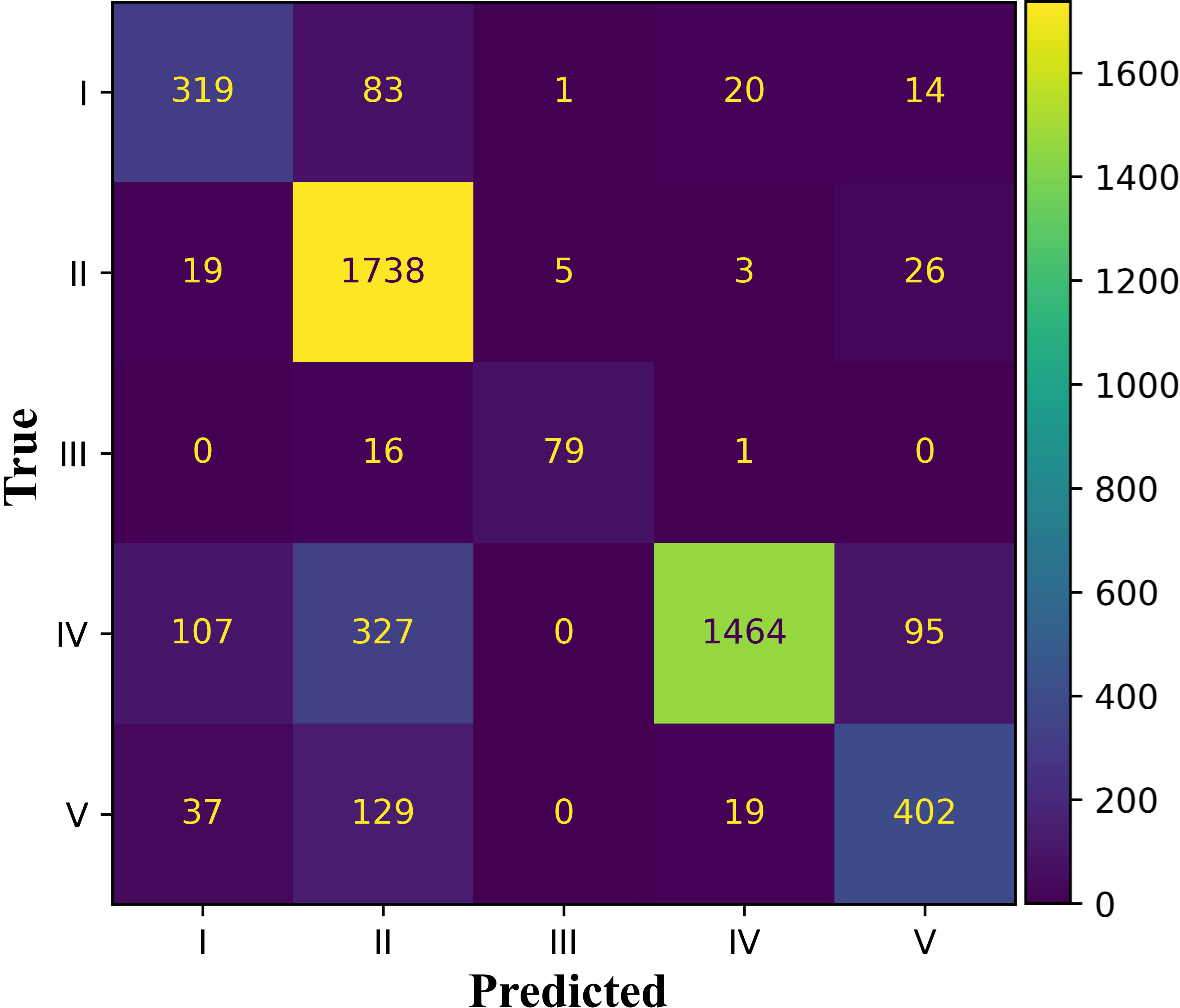}
    \caption{Absolute CFM for SiSo task}
    \label{fig_cfm_b}
    \end{subfigure}
    \begin{subfigure}{0.65\linewidth}
    \centering
    \includegraphics[width=1.0\linewidth]{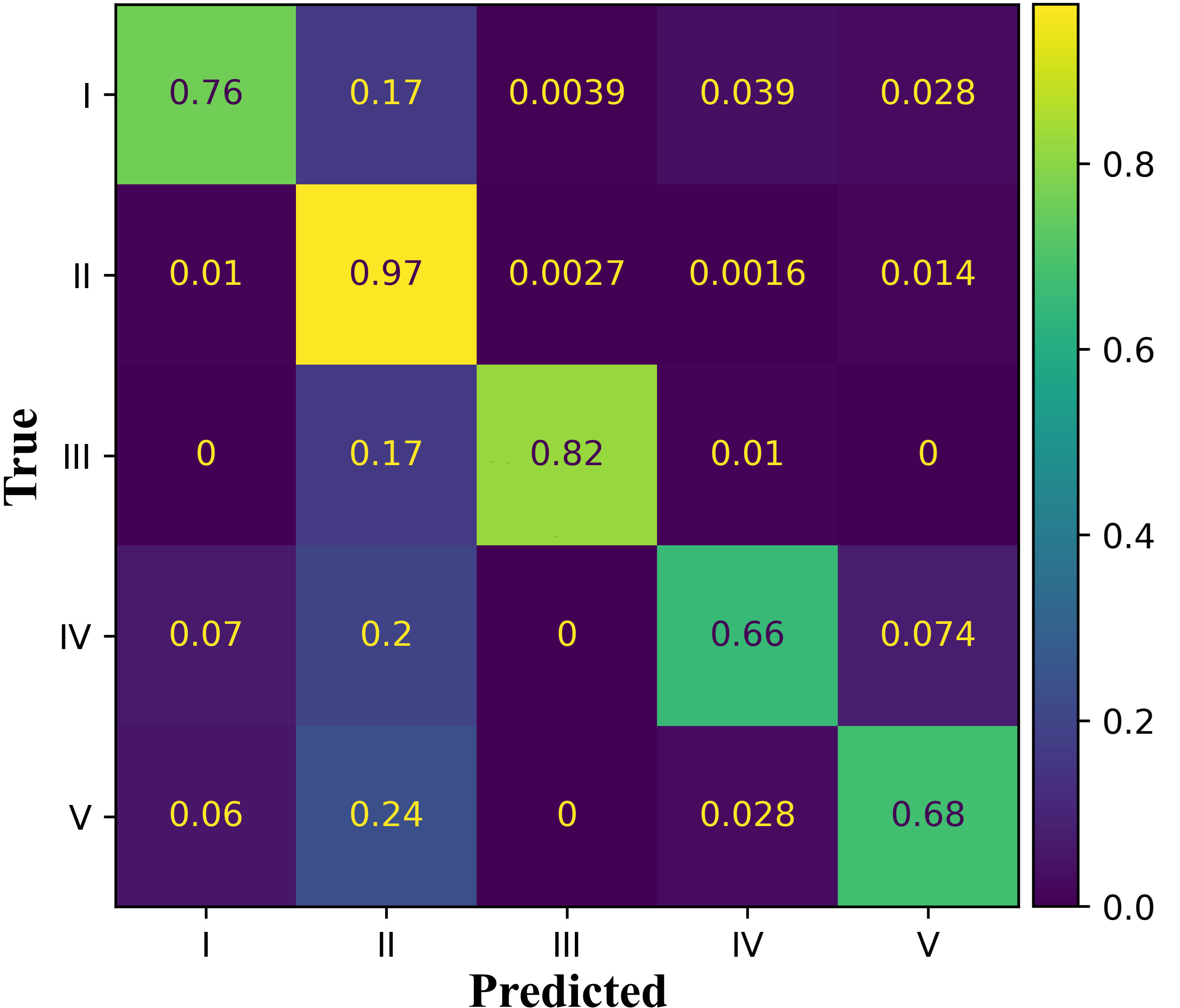}
    \caption{Normalized CFM for ViVo task}
    \label{fig_cfm_c}
    \end{subfigure}
    \hfill
    \begin{subfigure}{0.65\linewidth}
    \centering
    \includegraphics[width=1.0\linewidth]{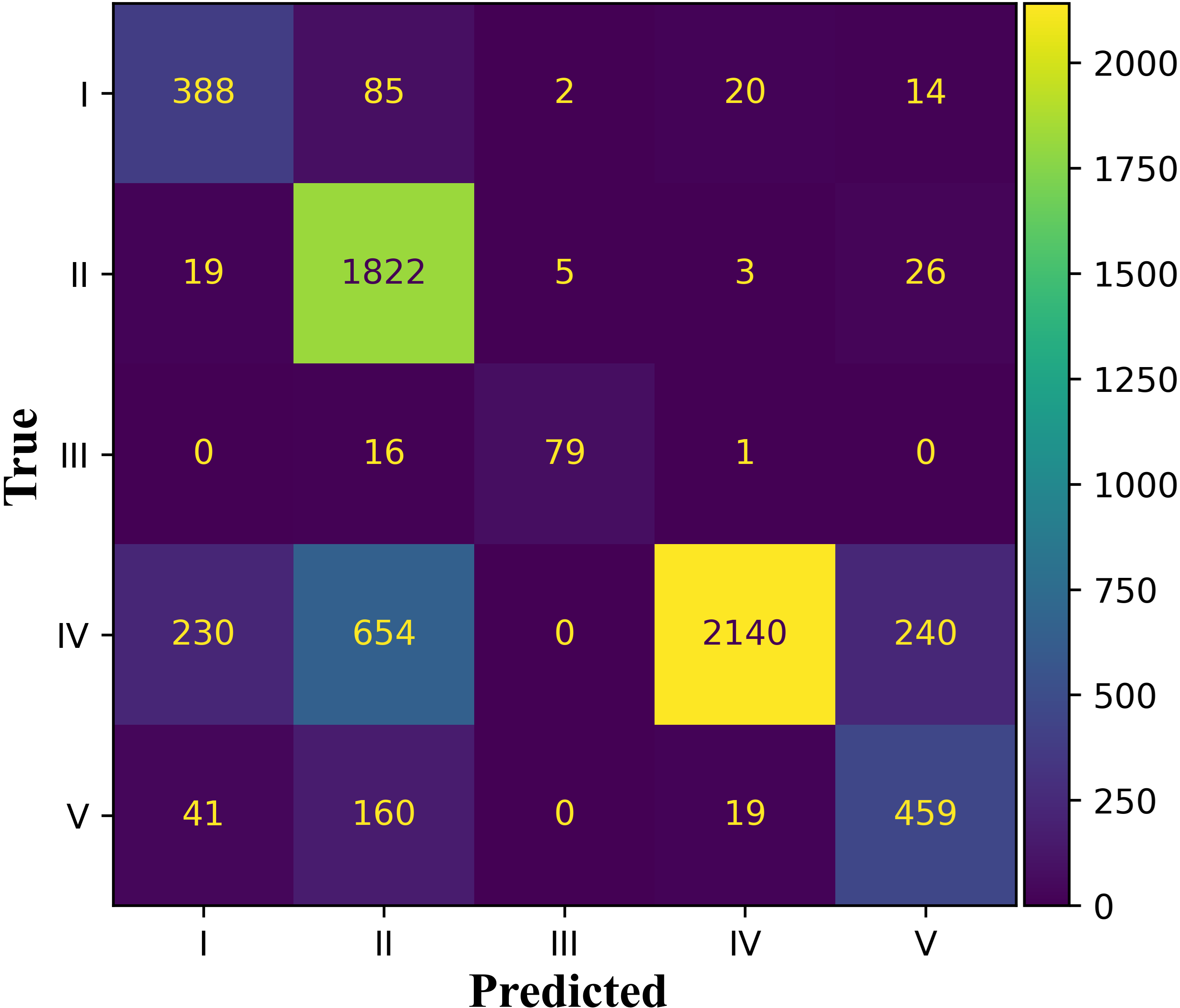}
    \caption{Absolute CFM for ViVo task}
    \label{fig_cfm_d}
    \end{subfigure}
    \begin{subfigure}{0.65\linewidth}
    \centering
    \includegraphics[width=1.0\linewidth]{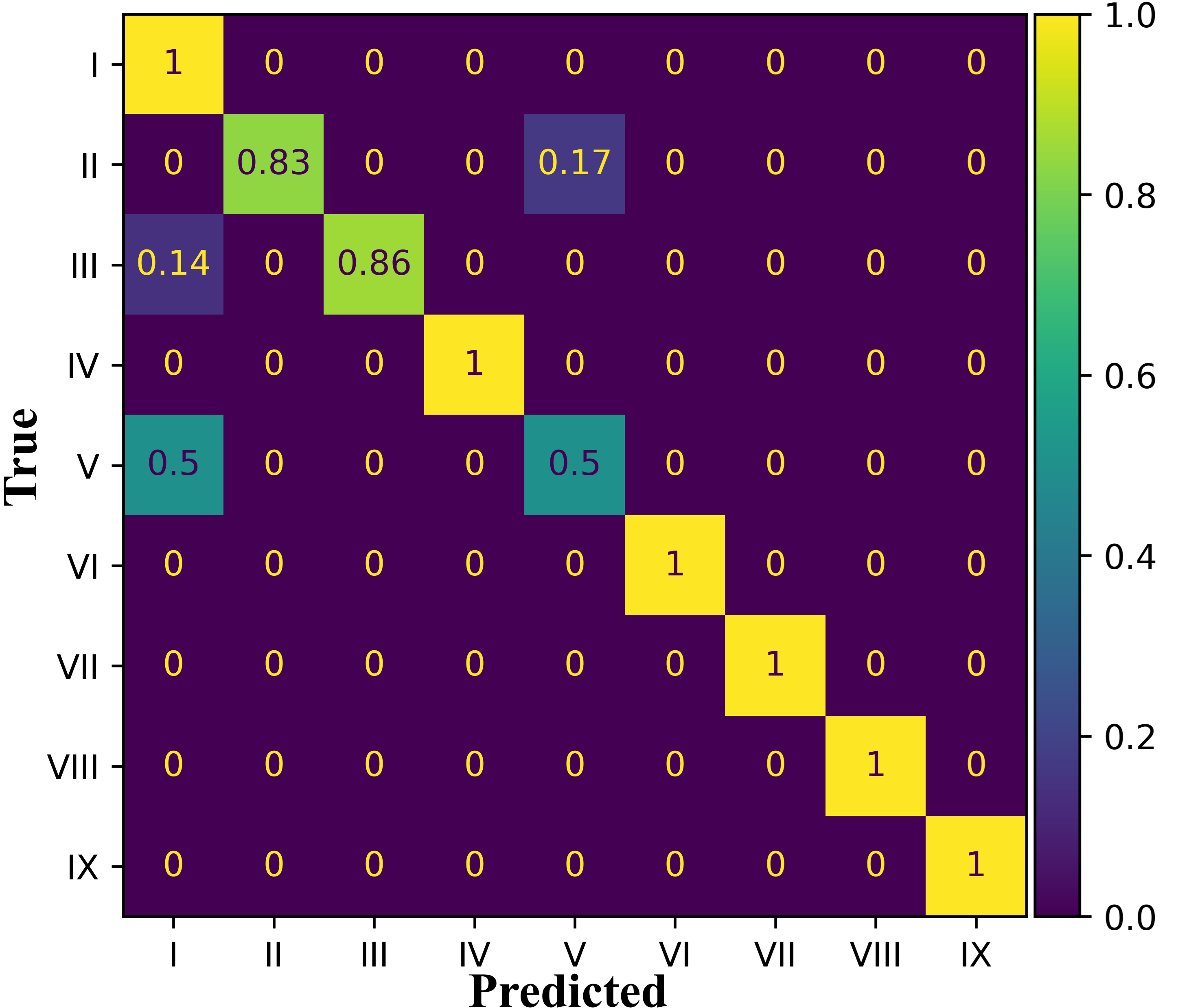}
    \caption{Normalized CFM for SiSo task}
    \label{fig_cfm_e}
    \end{subfigure}
    \hfill
    \begin{subfigure}{0.65\linewidth}
    \centering
    \includegraphics[width=1.0\linewidth]{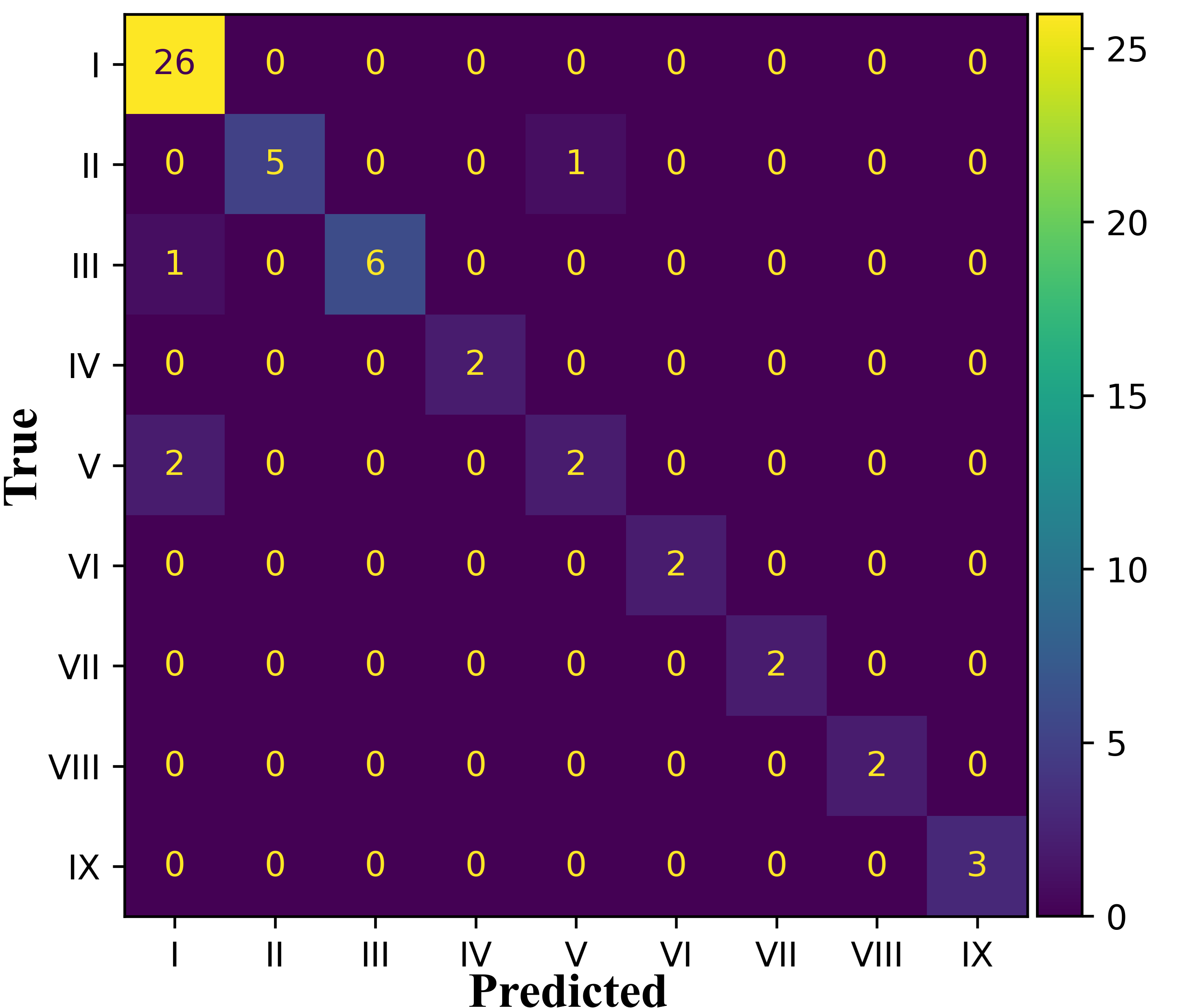}
    \caption{Absolute CFM for SiSo task}
    \label{fig_cfm_f}
    \end{subfigure}
    \begin{subfigure}{0.65\linewidth}
    \centering
    \includegraphics[width=1.0\linewidth]{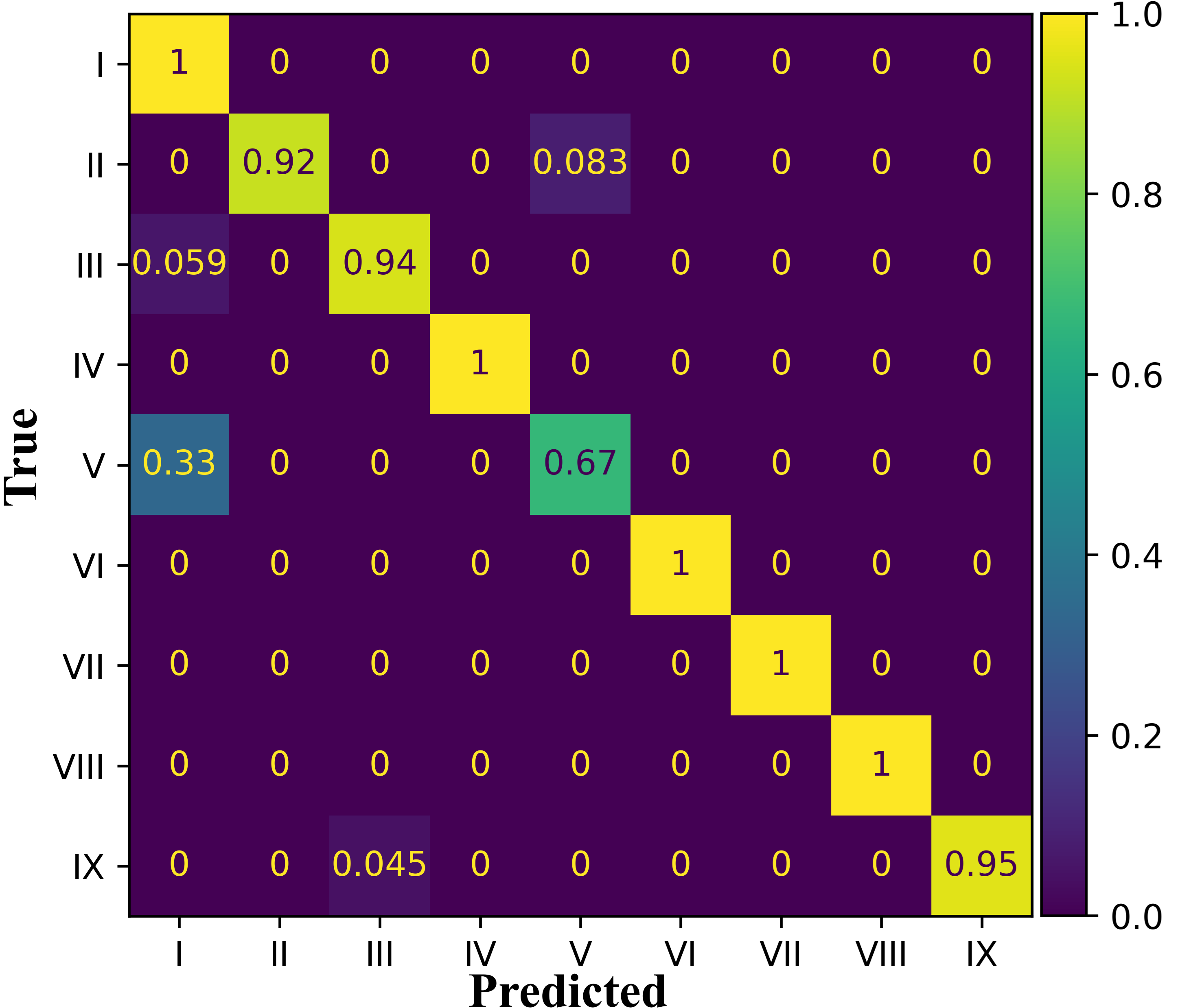}
    \caption{Normalized CFM for ViVo task}
    \label{fig_cfm_g}
    \end{subfigure}
    \hfill
    \begin{subfigure}{0.65\linewidth}
    \centering
    \includegraphics[width=1.0\linewidth]{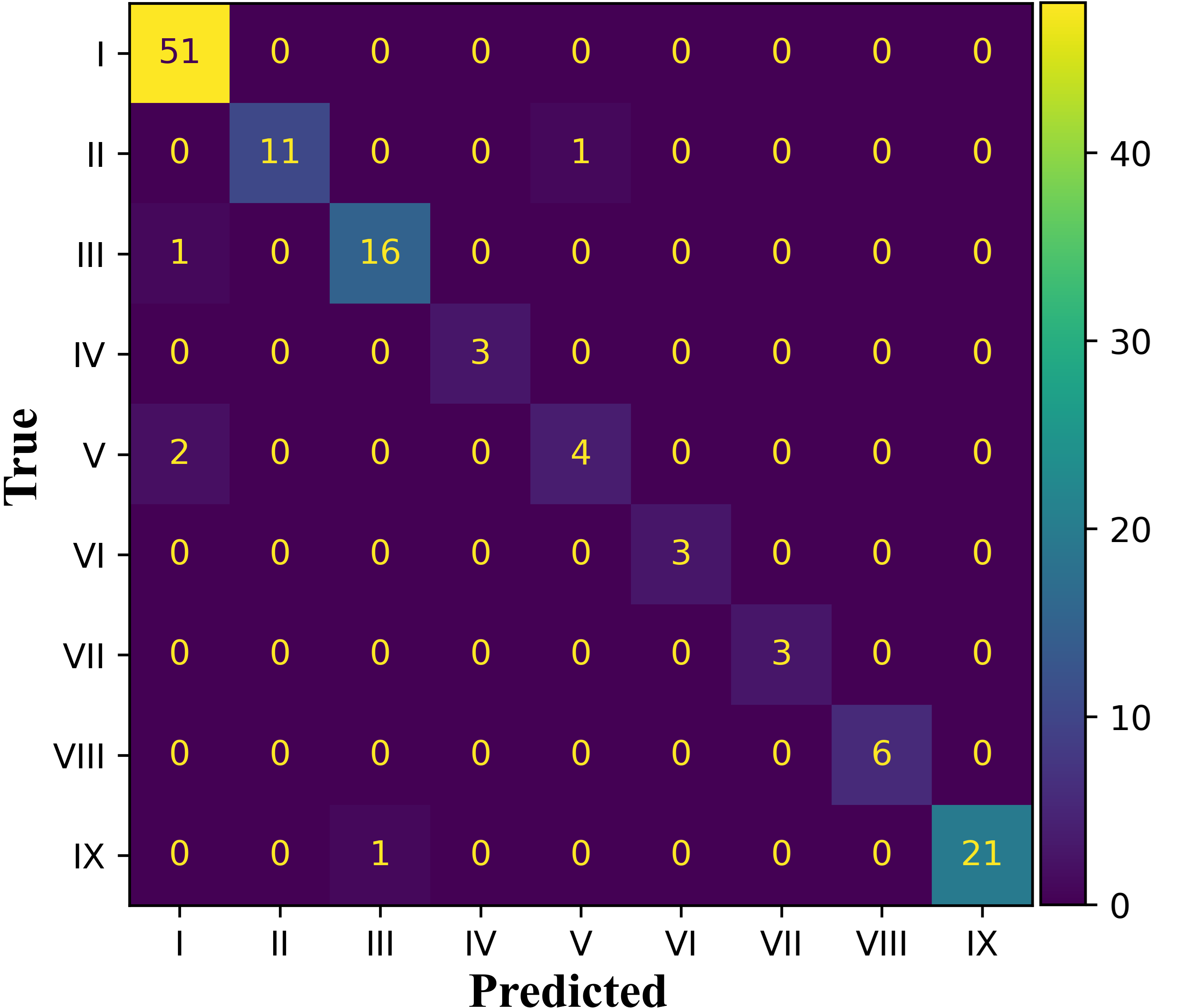}
    \caption{Absolute CFM for ViVo task}
    \label{fig_cfm_h}
    \end{subfigure}
    \end{multicols}
    \caption{\textbf{SARR--Depth-\textit{dataset*} confusion matrices for T-LESS (left) and ITODD (right).}}
    \label{fig_cfm}
\end{figure*}

\begin{figure*}[p]
    \begin{multicols}{2}
    \centering
    \begin{subfigure}{0.64\linewidth}
    \centering
    \includegraphics[width=1.0\linewidth]{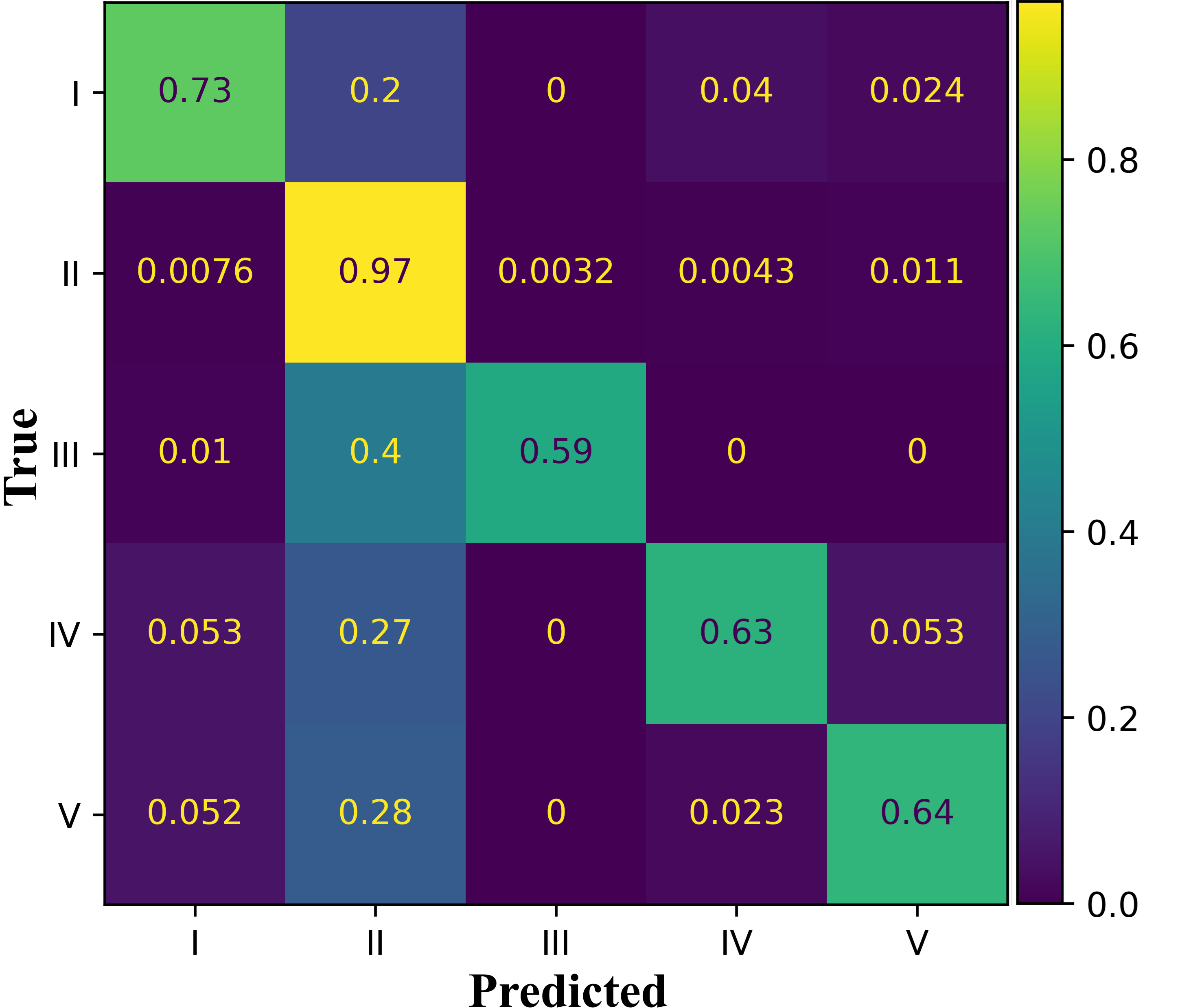}
    \caption{Normalized CFM for SiSo task}
    \label{fig_cfm_rgb_a}
    \end{subfigure}
    \hfill
    \begin{subfigure}{0.64\linewidth}
    \centering
    \includegraphics[width=1.0\linewidth]{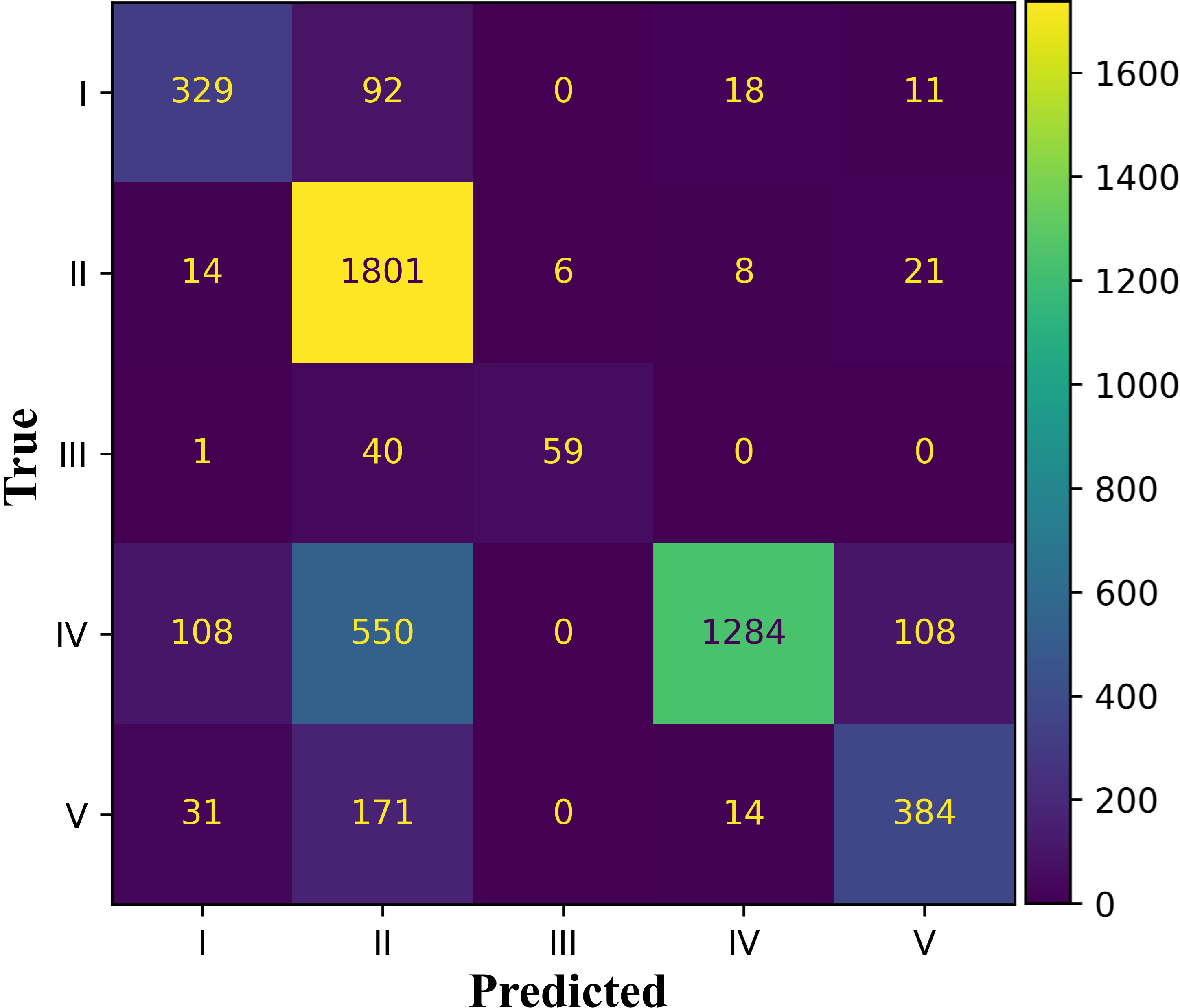}
    \caption{Absolute CFM for SiSo task}
    \label{fig_cfm_rgb_b}
    \end{subfigure}
    \begin{subfigure}{0.64\linewidth}
    \centering
    \includegraphics[width=1.0\linewidth]{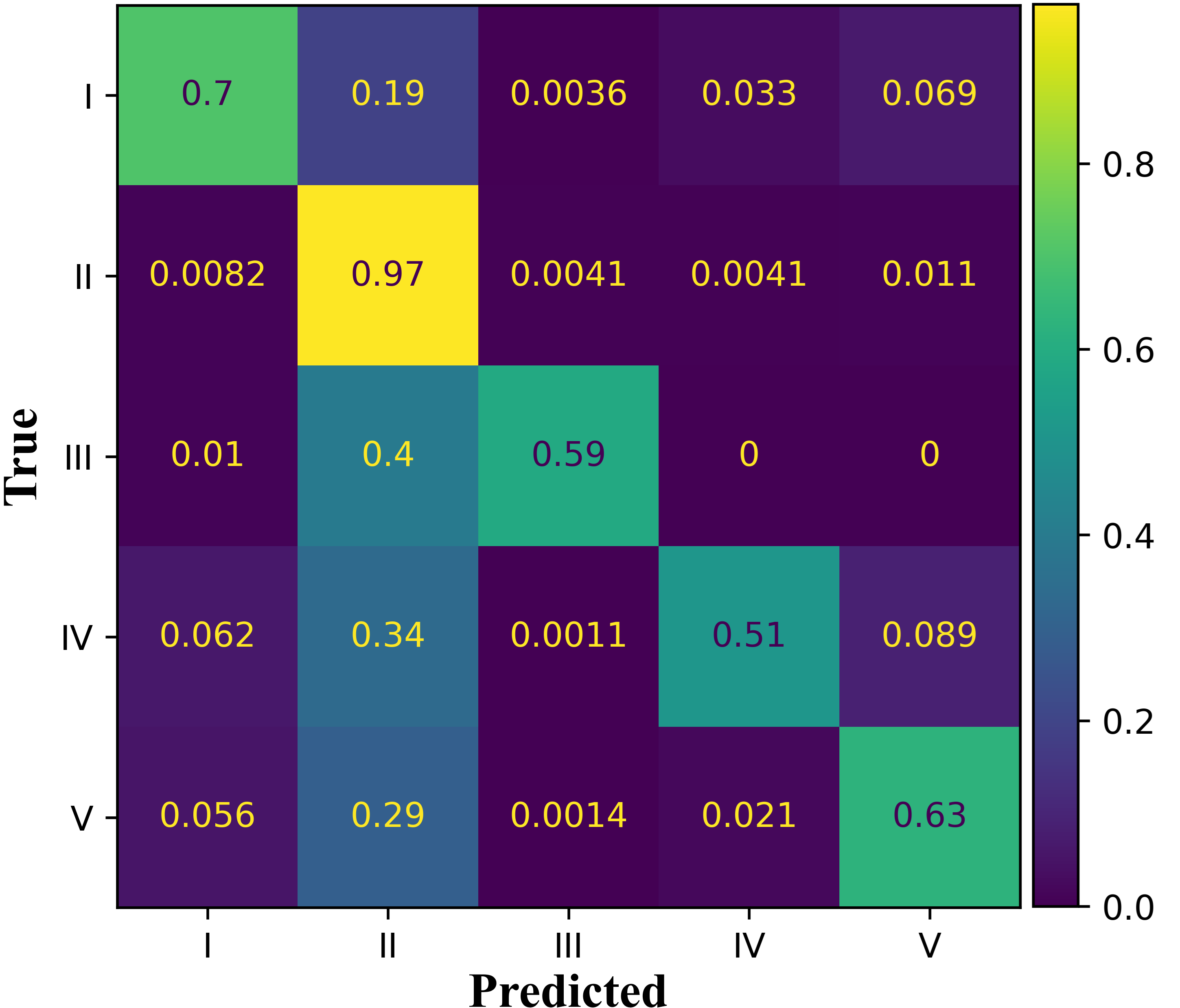}
    \caption{Normalized CFM for ViVo task}
    \label{fig_cfm_rgb_c}
    \end{subfigure}
    \hfill
    \begin{subfigure}{0.64\linewidth}
    \centering
    \includegraphics[width=1.0\linewidth]{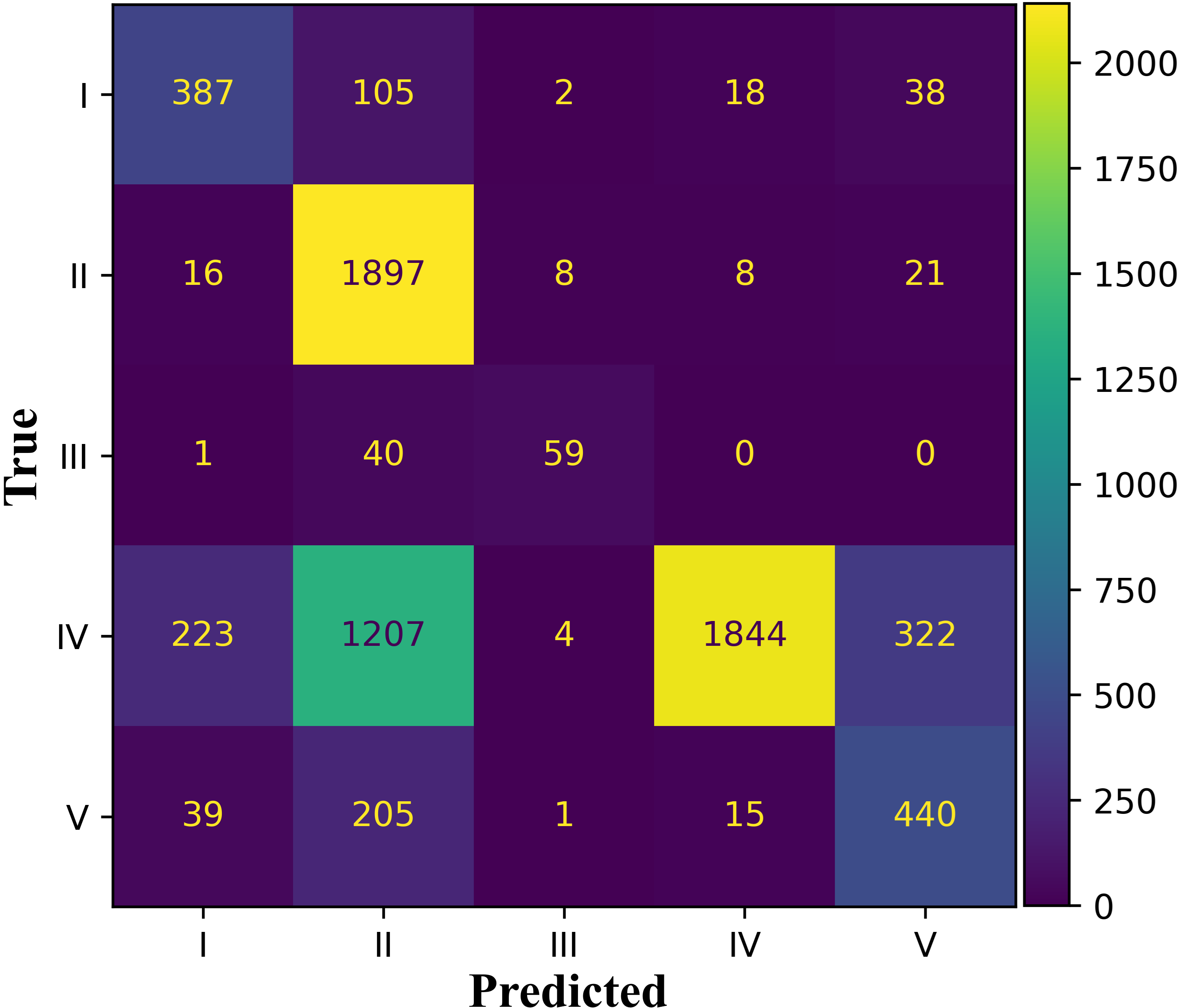}
    \caption{Absolute CFM for ViVo task}
    \label{fig_cfm_rgb_d}
    \end{subfigure}
    \begin{subfigure}{0.64\linewidth}
    \centering
    \includegraphics[width=1.0\linewidth]{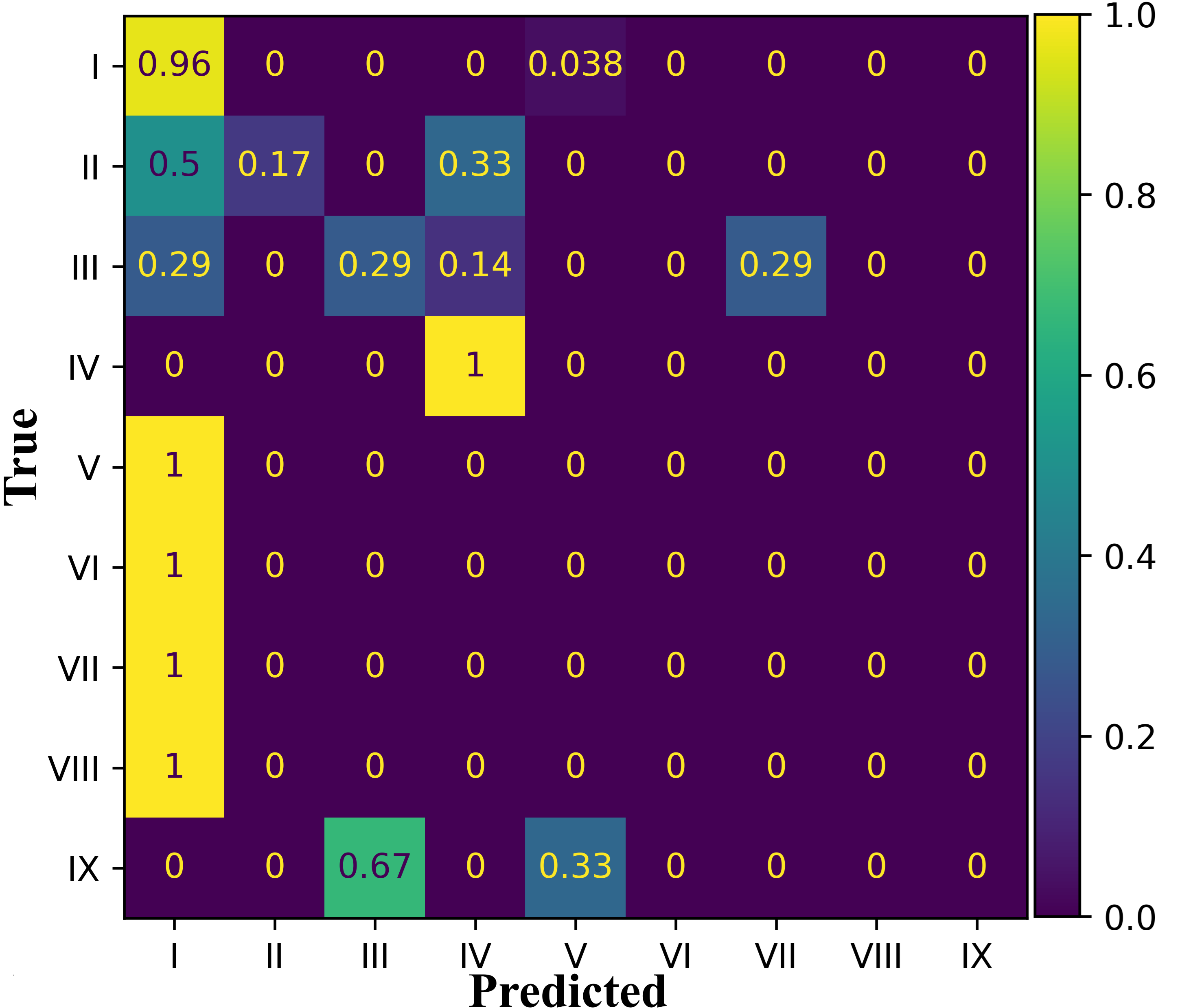}
    \caption{Normalized CFM for SiSo task}
    \label{fig_cfm_rgb_e}
    \end{subfigure}
    \hfill
    \begin{subfigure}{0.64\linewidth}
    \centering
    \includegraphics[width=1.0\linewidth]{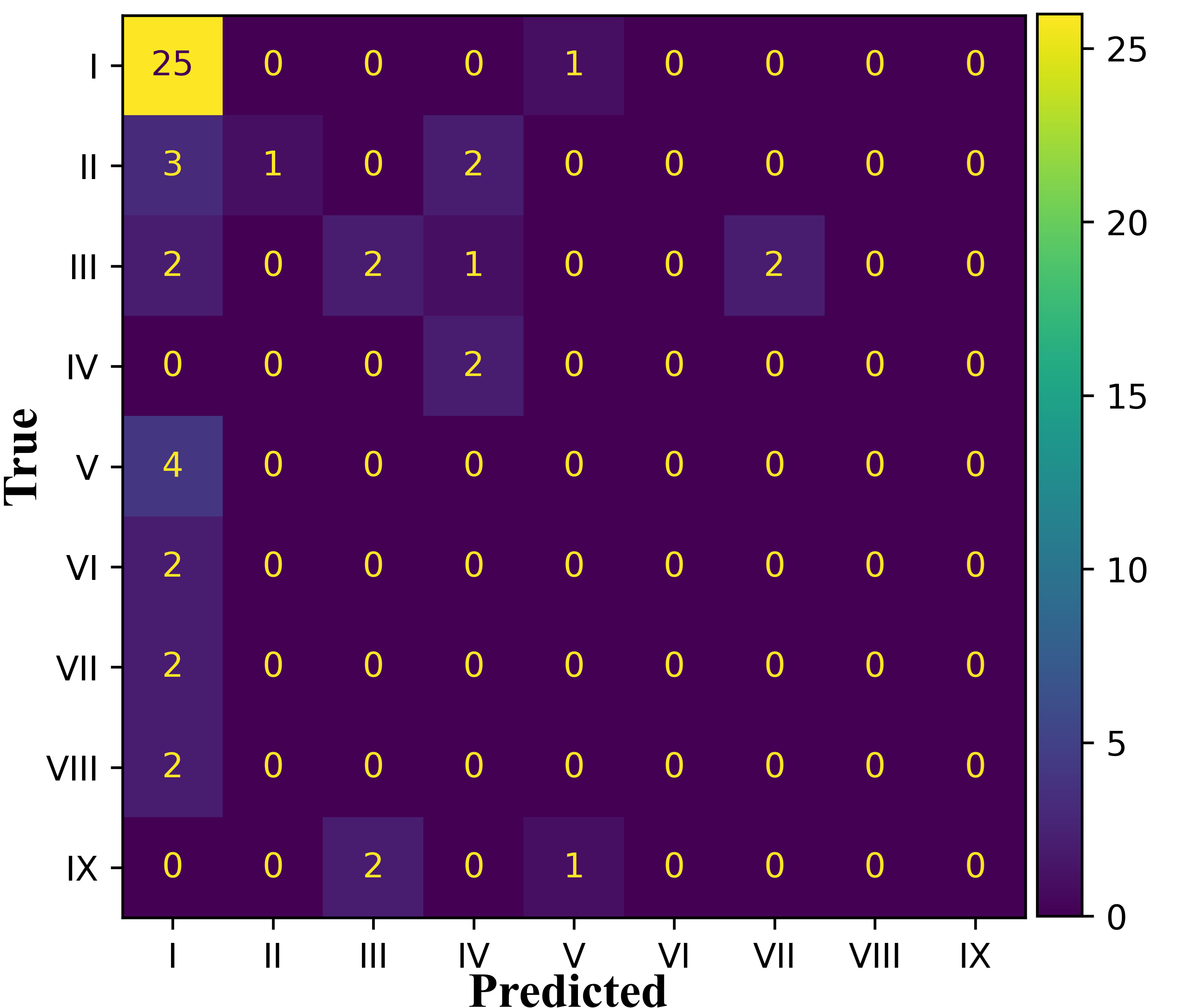}
    \caption{Absolute CFM for SiSo task}
    \label{fig_cfm_rgb_f}
    \end{subfigure}
    \begin{subfigure}{0.64\linewidth}
    \centering
    \includegraphics[width=1.0\linewidth]{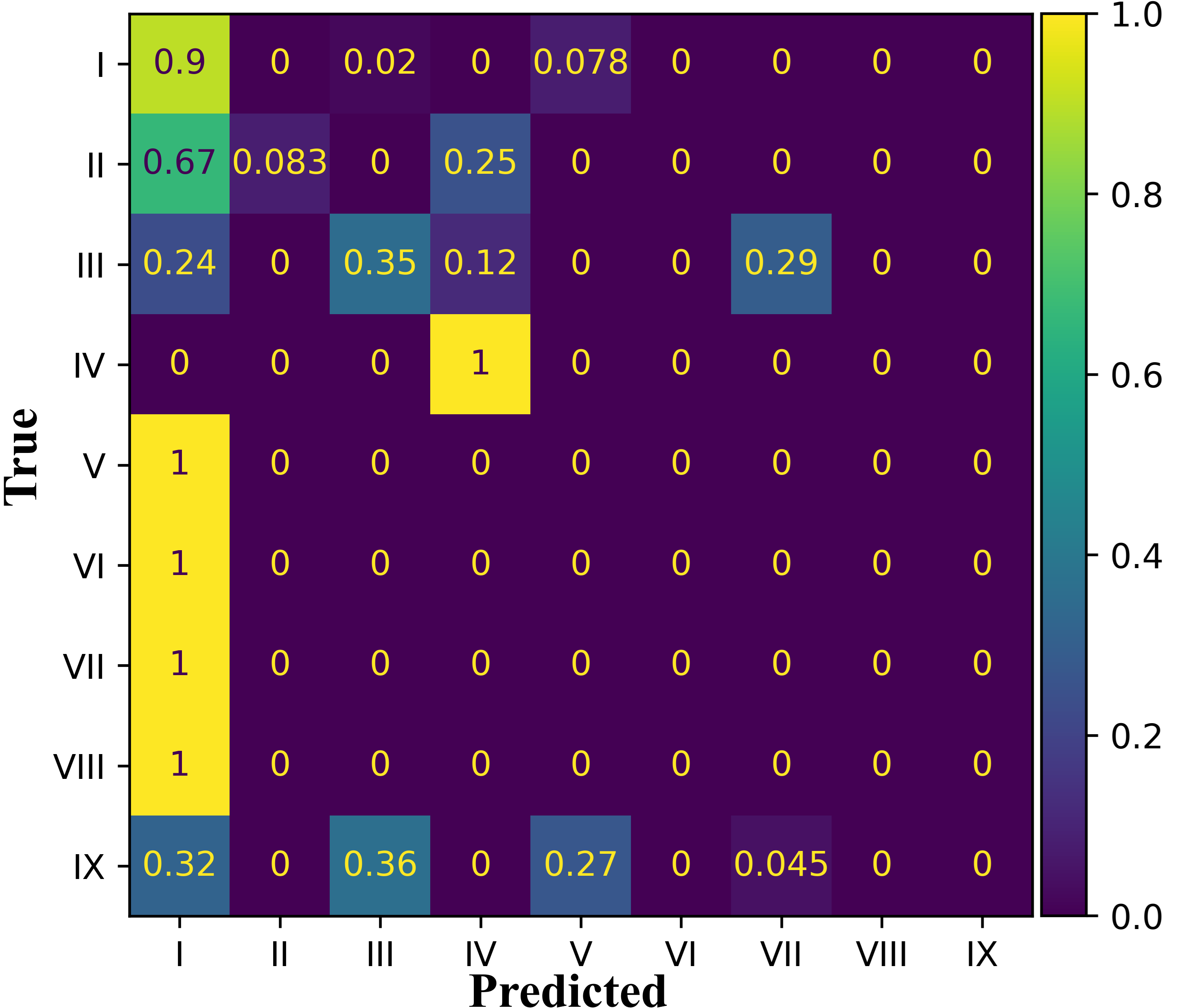}
    \caption{Normalized CFM for ViVo task}
    \label{fig_cfm_rgb_g}
    \end{subfigure}
    \hfill
    \begin{subfigure}{0.64\linewidth}
    \centering
    \includegraphics[width=1.0\linewidth]{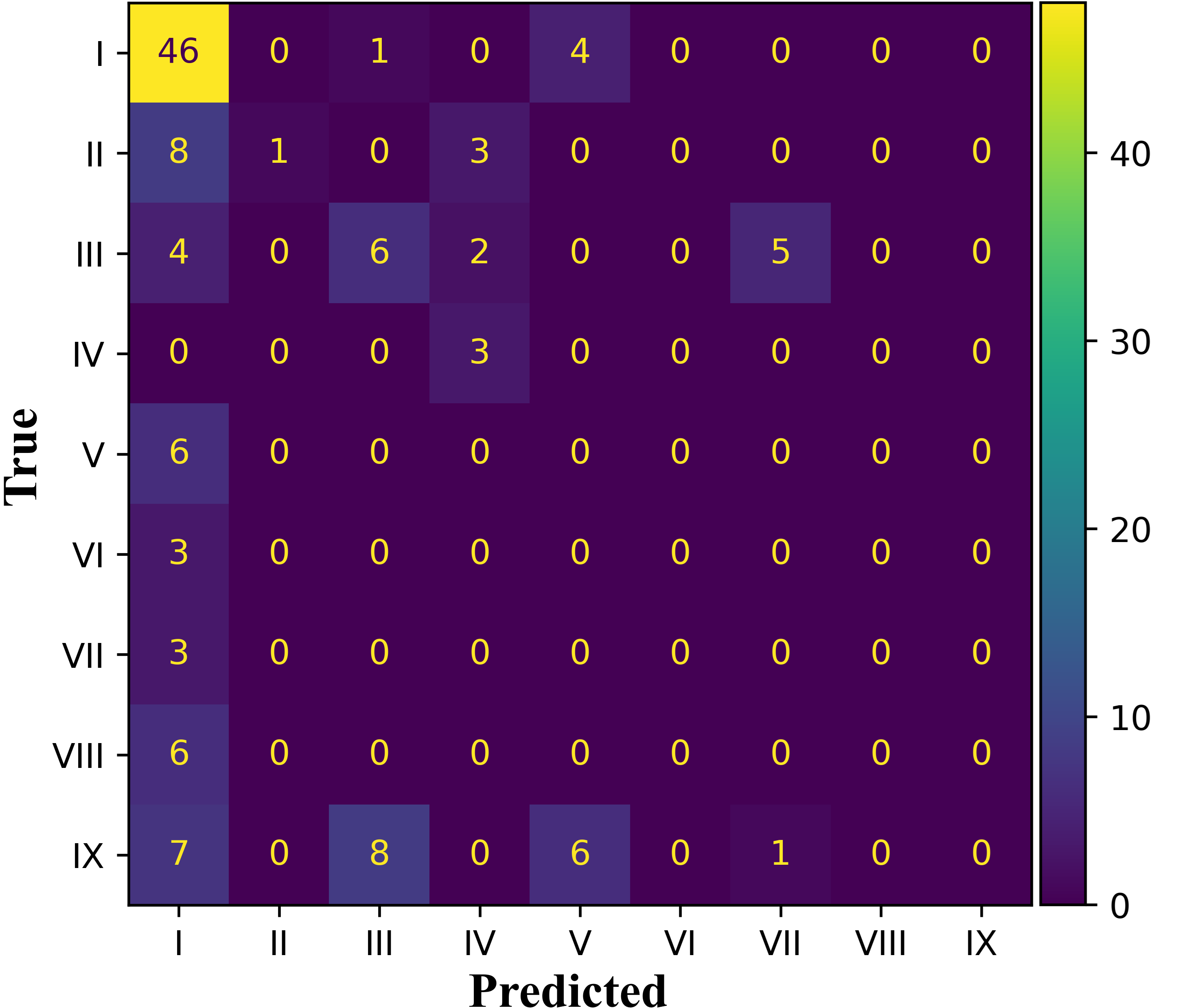}
    \caption{Absolute CFM for ViVo task}
    \label{fig_cfm_rgb_h}
    \end{subfigure}
    \end{multicols}
    \caption{\textbf{SARR--RGB-\textit{dataset*} confusion matrices for T-LESS on the left and SARR--Gray-\textit{dataset*} matrices for ITODD on the right.}}
    \label{fig_cfm_rgb}
\end{figure*}

The SARR-\textit{dataset*} networks were trained to classify the symmetry classes (5 for T-LESS and 9 for ITODD). This section provides detailed classification results.\\
\textbf{T-LESS:} As can be seen by the confusion matrices (CFMs) in \textbf{Fig.}~\ref{fig_cfm} on the left, the network achieves a mean classification accuracy of $78.6\%$ and $77.8\%$ for the SiSo and ViVo tasks, respectively. Class $\RN{1}$ gets confused with class $\RN{2}$ most likely because many fine details that break the symmetries for objects of class $\RN{1}$ are lost in the depth map (e.g. one-sided holes getting filled). Class $\RN{5}$ gets confused with  class $\RN{2}$, most likely due to the shapes being virtually identical in their symmetry, the only difference being the definition of the unrotated pose, or ``uprightness'': if the objects of class $\RN{5}$ (shown in \textbf{Table}~\ref{tab_tless_symmetries} of the main paper) were to be put upright, symmetry would be identical to class $\RN{2}$. This is amplified by the fact that these class $\RN{5}$ objects in the test images are often placed upright, so if we were to change some symmetry definitions of the dataset, we would first suggest treating objects of class $\RN{5}$ like those of class $\RN{2}$, as the two are equivalent from a symmetry standpoint. The absolute CFMs (subplots (b) and (d)) further show the strong class imbalance, with only $96$ samples belonging to class $\RN{3}$. The SARR--RGB-\textit{dataset*} network achieves mean classification accuracy of $71.2\%$ and $68.0\%$ for the SiSo and ViVo tasks, respectively, with additional objects from across the dataset erroneously classified as belonging to symmetry class $\RN{2}$. This is visible in \textbf{Fig.}~\ref{fig_cfm_rgb} on the left.\\
\textbf{ITODD:} As shown on the right side of \textbf{Fig.}~\ref{fig_cfm}, we obtained mean accuracies of $91.0\%$ and $94.2\%$ for SiSo and ViVo tasks, respectively. Class $\RN{1}$ receives the most false positives, most likely due to the strong dataset imbalance, see subplots (f) and (h). Symmetry classification accuracies of $26.9\%$ and $25.9\%$ from the SARR--Gray-\textit{dataset*} network is substantially worse compared to the depth-based counterpart, with numerous classes receiving no true-positive predictions, as is shown in \textbf{Fig.}~\ref{fig_cfm_rgb} on the right.
\end{appendices}
\end{document}